\documentclass[letterpaper,twocolumn,10pt]{article}
\usepackage{zhanggroup}

\def\BibTeX{{\rm B\kern-.05em{\sc i\kern-.025em b}\kern-.08emT\kern-.1667em\lower.7ex\hbox{E}\kern-.125emX}}

\usepackage{nicefrac}
\usepackage{siunitx}
\usepackage{array,framed}
\usepackage{booktabs}
\usepackage{
  color,
  float,
  epsfig,
  wrapfig,
  graphics,
  graphicx,
  subcaption
}
\usepackage{textcomp}
\usepackage{setspace}
\usepackage{latexsym,fancyhdr,url}
\usepackage{enumerate}
\usepackage[linesnumbered,ruled]{algorithm2e}
\usepackage{algpseudocode}
\usepackage{xparse}
\usepackage{xspace}
\usepackage{multirow}
\usepackage{csvsimple}
\usepackage{balance}

\usepackage{
  tikz,
  pgfplots,
  pgfplotstable
}
\usepackage{hyperref}

\usetikzlibrary{decorations.pathreplacing,calc}
\newcommand{\tikzmark}[1]{\tikz[overlay,remember picture] \node (#1) {};}

\usetikzlibrary{
  shapes.geometric,
  arrows,
  external,
  pgfplots.groupplots,
  matrix
}

\usepackage[absolute]{textpos}

\usepackage{mathtools}

\DeclareMathAlphabet{\mathcal}{OMS}{cmsy}{m}{n}

\usepackage{amsmath,amsfonts}

\usepackage{eucal}
\usepackage{endnotes}

\usepackage{pifont}
\usepackage{todonotes}

\usepackage[normalem]{ulem}
\usepackage{gensymb}
\usepackage{caption}
\usepackage{bbm}
\usepackage[keeplastbox]{flushend}
\usepackage{makecell}

\newcommand{\cmark}{\ding{51}}%
\newcommand{\xmark}{\ding{55}}%

\usepackage[export]{adjustbox}

\definecolor{Gray}{gray}{0.9}

\usepackage{cellspace}

\DeclareSymbolFont{cmsymbols}{OMS}{cmsy}{m}{n}
\SetSymbolFont{cmsymbols}{bold}{OMS}{cmsy}{b}{n}
\DeclareSymbolFontAlphabet{\mathcal}{cmsymbols}

\newcommand{\mypara}[1]{\smallskip\noindent\textbf{#1.} \xspace}

\usepackage{colortbl}

\usepackage{enumitem}
\setlist[itemize]{leftmargin=*}

\DeclareGraphicsExtensions{%
    .png,.PNG,%
    .pdf,.PDF,%
    .jpg,.mps,.jpeg,.jbig2,.jb2,.JPG,.JPEG,.JBIG2,.JB2}

\def\approach{{\sc MNEMON}\xspace}

\RestyleAlgo{ruled}

\pgfplotsset{compat=1.16}

\begin{document}


\begin{textblock}{13}(1.5,1)
\centering
To Appear in the 29th ACM Conference on Computer and Communications Security (CCS), November 7-11, 2022.
\end{textblock}

\def\thetitle{\Large \bf Finding \approach: Reviving Memories of Node Embeddings}
\title{\thetitle}

\date{}

\author{
{\rm Yun Shen\textsuperscript{1}\thanks{Work partially done while the author was with NortonLifeLock.}} \ \ \
{\rm Yufei Han\textsuperscript{2}} \ \ \
{\rm Zhikun Zhang\textsuperscript{3}} \ \ \
{\rm Min Chen\textsuperscript{3}} \ \ \ 
\\
{\rm Ting Yu\textsuperscript{4}} \ \ \ 
{\rm Michael Backes\textsuperscript{3}} \ \ \ 
{\rm Yang Zhang\textsuperscript{3}} \ \ \
{\rm Gianluca Stringhini\textsuperscript{5}} \ \ \ 
\\
\\
{\rm \textsuperscript{1}\textit{NetApp}} \ \ \
{\rm \textsuperscript{2}\textit{INRIA}} \ \ \
{\rm \textsuperscript{3}\textit{CISPA Helmholtz Center for Information Security}} \ \ \
{\rm \textsuperscript{4}\textit{QCRI}} \ \ \
{\rm \textsuperscript{5}\textit{Boston University}} \ \ \
}

\maketitle

\begin{abstract}
Previous security research efforts orbiting around graphs have been exclusively focusing on either (de-)anonymizing the graphs or understanding the security and privacy issues of graph neural networks.
Little attention has been paid to understand the privacy risks of integrating the output from graph embedding models (e.g., node embeddings) with complex downstream machine learning pipelines.
In this paper, we fill this gap and propose a novel model-agnostic graph recovery attack that exploits the implicit graph structural information preserved in the embeddings of graph nodes.  
We show that an adversary can recover edges with decent accuracy by only gaining access to the node embedding matrix of the original graph without interactions with the node embedding models. 
We demonstrate the effectiveness and applicability of our graph recovery attack through extensive experiments.

\end{abstract}

\section{Introduction}
\label{sec:introduction}

Many complex systems can be represented as graphs, such as social networks, communication networks, function call graphs, biomedical graphs, and the World Wide Web~\cite{QTMDWT18,LSRV20,KMBPR16}.
Graph embedding algorithms~\cite{cai2018comprehensive,goyal2018graph,zhang2018network} have been long researched to obtain effective graph representations to represent these networks concisely in low dimensional Euclidean vectors.
Upon such transformation, these embedding vectors can make graph analytics tasks efficient and facilitate numerous solutions to real world problems, e.g.,  node classification~\cite{wang2019graph}, community detection~\cite{malliaros2013clustering}, link prediction/recommendation~\cite{liben2007link},  binary similarity detection~\cite{gao2018vulseeker,zhao2018deepsim,zhou2019devign}, malware detection~\cite{fan2019graph,pektacs2020deep}, fraud detection~\cite{wang2019fdgars}, and bot detection~\cite{ali2019detect}.

It is well recognized that graphs contain sensitive and private information about the nodes (e.g., node attributes, the relationships among the nodes, etc.). 
Previous security research efforts orbiting around graphs have been focusing on either (de-)anonymizing the graphs~\cite{liu2008towards,zhou2008brief,ji2016graph} or understanding the security and privacy issues of graph neural networks~\cite{sun2018adversarial,he2021stealing,zhang2022inference,shen2022model,xi2021graph,zhang2021backdoor,wu2022linkteller}.  
Specifically, graph anonymization techniques~\cite{liu2008towards,zhou2008brief,ji2016graph} perturb the original graph data to protect users' privacy while preserving as much data utility as possible.
In contrast, graph de-anonymization techniques focus on unveiling sensitive private information from graphs. 
In recent years, inspired by the membership inference attack~\cite{shokri2017membership}, we have witnessed several successful link re-identification attacks against graph neural networks that extract private links contained in the training data via these GNN models ~\cite{he2021stealing,zhang2022inference,wu2022linkteller,zhang2021membership}.
Note that the node embeddings are not privacy preserving by design.
Yet, they are pervasively used in many graph analytics tasks as aforementioned. 
To our surprise, understanding and quantifying the privacy risks of integrating them with the complex ML pipeline v{in a model-agnostic setting} remains unexplored, hence our focus in this paper.

As such, we fill this gap and quantify the privacy risks of integrating node embeddings with downstream data analytics/machine learning pipelines.
Our attack's application scenarios (see \autoref{sec:attack_application_scenario}) lie in the complex ML systems where raw graph data is part of the learning process but cannot be directly obtained by the attackers due to data segregation policy and/or privacy policy.
Instead, the attackers only gain access to the transformed graph data (i.e., the node embeddings of the original graph).
They cannot interact with the node embedding models since such pipelines usually operate in one direction. 
For instance, the data holder may have integrated with the malicious machine learning solution providers (i.e., MLaaS providers) from the AWS Marketplace~\cite{song2017machine,malekzadeh2021honest}, or the data holder is part of a vertical federated learning environment in an enterprise~\cite{wu2022linkteller}.
In both cases, the node embeddings are part of the learning process and can be obtained by either the malicious MLaaS providers~\cite{song2017machine,malekzadeh2021honest} or the insiders~\cite{wu2022linkteller} in the pipeline. 

Concretely, our study addresses two research questions - \emph{can we recover the edges with decent accuracy from the node embedding matrix} and \emph{can we recover a graph structure that is similar to the original graph with respect to the graph properties?} - without knowledge of and the interactions with the node embedding models.
Note that these two research questions were discussed in the link re-identification attacks~\cite{he2021stealing,wu2022linkteller,duddu2020quantifying}.
They, however, follow the adversarial machine learning methodology and assume the interaction with the target model using shadow datasets and the supervision information from the feedback.
Our attack does not assume such capabilities (see \autoref{sec:threat_model}), which is more practical in the real world. 

\mypara{Our Contributions} 
In this paper, we propose \approach ~- a joint graph metric learning and self-supervised learning based graph recovery attack - to tackle these two questions. 
\approach first leverages the background information (i.e., the origin of the node embedding matrix) to estimate the average node degree.
It then uses graph metric learning with a multi-head attention mechanism to construct a data specific distance metric from a given node embedding matrix.
Coupling with graph metric learning, \approach employs graph autoencoder framework to iteratively optimize a graph structure through self-supervised graph regularization (i.e., the learning objectives are generated from the data itself).
Upon the termination of the process, the learned graph structure constitutes the recovered graph from the node embedding matrix.

We stress that our goal is not perfectly recovering a graph from its node embedding matrix.
Rather, we focus on understanding and quantifying the privacy risks of integrating them with the complex ML pipeline. 

A successful graph recovery attack can lead to severe consequences.
For instance, in the context of social networks, \approach allows an adversary to gain direct knowledge of sensitive and private social relationships.
Also, certain graph data is often expensive to obtain (e.g., protein interaction networks collected from lab studies). 
\approach can pose a direct threat to the intellectual property of the data holder as well.
In summary, we make the following contributions.
\begin{itemize}
\item We propose a novel model-agnostic graph recovery attack that exploits the implicit graph structural information preserved in the node embedding vectors. 
We show that the attacker can unveil the private and sensitive graph structural information with decent accuracy from the node embeddings.

\item We systematically define the threat model to characterize an adversary’s background knowledge and realistic application scenarios.
Extensive evaluation of four popular node embedding models using four benchmark graph datasets demonstrates the efficacy of our attacks.

\item We discuss a preliminary mitigation mechanism to defend against the graph recovery attack.
Our results demonstrate that \approach could be partially mitigated with some utility trade-off. 

\end{itemize}

\section{Preliminaries}
\label{sec:background}

\begin{table}[!t]
\centering
\caption{Summary of the notations. We use lowercase letters to denote scalars, bold lowercase
letters to denote vectors and bold uppercase letters to denote matrices.}
\resizebox{0.7\linewidth}{!} {
\begin{tabular}{l|l}
\toprule 
\textbf{Notation} & \textbf{Description} \\ \midrule
$\mathbf{G}=(\mathbf{V}, \mathbf{E}, \mathbf{X})$   & graph (network)   \\ 
$\mathbf{G}_O$ / $\mathbf{G}_R$ & original/recovered graph \\ 
$n$ & number of nodes \\ 
$\mathbf{A} \in \mathbb{R}^{n \times n}$ & (weighted) adjacency matrix \\
$\mathbf{X}$ &  node features \\ 
$v, u$ & node \\ 
$d$ & dimension of node embeddings \\ 
$\mathbf{H} \in \mathbb{R}^{n \times d}$ & node embedding matrix \\ 
$\mathbf{h}_v$ & node embedding of node $v$ \\
$t$ & $t$-th iteration \\
$k$ & (estimated) average node degree  \\
$f$ & node embedding model \\
$\phi$ & learnable embedding distance function \\ 
$\mathcal{L}$ & loss function \\
\bottomrule	 
\end{tabular}
}
\label{tab:notations}
\end{table}

\subsection{Notations}
We denote an undirected, attributed graph as $\mathbf{G}=(\mathbf{V}, \mathbf{E}, \mathbf{X})$, where $\mathbf{V}=\{v_i\}_{i=1}^n$ represents the nodes, $\mathbf{E} \subseteq \{(v, u)| v, u \in \mathbf{V} \}$ denotes the edges, and $\mathbf{X} = \{\mathbf{x}_i\}_{i=1}^n$ denotes the node features, where $\mathbf{x}_i$ represents the node feature of $v_i$. 
$|\mathbf{E}|$ denotes the graph size (i.e., the number of edges).
The original and the recovered graphs are denoted as $\mathbf{G}_O$ and $\mathbf{G}_R$ respectively. 
Let $\mathbf{A} \in \mathbb{R} ^ {n \times n}$ represent the (weighted) adjacency matrix. 
As such, $\mathbf{G}$ can also be represented as $\mathbf{G}=(\mathbf{A}, \mathbf{X})$.
The notations introduced here and used in the following sections are summarized in \autoref{tab:notations}.

\subsection{Node Embedding}
\mypara{Definition}
In this paper, we focus on \textit{node embedding}, which plays a central role in graph embedding techniques.
As the name suggests, a node embedding model $f$ maps nodes to $d$-dimensional vectors that capture their structural properties and node features (if available).
Formally, a node embedding model is defined as $f: \mathbf{G} \rightarrow \mathbf{H} $, where $\mathbf{H} \in \mathbb{R}^{n \times d}$ represents node embedding matrix where $d$ denotes the dimension of the embeddings ($d \ll n$) and $\mathbf{h}_v \in \mathbf{H}$ denotes the node embedding vector of node $v$.
The node embeddings of connected nodes maintain ``approximate closeness'' to each other in the latent space (e.g., $\mathbf{h}_v$ and $\mathbf{h}_u$ should be close in the Euclidean space if $v$ and $u$ are connected in the graph).

\mypara{Overview}
There exists abundant previous work on node embedding models~\cite{cai2018comprehensive,goyal2018graph,zhang2018network}.
Broadly speaking, these techniques can be grouped into two categories - matrix factorization based approaches and deep learning based approaches. 
\begin{itemize}
\item \emph{Matrix factorization based approaches.}
The essence of these approaches is treating node embedding as a dimensionality reduction problem and factorizing graph adjacency matrix or node proximity/similarity matrix to obtain node embedding~\cite{jolliffe2016principal}.
The core idea of these approaches is that the graph property to be preserved can be interpreted as pairwise node similarities or node proximity in a low dimensional space by matrix factorization.
In general, matrix factorization methods can be classified into two categories - node proximity matrix factorization and graph Laplacian eigenmaps factorization.
\item \emph{Deep learning (DL) based approaches.} 
The pioneer DL-based approaches include DeepWalk~\cite{perozzi2014deepwalk}, Node2Vec~\cite{grover2016node2vec}, and their variants. 
These approaches first generate a set of truncated random walk paths sampled from a graph, then apply deep learning techniques (e.g., SkipGram) to the sampled paths, consequently learning node embeddings.   
In recent years, we also witnessed the rise of Graph Neural Networks (GNNs).
These GNN models are a type of Neural Network which directly operates on the Graph structure via message passing between the nodes of graphs, and encoding the nodes into a low dimensional space (e.g., GCN~\cite{kipf2016semi}, GraphSAGE~\cite{hamilton2017inductive}).
They can take node features into consideration and do not need random walk paths. 
\end{itemize}

\noindent We refer the audience to ~\cite{cai2018comprehensive,goyal2018graph,zhang2018network} for the overview of node embedding techniques and other graph tasks (e.g., graph-level embedding, graph-level classification, etc.).

\section{Threat Model}
\label{sec:threat_model}

\subsection{Attack Setting}
\label{sec:attack_setting}

We frame our attack in a \emph{model agnostic} setting.
We assume that the adversary only has access to the \textit{node embedding matrix} $\mathbf{H}$ together with the \textit{background information} of the origin from which the embedding matrix was leaked (see \autoref{sec:attack_application_scenario} for detailed application scenario discussion).
The attackers do not have any knowledge of the node embedding model, and they cannot tamper with its internals (e.g., model parameters, model architecture).
We strictly require that the attackers cannot interact with the target model, and do not have the auxiliary data to train a shadow model using the feedback from the target model.

\mypara{Remarks}
It is important to note that our attack setting is different from the existing adversarial machine learning settings, whereas the interaction with the target model (i.e., querying the target model via publicly accessible API) and the availability of auxiliary data (e.g., nodes with features and labels, etc.) are indispensable. 
Our attack, however, assumes neither.
That is, \emph{we strictly require that the attackers cannot interact with the target model, and do not have the auxiliary data to query a target model and use the query results to train a shadow model.}
In other words, this setting eliminates the supervise information from the target model and consequently renders the previous link re-identification attacks inapplicable, hence the novelty of our attack.
We provide a detailed discussion in \autoref{sec:related_work} to distinguish our attack from the existing ones.

\subsection{Attack Scenarios}
\label{sec:attack_application_scenario}

We consider our attack's application scenarios lie in those complex ML systems where graph data is part of the learning process but cannot be directly obtained by the attackers due to data segregation policy and/or privacy policy.
As such, we discuss three real world scenarios below. 
\begin{itemize}

\item The first attack scenario is the insider threat in a complex enterprise ML environment.
In this scenario, a company enforces rigid data protection and segregation policies to guard the security of raw data.
As a result, one department may have sensitive user private profile and relationship information (i.e., graph data with node features), and another department has the user purchase history.
To train a joint model (e.g., a personalized recommender system) that leverages the data from different departments, the company needs to perform vertical federated learning~\cite{yang2019federated}. 
Instead of supplying the graph data to the central model, the department that holds the graph data may generate node embeddings that preserve the utility (i.e., user closeness without disclosing the exact edges) and facilitate the learning task.
The insider then obtains the node embeddings during this learning process and leaks them to the attackers.
This attack scenario is in line with the setting recently discussed by Wu et al.~\cite{wu2022linkteller}.

\item The second attack scenario is the malicious third-party provider that is already part of the data holder's data analytics or machine learning pipeline.
For example, the data holder may have integrated with the malicious machine learning solution providers (i.e., MLaas providers) from the AWS Marketplace.
In this case, the upstream data holder, without knowing the implications, passes the node embedding matrix to the rouge provider for downstream analytical tasks, such as data visualization, link prediction, node classification, profiling, etc.
The attackers can then obtain the node embedding matrix from the data holder through the rouge provider.
This attack scenario is in line with the malicious machine learning provider scenario discussed by Song et al.~\cite{song2017machine} and Malekzadeh et al.~\cite{malekzadeh2021honest}. 

\item The third attack scenario is security misconfiguration in the ML environment. 
For instance, researchers may leverage the free computing resources (e.g., GPUs) offered by Colab, and connect it to their private Github repository.
Due to such misconfigurations, the notebooks containing the node embeddings are leaked (i.e., wrongly using ``anyone on the Internet with this link can view'' instead of ``send to the specific users'').
This attack scenario is in line with the real world misconfigured S3 buckets leakage discussed by Continella et al.~\cite{continella2018there}.

\end{itemize}

\mypara{Background Information Acquisition} 
Besides, given the first two attack scenarios, the attackers can easily obtain the background information of the origin of the embedding matrix (e.g., from which companies the matrices come from).
With fair reconnaissance efforts (e.g., correlating the owner of Colab notebooks with Github handles), the attacker may also infer the origin of the embedding matrix in the third scenario.
In summary, these three attack scenarios are tangible and match our attack setting.  

\subsection{Attack Goals}
\label{sec:attack_goals}

The primary goal of the attackers is \emph{uncovering the edges with decent accuracy from the node embedding matrix}. 
Attaining this goal would enable the attacker to expose private and sensitive relationships among the nodes rather than the ``approximate closeness'' offered by the node embeddings (see \autoref{sec:background}).
Nevertheless, due to the strict attack setting, it is impractical for the attackers to faultlessly retrieve all the edges from the node embedding matrix.
As a result, the secondary goal of the attackers is \emph{recovering a graph structure $\mathbf{A}_R$ that is similar to the original graph $\mathbf{A}_O$ with respect to the graph properties}.
Achieving this goal would enable the attackers to gain additional knowledge of the original graph as a whole and perform graph mining tasks, which in turn violates the intellectual property of the data holder or can facilitate advanced attacks, such as re-identifying individuals~\cite{ji2016graph}, structural data de-anonymization~\cite{ji2014structural}, etc.
For example, recovering a graph with similar triangle counts and joint degree distribution to the original graph would enable the attacker to gain insights into the underlying user engagement in a social network.
This information itself is sensitive and proprietary. 

\mypara{Non-goals}
Recall our attack setting in \autoref{sec:attack_setting} that the attackers only have the node embedding matrix and the background information of the origin of the embedding matrix, and cannot interact with the target model with auxiliary data.
We thereby cannot infer node features (i.e., attribute inference attack) since we do not have any auxiliary data (i.e., we do not know the format of the original node features). 
Similarly, we cannot steal the target model (i.e., model extraction attack) nor can we understand the privacy leakage from the target model itself as we do not interact with it.
Finally, our attack focuses on the node-level embeddings.
We thus do not attack the graph-level embeddings~\cite{cai2018comprehensive,goyal2018graph,zhang2018network}.

\section{\approach: Graph Recovery Attack}
\label{sec:approach}

\subsection{Attack Overview}
\label{sec:attack_overview}

At a high level, \approach contains three main components.

\begin{itemize}

\item The first component (see \autoref{sec:estimate_average_node_degree}) leverages the background information (i.e., the origin of the node embedding matrix) to estimate the average node degree.
The goal is to estimate a rough average node degree $k$ and the graph size  (i.e., |$\mathbf{E}$| = $\frac{k \times n}{2}$).

\item The second component (see \autoref{sec:metric_learning}) leverages graph metric learning (GML) to learn a data-specific distance function since it is often difficult to choose a standard metric that fits all the datasets. 
The goal is to learn multi-head attention weights and tailor the distance function on a per node embedding matrix basis.

\item The third component (see \autoref{sec:iterative_GAE}) learns a graph structure through Graph
AutoEncoder (GAE) framework using self supervised graph regularization.
The goal is to optimize the graph structure and reduce the false positive edges incurred by the learned metric from the second component.

\end{itemize}

\noindent We iteratively optimize the second and third components as they are inter-connected.

\begin{figure}[t]
\begin{center}
\includegraphics[width=0.99\linewidth]{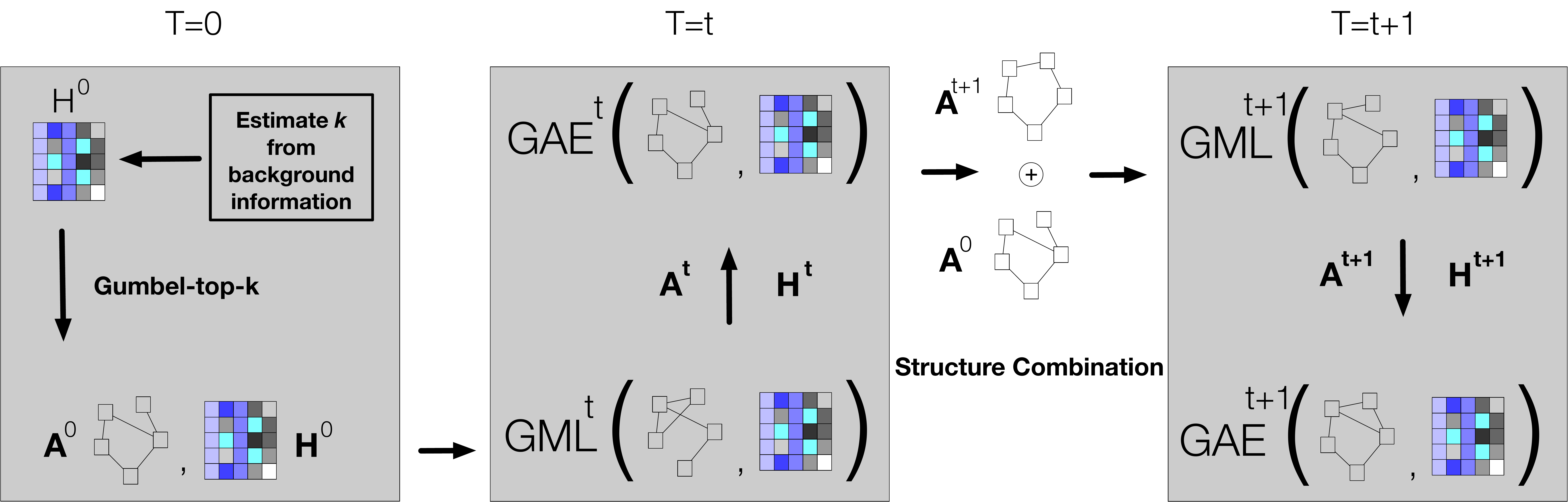}
\end{center}
\caption{
Overview of \approach. At timestamp $T=0$, \approach estimates the average node degree $k$ and initializes the seed graph using Gumbel-Top-$k$ trick. At timestamp $T=t$, it iteratively learns a data-specific graph distance metric using GML and optimizes a graph structure in a self-supervised way using GAE.
}
\label{fig:architecture}

\end{figure}

Specifically, GML learns a distance function to measure the closeness of two nodes and builds the input graph for GAE ($T=t$ in \autoref{fig:architecture}). 
GAE then learns to reconstruct this input graph.
If GAE finds certain parts of the input graph are hard to reconstruct, which is reflected by the self supervised graph learning loss, it may be due to the input graph built by GML partially capturing the graph structure. 
We then merge the graph structures by combining both the input graph and output graph of GAE, which enables us to retain the most confident edges (the transition from $T=t$ to $T=t+1$ in \autoref{fig:architecture}).
The combined graph is then used to guide GML to update its metric learning process in the next iteration.
We outline \approach's workflow in \autoref{fig:architecture}.
In the following sections, we discuss the technical details of our attack.

\subsection{Estimate the Average Node Degree}
\label{sec:estimate_average_node_degree}

The only clue that the attackers have is the background information about the origin of the node embedding matrix. 
For instance, in our first attack scenario where the node embedding matrix is leaked by an insider, it is straightforward for the attackers to obtain such background information.
The attacker's immediate task is thereby estimating the average node degree $k$. 
The rationale is straightforward.
The attackers already know the number of nodes from the embedding matrix (i.e., $n$).
Yet, due to the combinatorial nature of the graph, there exist $n(n-1)^2$ possible edges.  
As such, if the attacker can estimate the average node degree $k$, they can trivially obtain the estimated size of the original graph (i.e., the number of edges) which is equivalent to $(k \times n) / 2$.
In this way, the estimated graph size enables them to effectively learn the graph structure as we will discuss in \autoref{sec:metric_learning} and \autoref{sec:iterative_GAE}.

Abundant previous work~\cite{hay2009accurate,hardiman2013estimating,leskovec2006sampling,kleinberg2000small,dasgupta2014estimating} has already exemplified that the graphs of similar origins may share similar graph properties (e.g., node degree, graph density, small world phenomenon, local clustering coefficient, etc.).
Our core idea is that the attackers can estimate the average node degree from the graphs of similar origins and transfer the estimated node degree from these graphs to facilitate the attack.
This alleviates the attackers from stealing a sample training data from the data holder, which, in turn, makes our attack realistic.
For instance, if the attackers know that the node embeddings come from a Facebook network, they can leverage graph sampling methods to sample Facebook networks publicly available in the Network Data Repository~\cite{rossi2015the} and estimate the average node degree of the network that they target (see \autoref{sec:evaluation} for how we use graph sampling to sample real world data). 
These graph sampling methods have been proven accurate in estimating the average node degree~\cite{rozemberczki2020little}.
In this paper, we use the state-of-the-art spikyball sampling~\cite{leskovec2006sampling} implemented in the latest Little Ball of Fur python library~\cite{rozemberczki2020little} to estimate the average node degree.
Additional details can be found in \autoref{sec:exp_estimate_avg_degree}.

\mypara{Notes}
The graph sampling process does not interact with the original node embedding models.
It also does not need the supervision information from the target models as required by previous research~\cite{he2021stealing,wu2022linkteller,duddu2020quantifying}.
We also stress that \approach does not estimate or require the precise average node degree.
\approach can accommodate the inevitable estimation error.
We provide a detailed study in \autoref{sec:exp_impact_of_k} to illustrate this capability.
For instance, we show that our attack can still achieve good performance even when the estimated average node degree is twice the real average node degree (see \autoref{sec:exp_impact_of_k}) thanks to graph metric learning (see \autoref{sec:metric_learning}) and self-supervised graph structure learning (see \autoref{sec:iterative_GAE}).

\subsection{Graph Metric Learning} 
\label{sec:metric_learning}

Upon estimating the average node degree, the common approach to recover a graph from the node embedding matrix is using $k$NN algorithm. 
$k$NN builds a graph in which two nodes $v$ and $u$ are connected by an edge if the distance between the embedding vectors $h_v$ and $h_u$ is among the $k$-th smallest distances.
The drawback of $k$NN algorithm is that it requires a manually predefined distance function for neighbor selection. 
However, it is often difficult to choose a standard metric that fits all the datasets and tasks of interest. 
Take a barbell graph for example, which consists of two dense cliques connected by a long chain.
Reflected in the latent space, the node embedding vectors from two dense cliques are close to each other  (i.e., dense regions), while those from the long chain are relatively farther to each other (i.e., sparse regions).
A standard distance function, such as Eucliean or cosine distance, used by $k$NN may not recover the edges from the long chain as the distances among them are inevitably large. 
Yet, they are equally connected from a graph perspective. 
As such, we propose to leverage graph metric learning in the node embedding space to learn a data-specific distance function and automatically adjust for both the dense and sparse regions in the node embedding matrix.

\mypara{Graph Initialization with Gumbel-Top-$k$ Trick}
We follow the approach discussed by Kazi~\cite{kazi2020differentiable} to initialize a seed graph.
We first generate a fully connected graph with edge normalized distance score using \autoref{eq:edge_probability}.
\begin{equation}
    \label{eq:edge_probability}
    p_{vu} = e^{-\tau \delta(h_v, h_u)}, \forall v, u \in \mathbf{V}
\end{equation}

\noindent Here $\delta$ is a distance function and $\tau$ is a temperature parameter controlling the smoothness of the distance scores between node embedding vectors.
Instead of using the Euclidean distance function adopted by Kazi~\cite{kazi2020differentiable}, we opt for cosine distance as the distance function $\delta$, i.e., $\delta(h_v, h_u) = 1-cos(h_v, h_u)$. 

Note that most of the node embeddings are normalized to facilitate downstream tasks. 
In this case, Euclidean distance is proportional to cosine distance, given a well normalized value range of the node embeddings. 
When node embeddings are not normalized, our framework can also be adjusted to Euclidean distance.
Let $\mathbf{P} = \{ p_{vu} \}$ denote the edge probability matrix.
We then leverage Gumbel-Top-$k$ trick~\cite{kool2019stochastic} to sample from $\mathbf{P}$, which generalizes Gumbel-Max trick~\cite{gumbel1954statistical} to draw an ordered sample of size $k$ without replacement from a categorical distribution by taking the indices of the $k$ largest perturbed log-probabilities.

That is, we perturb each $p_{vu}$ by adding a Gumbel random variate $\vartheta_{vu} \sim Gumbel(0,1)$.
We then select the indices of the $k$ largest perturbed log-probabilities without replacement.
This process makes the sampling a stochastic relaxation of $k$NN~\cite{kazi2020differentiable}.
This sampled adjacency matrix (denoted as $\mathbf{A}^0$) constitutes our seed graph structure. 
Note that $\delta$ is used for sampling purposes only and is not part of our learning targets. 
This corresponds to $T=0$ in \autoref{fig:architecture}.

\mypara{Learnable Distance Function ($\phi$)}
Due to the stochastic nature of Gumbel-Top-$k$ trick, we inevitably obtain an initial noisy graph structure from the above graph initialization process. 
That is, an edge $(v, u)$ in $\mathbf{A}^0$ may not exist in the original graph $\mathbf{G}_O$, i.e., a false positive edge.
To reduce such false positives, we propose a learnable distance function $\phi$ to learn a better graph structure.
The core idea is that, instead of using a predefined distance function, we leverage metric learning~\cite{yang2006distance} to learn a distance metric for the input space of data (i.e., the node embedding matrix $\mathbf{H}$) from the adjacency matrix $\mathbf{A}$ that preserves the node relationships (i.e., $\mathbf{A}$ is used to supervise the distance learning).
In this paper, we adopt a weighted cosine distance (defined in \autoref{eq:phi})~\cite{zhao2021heterogeneous,chang2021sequential} as our learnable distance function $\phi$.

\begin{equation}
    \label{eq:phi}
    \phi(\mathbf{h}_v, \mathbf{h}_u) = 1 - cos(\mathbf{w} \circ \mathbf{h}_v, \mathbf{w} \circ \mathbf{h}_u)
\end{equation}

\noindent Here $\mathbf{w}$ is a learnable weight vector that is the same dimension as $\mathbf{h}_v$ and $\mathbf{h}_u$, and $\circ$ denotes the Hadamard product. 
Following the procedure discussed in~\cite{chen2020iterative,vaswani2017attention}, we further extend \autoref{eq:phi} to a multi-head version as in \autoref{eq:phi_multi_head} to increase the expressiveness and stablize the learning process.

\begin{equation}
    \label{eq:phi_multi_head}
    \phi(\mathbf{h}_v, \mathbf{h}_u) = 1 - \frac{1}{m} \sum_{i=1}^{m} cos(\mathbf{w}_i \circ \mathbf{h}_v, \mathbf{w}_i \circ \mathbf{h}_u)
\end{equation}

\noindent Here $m$ refers to the number of attention heads.
In this way, we can learn the distance function from multiple perspectives.
Note that all node embeddings share the same metric parameters $\mathbf{W} = \{\mathbf{w}_i\}_{i=1}^m$. 

\mypara{Graph Sparsification}
We plug $\phi$ into \autoref{eq:edge_probability} (i.e., replacing $\delta$) and use the aforementioned Gumbel-Top-$k$-based sampling trick to extract an adjacency matrix.
Our graph sparsification method is different from the $\epsilon$-neighborhood approach used by~\cite{chen2020iterative} which cannot easily control the graph size (i.e., $\epsilon$ is fixed and may lead to different graph sizes as the learned weighted adjacency matrix also evolves during the learning process).

\subsection{Self-Supervised Graph Structure Learning}
\label{sec:iterative_GAE}

Having introduced how we apply the graph metric learning technique to tune a data-specific distance function in the previous section, we move on to discuss how we optimize a graph structure and learn the graph distance metric jointly via self-supervised learning.
The core idea is that we refine the initial noisy graph structure through self-supervised graph regularization. 
To this end, we propose to use Graph AutoEncoder~\cite{kipf2016variational} (GAE) with an adaptive graph structure combination mechanism to iteratively refine the graph structure learned from the node embedding matrix.

\begin{algorithm}[t]
\LinesNumbered
\SetKwInOut{Input}{Input}
\SetKwInOut{Output}{Output}
\caption{Graph recovery attack \approach}
\label{alg:attack}
\Input{Node embedding matrix $\mathbf{H}^0$, background information $B$, maximum iteration $T$, 
hyperparameters $\tau, \alpha$, $\beta$, $\eta$, $m$}
\Output{Learned graph structure $\mathbf{A}^T$ (i.e., $\mathbf{G}_R$)}

$k \gets EstimateAvgDegree(B)$ \\ 
$\mathbf{A}^0 \gets$ Apply Gumbel-Top-$k$ trick on the fully connected probabilistic graph $\mathbf{P}$ with $\tau$ (\autoref{eq:edge_probability}) to generate the initial seed graph

\For{$t \leftarrow 0$ \KwTo $T-1$}{
$\mathbf{A}^t \gets GML(\mathbf{A}^{t}, \mathbf{H}^{t}, m)$  \\  
$\mathbf{A}^{t+1}, \mathbf{H}^{t+1} \gets GAE(\mathbf{A}^t, \mathbf{H}^t)$  \\ 
$\mathcal{L} \gets \mathcal{L}_{lap} (\mathbf{A}^{t+1}, \mathbf{H}^{0}) + \mathcal{L}_{spa} (\mathbf{A}^{t+1}, \alpha, \beta)$  \\ 
\Indp \Indp + $\mathcal{L}_{rec} (\mathbf{A}^{t+1}, \mathbf{A}^{t} )$ \\
\Indm \Indm Backpropagate $\mathcal{L}$ \tikzmark{bottom} \\ 
$\mathbf{A}^{t+1} \gets Combine(\mathbf{A}^{0}, \mathbf{A}^{t+1}, \eta)$ \\
$\mathbf{A}^{t+1} \gets Binarize(\mathbf{A}^{t+1})$ \\ 
}

\end{algorithm}

\mypara{Graph Autoencoder (GAE)}
Given the adjacency matrix roughly estimated using the multi-headed distance metric in Eq.\ref{eq:phi_multi_head} and the node embedding vectors as the input, GAE learns to refine the adjacency matrix as the output.
The initial input of our GAE is $\mathbf{G}^0 = (\mathbf{A}^0, \mathbf{H}^0)$. 
Here $\mathbf{H}^0$ represents the node embedding matrix obtained by the attackers. 
$\mathbf{A}^{0}$ represents the initialized seed graph. 
In this way, we treat $\mathbf{H}^0$ as the node features $\mathbf{X}$ of $\mathbf{G}^0$.
Note that we add a superscript for ease of description of the following iterative learning process. 

\begin{itemize}

\item \emph{Encoder.} 
The encoder is a $Z$-layer graph convolutional network (GCN)~\cite{kipf2016semi}.
At the $t$-th iteration, its input is a graph $\mathbf{G}^t = (\mathbf{A}^t, \mathbf{H}^t)$. 
The encoder (see \autoref{eq:encoder}) learns a latent representation $\mathbf{H}^{t+1} \in \mathbb{R} ^{n \times d}$ where each row represents a node $v$'s latent representation after encoding.
\begin{equation} 
\label{eq:encoder}
\mathbf{H}^{t+1} = GCN(\mathbf{A}^t, \mathbf{H}^t)
\end{equation}

\item \emph{Decoder.} 
We use an inner-product decoder in this paper~\cite{kipf2016semi}.
The adjacency matrix can be reconstructed using \autoref{eq:decoder}, where  $\sigma(x) = 1/(1+e^{-x})$ and the output $\mathbf{A}^{t+1}$ is a weighted adjacency matrix.

\begin{equation} 
\label{eq:decoder}
\mathbf{A}^{t+1} = \sigma(\mathbf{H}^{t+1} \mathbf{H}^{{t+1}^T})
\end{equation}

\end{itemize}

\noindent Note that GAE is a generic framework. 

We follow the design by Kipf et al.~\cite{kipf2016variational} and use GCN as the encoder and inner-product as the decoder.
This design allows us to use linear GCN~\cite{salha2021simple} to accelerate the computation and compare to our baseline~\cite{duddu2020quantifying} in \autoref{sec:evaluation}.
The adversary can plug in other GNN models into GAE framework.
The audience can use different architectures as encoders and decoders.

\mypara{Self-Supervised Graph Regularization}
\approach cannot interact with the target model (i.e., the node embedding model).
We therefore rely on several graph regularization objectives to guide the above GAE-based learning process in a self-supervised way.  

\begin{itemize}
\item Graph Laplacian regularization ($\mathcal{L}_{lap}$)~\cite{belkin2001laplacian}. 
A graph Laplacian regularization assumes that the learned weighted adjacency matrix is smooth with respect to a set of node features. 
In our case, the weighted adjacency matrix is $\mathbf{A}^{t+1}$ and the set of node features is the node embedding matrix $\mathbf{H}^{0}$.
Note that our goal is to optimize the graph structure (i.e., $\mathbf{A}^{t+1}$).
As we can see in \autoref{eq:graph_laplacian_regularization}, we stress that we always force that the learned weighted adjacency matrix $\mathbf{A}^{t+1}$ is smooth with respect to the initial node embedding matrix $\mathbf{H}^{0}$. 
As such, graph Laplacian regularization can be interpreted that two connected nodes in the learned graph structure should be close enough in the latent node embedding space defined by $\mathbf{H}^{0}$.

\begin{equation} 
\label{eq:graph_laplacian_regularization} 
\mathcal{L}_{lap} (\mathbf{A}^{t+1}, \mathbf{H}^{0})  = \frac{1}{2n^2} \sum_{v,u} \mathbf{A}_{vu}^{t+1} \|h_v^{0} - h_u^{0}\| = \frac{1}{2} \mathbf{tr}(\mathbf{H}^{{0}^T} \mathbf{L}^{t+1} \mathbf{H}^{0})
\end{equation}

\noindent where $\mathbf{tr}$ denotes trace of matrix, $\mathbf{L}^{t+1} = \mathbf{D}^{t+1} - \mathbf{A}^{t+1}$ and $\mathbf{D}^{t+1} = \sum_v \mathbf{A}_{vu}^{t+1}$.

\item Graph sparsity regularization ($\mathcal{L}_{spa}$)~\cite{kalofolias2016learn}.
In the real world, the graphs are normally sparse.
We use graph sparsity regularization proposed by Kalofolias et al.~\cite{kalofolias2016learn}  to learn graphs that meet such expectations. 
As we can see in \autoref{eq:graph_sparsity_regularization}, graph sparsity regularization encourages that each node connects to at least another node in the first term, and penalizes large degrees in the second term naturally arising from the first term.
Graph sparsity regularization can be interpreted as using $\alpha$ to force the graph degrees to be positive and $\beta$ to control the graph sparsity. 

\begin{equation} 
\label{eq:graph_sparsity_regularization} 
\mathcal{L}_{spa} (\mathbf{A}^{t+1}, \alpha, \beta) = -\alpha \mathbf{1}^T log(\mathbf{A}^{t+1} \mathbf{1}) + \frac{\beta}{2} \|\mathbf{A}^{t+1}\| 
\end{equation}

\noindent where $\alpha > 0$ and $\beta \geq 0$ are two controlling hyperparameters.

\item Graph reconstruction loss ($\mathcal{L}_{rec}$)~\cite{kipf2016variational}. 
Graph reconstruction loss forces GAE to learn a latent representation $\mathbf{H}^{t+1}$ to faithfully rebuild the input adjacency matrix $\mathbf{A}^{t}$. 

In this paper, we adopt the link prediction as the way to interpret the reconstruction loss~\cite{liu2019graph} and minimize the binary cross entropy loss between negative (i.e., non-existing edges) and positive samples (i.e., existing edges).
The loss function can be found in \autoref{eq:graph_reconstruction_loss}.

\begin{equation} 
\label{eq:graph_reconstruction_loss} 
\mathcal{L}_{rec} =\frac{1}{2n^2}  \| \mathbf{A}^{t} {\circ} log(A^{t+1}) + (\mathbf{1}-\mathbf{A}^{t}) {\circ} log (\mathbf{1}-\mathbf{A}^{t+1}) \|_F ^2
\end{equation}

where $\circ$ is elementwise product and $\mathbf{1}$ is an all-ones matrix.

\end{itemize}
\noindent 

\noindent To summarize,  $\mathcal{L}_{rec}$ forces GAE to learn simultaneously the updated latent representation $\mathbf{H}^{t+1}$ and a graph adjacency matrix $\mathbf{A}^{t+1}$ decoded from $\mathbf{H}^{t+1}$ to faithfully rebuild the input adjacency matrix $\mathbf{A}^{t}$.
$\mathcal{L}_{lap}$ and $\mathcal{L}_{spa}$ makes the learned graph smooth and sparse.
Note that all these three supervisory signals are from the data itself. 

\mypara{Learning Objective} \autoref{eq:loss_function} summarizes the objective function of the self-supervised graph structure learning. 

\begin{eqnarray} 
\label{eq:loss_function} 
\mathcal{L} & = & \mathcal{L}_{lap} + \mathcal{L}_{spa} + \mathcal{L}_{rec} \nonumber \\
\mathbf{W}^* , ~\mathbf{A}^{*} & = & \underset{\mathbf{W}, ~\mathbf{A}} {\arg\,\min} ~~~~\mathcal{L}(\mathbf{W} ,\mathbf{A},\mathbf{H}^{0}) 
\end{eqnarray}

\noindent 
By minimizing \autoref{eq:loss_function}, we can jointly refine the graph structure and the graph metric function $\phi$. 
The learning process is executed with two alternating steps. 
First, we refine the distance metric $\phi$ by updating the multi-heads parameters $\mathbf{w}_{i=1 \dotsc m}$, given the current estimation of the graph structure (\autoref{sec:metric_learning}). 
Second, we estimate the graph structure using the current distance metric $\phi$ (\autoref{sec:iterative_GAE}). 
The two steps are complementary to each other and boost the overall accuracy of graph structure recovery. 

It is worth noting that all three loss functions are empirically comparable in magnitude in our evaluation. 
The adversary can weigh the losses to accommodate their specific attack targets in \autoref{eq:loss_function}.

\mypara{Adaptive Graph Structure Combination}
The learned weighted graph structure $\mathbf{A}^{t+1}$ is then combined with the input graph structure $\mathbf{A}^0$ using \autoref{eq:gae_combination}.
This structure combination step can be interpreted as a denoising function to reduce false positive edges incurred by the initial adjacency matrix $\mathbf{A}^0$.
That is, GAE learns to reconstruct a graph structure $\mathbf{A}^{t+1}$ given its structure $\mathbf{A}^{t}$ and node feature $\mathbf{H}^{t}$.
The edges reconstructed with high confidence are likely to appear the original graph.
we use $\eta$ in \autoref{eq:gae_combination} to control the update rate of $\mathbf{A}^{0}$ using $\mathbf{A}^{t+1}$, and iteratively filter out the false positive edges from the initial graph structure.

\begin{equation} 
\label{eq:gae_combination} 
\mathbf{A}^{t+1} = (1- \eta) \mathbf{A}^{0} + \eta \mathbf{A}^{t+1}
\end{equation}

\noindent Note that $\mathbf{A}^{t+1}$ remains a weighted adjacency matrix after combination.
At the end of each iteration, however, we need to obtain the learned graph structure in a binary form to guide graph metric learning in the next iteration.
To this end, we first apply an entrywise clipping function, $clip(x) = min(max(0, x), 1)$, to $\mathbf{A}^{t+1}$.
We then use the same Bernoulli binarization strategy outlined in~\cite{chanpuriya2021deepwalking} to obtain the binary adjacency matrix $\mathbf{A}^{t+1}$.

Specifically, we treat each element of the weighted adjacency matrix $\mathbf{A}^{t+1}$ as the parameter of a Bernoulli distribution and sample independently to produce the final binary adjacency matrix.

\mypara{Summary}
We summarize the whole learning process (i.e., \autoref{sec:estimate_average_node_degree}, \autoref{sec:metric_learning} and \autoref{sec:iterative_GAE}) in \autoref{alg:attack}.
Additional details (e.g., complexity analysis) can be found in \autoref{sec:attack_algorithm}.

\section{Evaluation}
\label{sec:evaluation}

\subsection{Experimental Setup}
\label{sec:exp_setup}

\mypara{Datasets} 
We use 4 public benchmark datasets to evaluate the performance of our graph reconstruction attack, including Cora~\cite{yang2016revisiting}, Citeseer~\cite{yang2016revisiting}, Actor~\cite{pei2020geom}, and Facebook~\cite{mcauley2012learning}. 
Cora and Citeseer are citation networks with nodes representing publications and edges indicating citations among these publications.
Actor is the actor-only induced subgraph of the film-director-actor-writer network used in~\cite{pei2020geom}. 
Each node corresponds to an actor, and the edge between two nodes denotes co-occurrence on the same Wikipedia page.
Facebook is a social network where nodes represent Facebook users and edges are friendships. 
We use these datasets to verify the efficacy of our attack given graphs with different characteristics (e.g., origin, graph size, density, node feature size, etc.). 
For example, Facebook is a social network, it has a well known small world phenomenon and tight community structures among the nodes while the other networks are relatively sparse. 
Statistics of these datasets are summarized in \autoref{tab:graph_stats}.

\begin{table}[t]
\centering
\caption{Summary of datasets.}
\resizebox{0.95\linewidth}{!} {
\begin{tabular}{lcccccc}
\toprule
\textbf{Dataset} & \textbf{Category} & $|\mathbf{V}|$ & $|\mathbf{E}|$ & $|\mathbf{X}|$ & $\lceil |\mathbf{E}|$/$|\mathbf{V}| \rceil$ & \textbf{Density} \\
\midrule
\textbf{Cora} & Citation  & 2,708 & 5,429 & 1,433  & 4 & 0.0014         \\
\textbf{Citeseer} & Citation    & 4,230 & 5,358 & 602 & 3 & 0.0006 \\
\textbf{Actor} & Co-Occurrence    & 7,600 & 33,544 & 931 & 9 & 0.0011         \\
\textbf{Facebook} & Social       & 4,039 & 88,234  & 1,283 &  43 & 0.011  \\
\bottomrule
\end{tabular}
}
\label{tab:graph_stats}
\end{table}

\mypara{Node Embedding Models ($f$)}
We use four popular node embedding models - network embedding as sparse matrix factorization (NetSMF)~\cite{qiu2019netsmf}, Deepwalk (abbreviated as DW)~\cite{perozzi2014deepwalk}, Node2Vec (abbreviated as N2V)~\cite{grover2016node2vec} and graph convolutional network (GCN)~\cite{kipf2016semi} - to generate node embeddings for our evaluation.
These four node embedding models are representative of the existing node embedding model families. 
Network embedding as sparse matrix factorization (NetSMF)~\cite{qiu2019netsmf} improves NetMF~\cite{qiu2018network} and represents the state-of-the-art matrix factorization based approach to generate node embeddings.
Deepwalk and Node2Vec are two well known shallow neural network-based (i.e., a neural network with one hidden layer) node embedding techniques.
Graph convolutional network (GCN) is a widely used deep neural network based approach for graph representation learning.
Note that NetMF, Deepwalk, and Node2Vec generate node embedding using graph structural information only, while GCN considers both node feature and graph structure.
As such, these models also cover different real world use cases whereas node embeddings can be generated with different inputs.
For reproducibility purposes, we outline their details below.

\begin{itemize}
    \item \textbf{NetSMF.}
    We use the Pytorch implementation by the original authors \cite{cen2021cogdl}.
    The window size of approximate matrix is 10.
    The number of negative nodes in sampling is 1.
    We run the path sampling algorithm for 100 iterations.
    \item \textbf{Deepwalk.}
    We use the DGL implementation of Deepwalk.
    The learning rate is set to 0.1.
    The number of negative nodes in sampling is 5.
    The random walk length is fixed at 80, and we run 10 random walks per node. 
    \item \textbf{Node2Vec.}
    We also use the DGL implementation of Node2Vec.
    The number of negative nodes in sampling is 5.
    The random walk length is fixed at 50, and we run 100 random walks per node.
    $p$ and $q$ are set to 0.25 and 4 respectively by default.
    \item \textbf{Graph Convolutional Network (GCN).}
    We use the Pytorch Geometric implementation of GCN.
    Our GCN model consists of 2 layers as suggested by the original authors. 
    For the first hidden layer, we set the hidden unit size to twice the size of input vectors.
    For the second layer, we set the hidden unit size to the embedding size.
    We use ReLU as the activation function between layers.
    Node embeddings are generated using link prediction as the objective function.
    We train the GCN  model for 400 epochs. 
\end{itemize}

\noindent For all node embedding models, we set their output embedding size (i.e., $d$) to 64, 128, and 256 for our evaluation.
These sizes are commonly used in the real world practices balancing between the expressiveness of the node embeddings and the computational complexity of the downstream tasks.
Besides, we use the largest connected components from all four datasets to accommodate these node embedding models in our evaluation.

\mypara{Competitors}
We implement three baseline methods detailed below for comparison study.

\begin{itemize}
\item \textbf{Direct Recovery.}
This baseline computes the pairwise similarity matrix from the embeddings of the original graph and reconstructs the graph by choosing the top $k \times n / 2$ 
pairs (i.e., edges) of the largest pairwise similarity scores. 
It is a straightforward attack strategy that can be leveraged by the adversaries since the embeddings of similar nodes should be close in the latent spaces (see \autoref{sec:background}). 
Note that our implementation of direct recovery is identical to the decoder used by Duddu et al.~\cite{duddu2020quantifying} to reconstruct graphs.

\item \textbf{$k$NN Graph.} 
We employ the widely used $k$NN algorithm (see \autoref{sec:background}) as the second baseline.
$k$NN builds a graph in which two nodes $v$ and $u$ are connected by an edge if the distance between $h_v$ and $h_u$ is among the $k$-th smallest distances.
We use cosine similarity as the distance function.

\item \textbf{Invert Embedding~\cite{chanpuriya2021deepwalking}.}
We adapt the optimization algorithm (Algorithm 2\&3 in Chanpuriya et al.~\cite{chanpuriya2021deepwalking}) as our third baseline to recover a graph from the node embeddings. 
Since the attackers cannot obtain the real eigenvalues from the PPMI matrix in a model agnostic setting, we thereby use a random diagonal eigenvalue matrix together with the node embedding matrix to generate the low-rank approximation matrix. 
We set the other hyperparameters as outlined in Chanpuriya et al.~\cite{chanpuriya2021deepwalking}.
Additional discussion about invert embedding can be found in \autoref{sec:related_work}.

\end{itemize}

\noindent The graph size (i.e., the number of edges) of all baselines are set to $k \times n / 2$.
We detail how we estimate $k$ in \autoref{sec:exp_estimate_avg_degree} and how $k$ influences the graph recovery performance in \autoref{sec:exp_impact_of_k}.

\mypara{Hyperparamter Configurations}
We set the number of attention heads $m$ to 16.
The temperature $\tau$, graph sparsity hyperparameters $\alpha$ and $\beta$, and the update rate $\eta$ are set to 1, 0.3,  0.1 and 0.5 respectively.
We set the maximum iteration $T$ to 400.
We use a linear graph autoencoder (i.e., $Z$ is set to 1) proposed by Salha et al.~\cite{salha2021simple}, which is an effective alternative to multilayer GCNs.
These hyperparameter values offer consistent performance across different datasets and models in our evaluation. 

\mypara{Evaluation Metrics}
Recall that the attackers have two main goals. 
Their primary goal is uncovering the edges with decent accuracy from the node embedding matrix, and their secondary goal is recovering a graph structure that is similar to the original graph with respect to the graph properties.
Bearing them in mind, we use two categories of metrics to evaluate \approach's performance.
\begin{itemize}
\item \textbf{Edge Metrics.} 
We first use four edge related metrics - precision (P), recall (R), F1 score (F1), and joint degree distribution (JDD) - to measure how \approach attains the primary goal. 
Precision, recall, and F1 are commonly used, and we apply them to measure the overall capability of \approach recovering the exact edges.
The joint degree distribution is a metric relating to the edge distribution and provides an additional measurement about 1-hop neighborhoods around a node.
It examines each pair of connected nodes and notes their respective nodal degrees. It is defined as $P(k_1, k_2) = \mu(k_1, k_2) \times m(k_1, k_2)$, where $\mu(k_1, k_2)=1$ if $k_1=k_2$ otherwise 2, and $m(k_1, k_2)$ denotes the number of edges connecting nodes of degree $k_1$ and $k_2$. 

We use SecGraph~\cite{ji2015secgraph} to calculate the Jaccard similarity among two JDDs.
For all edge metrics, values close to 1 are the best.

\item \textbf{Global Metrics.}
We then employ three global metrics - relative Frobenius error, relative triangle error, and relative average clustering coefficient error - to measure how \approach achieves its secondary goal.
The relative error is defined as the absolute error (i.e., the difference between the measured value and ground truth value) divided by the ground truth value. 
It gives an indication of how good a measurement is relative to the ground truth value, or in other words, how much the observed value deviates from actual value.
We use the relative Frobenius error, which measures the difference between the adjacency matrix $\mathbf{A}_O$ and $\mathbf{A}_R$, i.e., $\| \mathbf{A}_O - \mathbf{A}_R \|_F / \| \mathbf{A}_O \|_F$. 
Similarly, we count the absolute difference between the number of triangles (respectively average clustering coefficient) of $\mathbf{G}_R$ and that of $\mathbf{G}_O$, then divided by the number of triangles (respectively average clustering coefficient) of $\mathbf{G}_O$ to calculate the relative triangle error (the relative average clustering coefficient error).
For all global metrics, values close to 0 are the best.

Similar relative error metrics are also used in Chanpuriya et al.~\cite{chanpuriya2021deepwalking}.
\end{itemize}

\noindent In practice, the audience could potentially leverage Narayanan-Shmatikov's attack~\cite{narayanan2009anonymizing} (and other appropriate de-anonymization attacks) to measure, to what extent, the recovered graph can assist the graph de-anonymization task given different graphs and different levels of background knowledge.

\mypara{Runtime Configuration}
All the experiments in this paper are repeated 5 times.
For each run, we follow the same experimental setup laid out before. 
We report the mean and standard deviation of each metric to evaluate the attack performance.
In this way, we can delineate objective performance results without reporting opportunistically optimal results.

\subsection{How to Estimate the Average Node Degree?}
\label{sec:exp_estimate_avg_degree}

\begin{figure}[t]
     \centering
    \begin{subfigure}[t]{0.495\linewidth}
        \includegraphics[width=\linewidth]{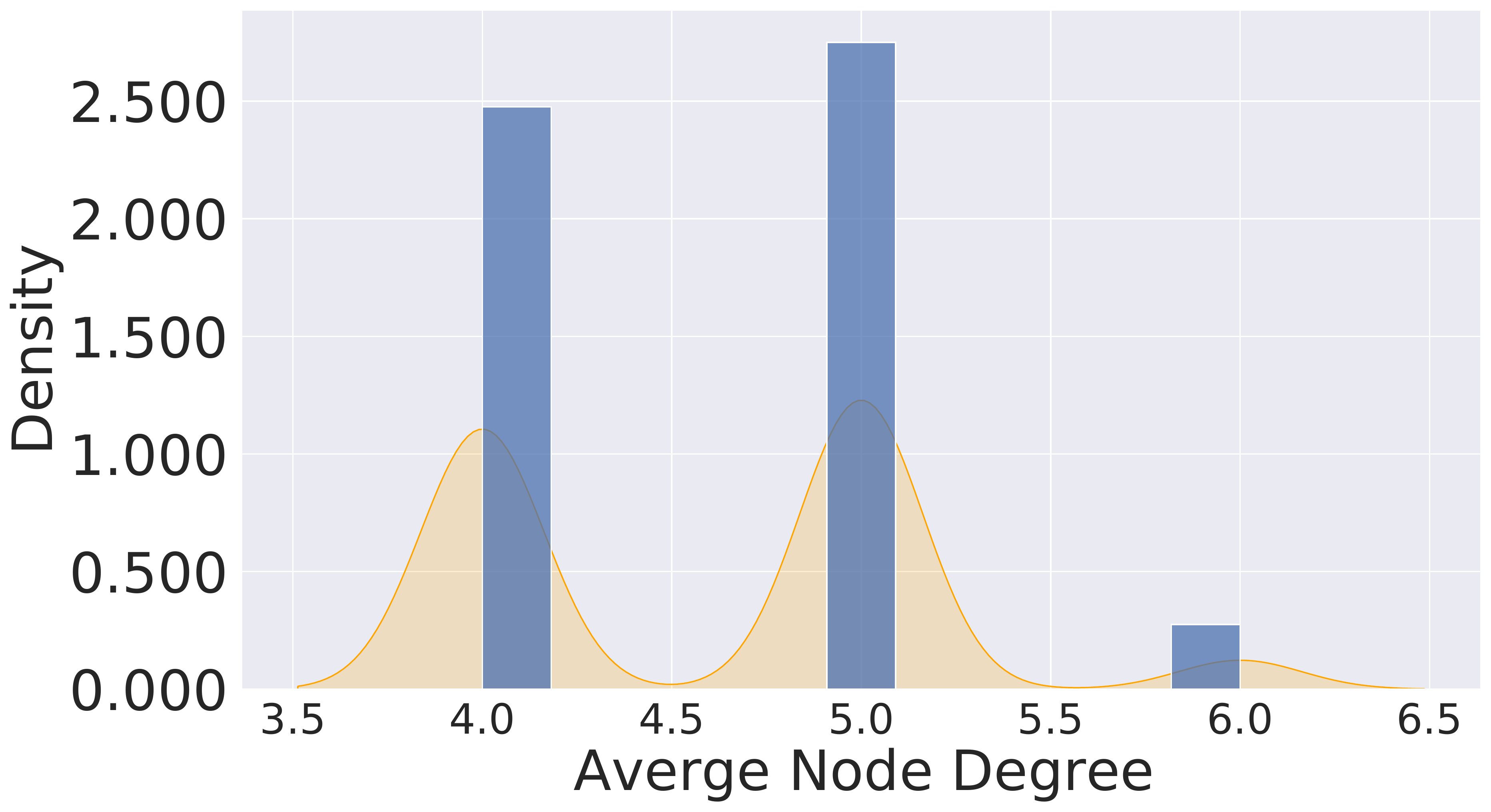}
        \caption{citation/co-occurrence}
        \label{fig:citation_network_estimated}
    \end{subfigure}
    \hfill
    \begin{subfigure}[t]{0.495\linewidth}
        \includegraphics[width=\linewidth]{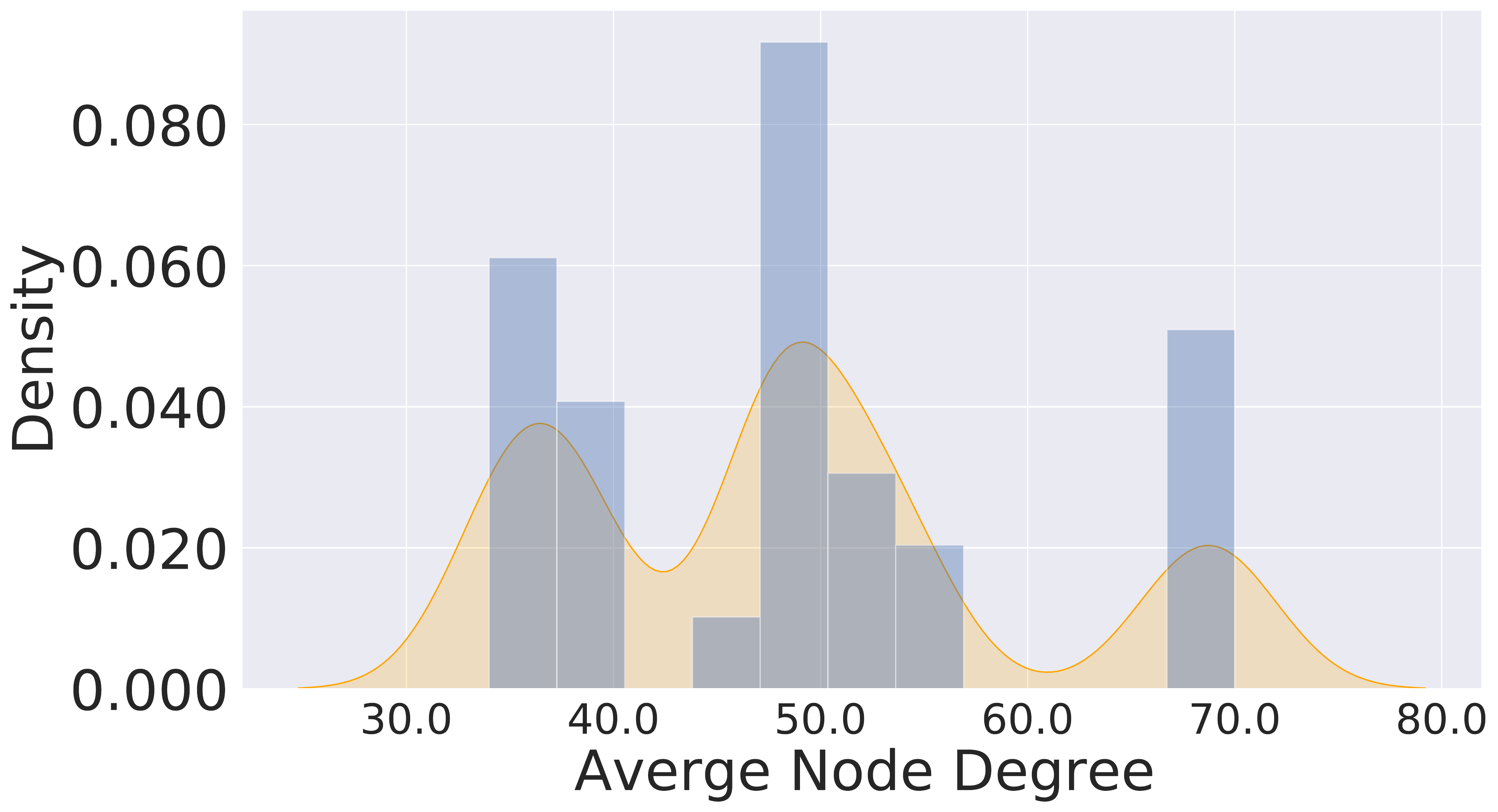}
        \caption{social}
        \label{fig:social_network_estimated}
    \end{subfigure}
    \hfill
    \caption{Distribution of estimated average node degrees. 
    The mean and standard deviation of our estimated average node degree of the citation/co-occurrence graphs (\autoref{fig:citation_network_estimated}) are 4.6 and 0.8. 
    The respective values of social networks (\autoref{fig:social_network_estimated}) are 45.7 and 8.4.}
\label{fig:estimated_k}
\end{figure}

\begin{figure}[t]
\centering
\begin{subfigure}[t]{0.495\linewidth}
\includegraphics[width=\textwidth]{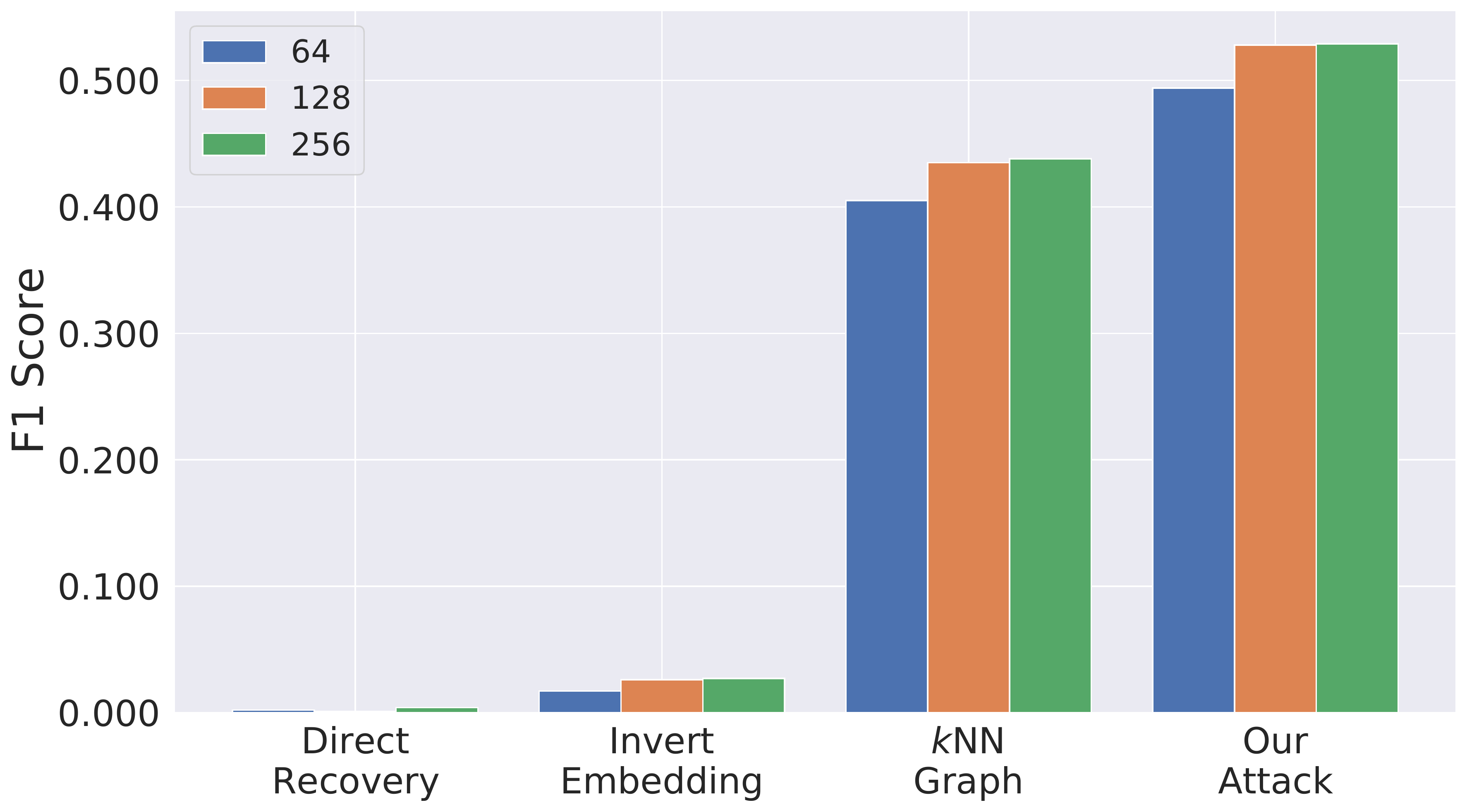}
\caption{F1 Score} 
\label{fig:acm_GAT_batch_hyper}
\end{subfigure}
\hfill
\centering
\begin{subfigure}[t]{0.495\linewidth}
\includegraphics[width=\textwidth]{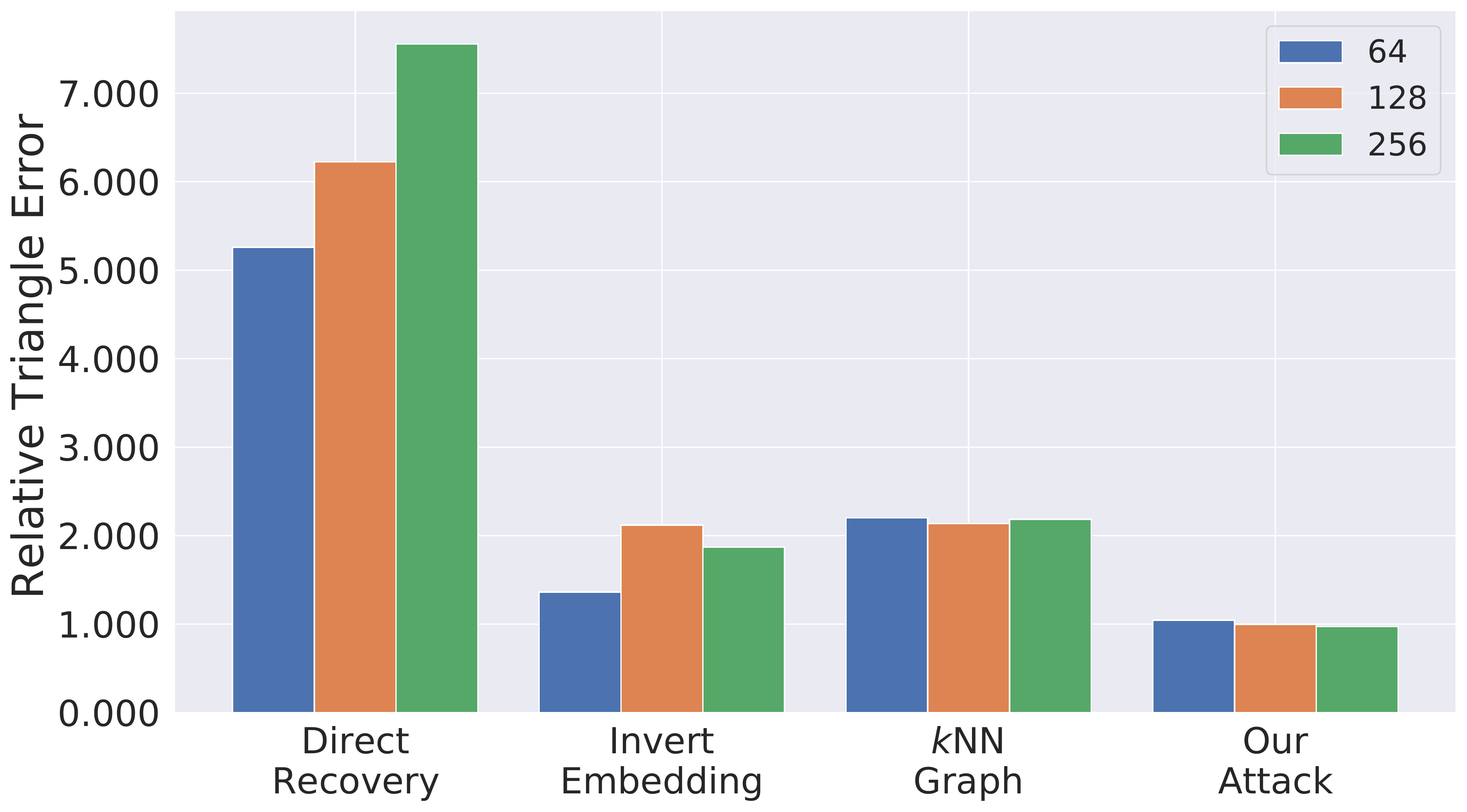}
\caption{Relative Triangle Error
} 
\label{fig:acm_GIN_batch_hyper}
\end{subfigure}
\hfill
\caption{F1 scores and relative triangle error scores of all basesline methods and \approach when given different node embedding sizes (i.e., 64, 128 and 256). We use Node2Vec to generate node embedding matrices.}
\label{fig:comparison_study}
\end{figure}

Recall that the only clue that the attackers have is the background information about the origin of the node embedding matrix. 
In this section, we exemplify how the attackers can estimate the average node degree $k$ from the graphs of similar origins by leveraging state-of-the-art graph sampling methods.
Note that the attackers can estimate the graph size (i.e., the number of edges) which equals to $k \times n / 2$.
We use the state-of-the-art spikyball sampling~\cite{ricaud2020spikyball} to estimate the average node degree for our evaluation.
It generalizes several exploration-based sampling schemes (e.g., Snowball sampling, Forest Fire sampling, graph-expander sampling etc.), and can be applied to any large graphs due to its flexibility~\cite{ricaud2020spikyball}.

Specifically, for citation/co-occurrence graphs (e.g., Cora, Citeseer, and Actor), we use the publicly available citation graphs - Pubmed and DBLP - to estimate the average node degree. 
For each graph, we use spikyball sampling to sample 30\% of the whole graph then estimate the average node degree from the sampled graph. 
This process is repeated 300 times.  
We calculate the mean average node degree as our final estimation of citation/co-occurrence graphs.
For social network graphs (e.g., Facebook), we randomly select six graphs (e.g., socfb-BU10, socfb-Carnegie49, socfb-JMU79, socfb-Lehigh96, socfb-Maine59 and socfb-UCSC68) from the publicly available FB100 dataset~\cite{red2011comparing} plus one Twitter graph~\cite{mcauley2012learning} from SNAP.
We also sample 30\% of each graph to estimate the average node degree and repeat this process 300 times per graph.
This strategy enables the adversary to sample enough graphs to cover a wide spectrum of graph properties.

The estimation results are shown in \autoref{fig:estimated_k}.
What can be seen in \autoref{fig:estimated_k} is that the estimated average node degrees may not exactly match the real values but are roughly within the same order of magnitude.
For instance, the mean and standard deviation of our estimated average node degree of the citation/co-occurrence graphs (\autoref{fig:citation_network_estimated}) are 4.6 and 0.8, while the mean and standard deviation of our estimated average node degree of social networks are 45.7 and 8.4 respectively.
Comparing to the real values in \autoref{tab:graph_stats}, the estimated values are not precise.
For instance, the above graph sampling process overestimates the average node degree for the Cora and Citeseer datasets, while underestimating the average node degree of the Actor dataset.
However, they can offer the attackers a reasonable starting point to estimate the graph sizes.
We use these estimated values (i.e., 5 for citation/co-occurrence graphs and 46 for social network graphs) in the rest of our evaluation. 
We provide a detailed study on how \approach can attain good performance even when the estimated average node degree is almost twice the ground truth value in \autoref{sec:exp_impact_of_k}.

\mypara{Takeaways}
When only having the background information about the origin of the node embedding matrix, our sampling process represents a feasible way that the attackers take to estimate the average node degree.
The estimated average node degrees are in the vicinity of the ground truth values but not exactly matching them.

\subsection{Is \approach Better than the Baselines?}
\label{sec:exp_comparison}

\begin{table*}[t]
\centering
\caption{Comparison of all baseline methods and \approach. We use the Cora dataset and the node embedding size is 256.}
\resizebox{0.7\linewidth}{!} {
\begin{tabular}{ccccccccc}
\toprule
\multirow{2}{*}{\textbf{\begin{tabular}[c]{@{}c@{}}Graph\\ Reocvery\\ Method\end{tabular}}} &
  \multirow{2}{*}{$f$} &
  \multicolumn{4}{c}{\textbf{Edge Metric}} &
  \multicolumn{3}{c}{\textbf{Global Metric}} \\
\cmidrule(lr){3-6} \cmidrule(lr){7-9}
 &
   &
  \multicolumn{1}{c}{\textbf{Precision}} &
  \multicolumn{1}{c}{\textbf{Recall}} &
  \multicolumn{1}{c}{\textbf{F1}} &
  \multicolumn{1}{c}{\textbf{JDD}} &
  \multicolumn{1}{c}{\textbf{\begin{tabular}[c]{@{}c@{}}Frobenius\\ Error\end{tabular}}} &
  \multicolumn{1}{c}{\textbf{\begin{tabular}[c]{@{}c@{}}Triangle\\ Error\end{tabular}}} &
  \multicolumn{1}{c}{\textbf{\begin{tabular}[c]{@{}c@{}}Clustering Coef.\\ Error\end{tabular}}} \\
\midrule
\multirow{4}{*}{\textbf{\begin{tabular}[c]{@{}c@{}}Direct\\ Recovery\end{tabular}}}  & DW & 0.001$\pm$0.000 & 0.002$\pm$0.000 & 0.001$\pm$0.000 & 0.000$\pm$0.000 & 1.647$\pm$0.000 & 6.867$\pm$0.000 & 0.753$\pm$0.000 \\
         & N2V & 0.003$\pm$0.000 & 0.006$\pm$0.000 & 0.004$\pm$0.000 & 0.000$\pm$0.000 & 1.645$\pm$0.000 & 7.559$\pm$0.000 & 0.759$\pm$0.000 \\
         & NetSMF & 0.013$\pm$0.000 & 0.022$\pm$0.000 & 0.016$\pm$0.000 & 0.311$\pm$0.000 & 1.647$\pm$0.000 & 0.621$\pm$0.000 & 0.228$\pm$0.000 \\
         & GCN & 0.001$\pm$0.000 & 0.002$\pm$0.000 & 0.002$\pm$0.000 & 0.000$\pm$0.000 & 1.647$\pm$0.000 & 6.391$\pm$0.000 & 0.653$\pm$0.000 \\
\midrule
\multirow{4}{*}{\textbf{\begin{tabular}[c]{@{}c@{}}Invert\\ Embedding\end{tabular}}} & DW & 0.007$\pm$0.005 & 0.012$\pm$0.010 & 0.009$\pm$0.007 & 0.667$\pm$0.092 & 1.660$\pm$0.010 & 0.866$\pm$0.661 & 0.198$\pm$0.008 \\
         & N2V & 0.021$\pm$0.008 & 0.037$\pm$0.014 & 0.027$\pm$0.010 & 0.665$\pm$0.030 & 1.645$\pm$0.008 & 1.869$\pm$0.160 & 0.148$\pm$0.005 \\
         & NetSMF & 0.003$\pm$0.001 & 0.005$\pm$0.003 & 0.004$\pm$0.002 & 0.462$\pm$0.064 & 1.676$\pm$0.007 & 0.842$\pm$0.789 & 0.221$\pm$0.011 \\
         & GCN & 0.015$\pm$0.003 & 0.026$\pm$0.006 & 0.019$\pm$0.004 & 0.675$\pm$0.004 & 1.657$\pm$0.005 & 0.255$\pm$0.158 & 0.188$\pm$0.005 \\
\midrule
\multirow{4}{*}{\textbf{\begin{tabular}[c]{@{}c@{}}$k$NN\\ Graph\end{tabular}}} & DW & 0.401$\pm$0.000 & 0.492$\pm$0.000 & 0.442$\pm$0.000 & 0.340$\pm$0.000 & 1.114$\pm$0.000 & 3.029$\pm$0.000 & 0.286$\pm$0.000 \\
         & N2V & 0.397$\pm$0.000 & 0.487$\pm$0.000 & 0.438$\pm$0.000 & 0.338$\pm$0.000 & 1.119$\pm$0.000 & 2.185$\pm$0.000 & 0.276$\pm$0.000 \\
         & NetSMF & 0.469$\pm$0.000 & 0.575$\pm$0.000 & 0.517$\pm$0.000 & 0.334$\pm$0.000 & 1.037$\pm$0.000 & 2.379$\pm$0.000 & 0.325$\pm$0.000 \\
         & GCN & 0.378$\pm$0.000 & 0.463$\pm$0.000 & 0.416$\pm$0.000 & 0.333$\pm$0.000 & 1.140$\pm$0.000 & 2.172$\pm$0.000 & 0.286$\pm$0.000 \\
\midrule
\multirow{4}{*}{\approach}  & DW & 0.492$\pm$0.004 & 0.578$\pm$0.003 & 0.531$\pm$0.003 & 0.840$\pm$0.011 & 1.010$\pm$0.005 & 1.118$\pm$0.036 & 0.215$\pm$0.006 \\
         & N2V & 0.506$\pm$0.001 & 0.554$\pm$0.003 & 0.529$\pm$0.001 & 0.724$\pm$0.007 & 0.993$\pm$0.001 & 0.973$\pm$0.027 & 0.228$\pm$0.004 \\
         & NetSMF & 0.579$\pm$0.002 & 0.640$\pm$0.003 & 0.608$\pm$0.003 & 0.732$\pm$0.006 & 0.908$\pm$0.003 & 1.263$\pm$0.019 & 0.288$\pm$0.004 \\
         & GCN & 0.462$\pm$0.002 & 0.506$\pm$0.001 & 0.483$\pm$0.001 & 0.753$\pm$0.006 & 1.040$\pm$0.003 & 0.864$\pm$0.032 & 0.230$\pm$0.005 \\
\bottomrule
\end{tabular}
}
\label{tab:exp_comparison_study_cora_256}
\end{table*}

In this section, we aim at studying whether \approach is effective to recover a graph from the node embedding matrix, or whether the existing baseline methods would be enough for the task at hand.
To address this research question, we compare \approach to the baseline methods discussed in \autoref{sec:exp_setup}.
All methods use the estimated average node degrees outlined in \autoref{sec:exp_estimate_avg_degree}. 
The node embedding size is fixed to 256.
Due to space limitations, we only report the attack results on the Cora dataset.  
The results of the other three datasets can be found in \autoref{sec:appendix_comparison}.

\mypara{Performance}
The performance comparison results are shown in \autoref{tab:exp_comparison_study_cora_256}.
Overall, direct graph recovery and invert embedding graph recovery cannot recover graphs from the node embedding matrices given all four node embedding models.
For instance, the F1 scores of these two methods are no greater than 0.027, indicating that they cannot attain the attacker's primary goal.
At the same time, the global metrics of these two baselines are equally underwhelming.
Our results show that such optimization based approach is less effective in a model agnostic setting.
$k$NN algorithm represents a \emph{de facto} approach to recover the edges from node embeddings.
Our results show that $k$NN graph can partially recover the edges from the node embedding matrix.
For example, it can recover the edges from the node embeddings generated by Node2Vec with a 0.438 F1 score.
As we can see in \autoref{tab:exp_comparison_study_cora_256}, \approach outperforms all baseline methods.
Take the node embeddings generated by Node2Vec for example, \approach achieves 0.529 F1 score, which is 0.091 higher than that of $k$NN graph recovery.
In other words, \approach's F1 score relatively improves that of $k$NN graph recovery by 0.208 (i.e., 0.091/0.438=0.208).
If we take the edges recovered by $k$NN graph as the upper bound of the existing privacy risk assessment, \approach empirically improves this upper bound by 0.208 per our evaluation results.
Given the combinatorial nature of graph edges (i.e., n(n-1)/2 possibilities) and our strict attack setting (i.e., no interaction with the node embedding models), such 0.208 relative improvement by \approach is substantial. 
Practically speaking, if we position \approach in the privacy risk assessment framework, it would lead to 0.208 increase of the estimated privacy loss than the de facto risk assessment using $k$NN algorithm. 
We also provide a visual explanation to exemplify \approach's capability in recovery graphs from the node embedding matrices in \autoref{sec:appendix_comparison}.

\mypara{Impact of Node Embedding Size}
We use two metrics - F1 score and relative triangle error - to understand the impact of node embedding size on both baselines and \approach.
We use Node2Vec to generate node embedding matrices.
The results are shown in \autoref{fig:comparison_study}.
It is straightforward to see that \approach consistently performs better than the baselines given different node embedding sizes.

\mypara{Takeaways}
The proposed learnable distance function and adaptive graph structure combination can reduce a reasonable amount of false edges.
They, in turn, enable \approach to recover better graph structure from the node embedding matrix given different node embedding models and embedding sizes.
Besides, $k$NN graph remains a viable approach to recover edges from the node embedding matrix.
However, due to its non-learning-based nature, $k$NN graph is outperformed by \approach.

\subsection{How Effective is \approach?}

\begin{table*}[t]
\centering
\caption{
The performance results of \approach using all four datasets. 
We fix the node embedding size to 256. 
We show the relative improvement scores in edge metrics to demonstrate to what extent \approach can relatively improve from $k$NN graph.
We add a positive sign (+) next to the relative improvement score to highlight the improvement.
We also show the relative error reduction scores in global metrics to demonstrate to what extent \approach can relatively reduce errors incurred by $k$NN graph.
We add a negative sign (-) next to the relative error reduction score to highlight the difference.
}
\resizebox{0.95\linewidth}{!} {
\begin{tabular}{ccccccccc}
\toprule
\multirow{2}{*}{\textbf{Dataset}} &
  \multirow{2}{*}{$f$} &
  \multicolumn{4}{c}{\textbf{Edge Metrics}} &
  \multicolumn{3}{c}{\textbf{Global Metrics}} \\
\cmidrule(lr){3-6} \cmidrule(lr){7-9}
 &
   &
  \multicolumn{1}{c}{\textbf{Precision}} &
  \multicolumn{1}{c}{\textbf{Recall}} &
  \multicolumn{1}{c}{\textbf{F1}} &
  \multicolumn{1}{c}{\textbf{JDD}} &
  \multicolumn{1}{c}{\textbf{\begin{tabular}[c]{@{}c@{}}Frobenius\\ Error\end{tabular}}} &
  \multicolumn{1}{c}{\textbf{\begin{tabular}[c]{@{}c@{}}Triangle\\ Error\end{tabular}}} &
  \multicolumn{1}{c}{\textbf{\begin{tabular}[c]{@{}c@{}}Clustering Coef.\\ Error\end{tabular}}} \\
\midrule
\multirow{4}{*}{\textbf{Cora}} & DW & 0.492$\pm$0.004 (+0.226) & 0.578$\pm$0.003 (+0.174) & 0.531$\pm$0.003 (+0.201) & 0.840$\pm$0.011 (+1.471) & 1.010$\pm$0.005 (-0.104) & 1.118$\pm$0.036 (-1.911) & 0.215$\pm$0.006 (-0.071) \\
         & N2V & 0.506$\pm$0.001 (+0.276) & 0.554$\pm$0.003 (+0.137) & 0.529$\pm$0.001 (+0.208) & 0.724$\pm$0.007 (+1.143) & 0.993$\pm$0.001 (-0.126) & 0.973$\pm$0.027 (-1.212) & 0.228$\pm$0.004 (-0.048) \\
         & NetSMF & 0.579$\pm$0.002 (+0.235) & 0.640$\pm$0.003 (+0.113) & 0.608$\pm$0.003 (+0.176) & 0.732$\pm$0.006 (+1.191) & 0.908$\pm$0.003 (-0.129) & 1.263$\pm$0.019 (-1.116) & 0.288$\pm$0.004 (-0.037) \\
         & GCN & 0.462$\pm$0.002 (+0.223) & 0.506$\pm$0.001 (+0.092) & 0.483$\pm$0.001 (+0.162) & 0.753$\pm$0.006 (+1.260) & 1.040$\pm$0.003 (-0.100) & 0.864$\pm$0.032 (-1.308) & 0.230$\pm$0.005 (-0.056) \\
\midrule
\multirow{4}{*}{\textbf{Citeseer}} & DW & 0.403$\pm$0.002 (+0.193) & 0.555$\pm$0.005 (+0.149) & 0.467$\pm$0.003 (+0.174) & 0.617$\pm$0.011 (+1.635) & 1.125$\pm$0.003 (-0.085) & 1.877$\pm$0.075 (-2.651) & 0.341$\pm$0.009 (-0.080) \\
         & N2V & 0.445$\pm$0.001 (+0.271) & 0.575$\pm$0.002 (+0.149) & 0.502$\pm$0.001 (+0.217) & 0.506$\pm$0.007 (+1.137) & 1.069$\pm$0.002 (-0.127) & 1.734$\pm$0.039 (-1.665) & 0.357$\pm$0.005 (-0.059) \\
         & NetSMF & 0.530$\pm$0.002 (+0.229) & 0.672$\pm$0.001 (+0.091) & 0.592$\pm$0.001 (+0.168) & 0.461$\pm$0.005 (+1.048) & 0.961$\pm$0.002 (-0.133) & 2.001$\pm$0.056 (-1.419) & 0.432$\pm$0.003 (-0.056) \\
         & GCN & 0.414$\pm$0.003 (+0.206) & 0.529$\pm$0.002 (+0.080) & 0.465$\pm$0.001 (+0.153) & 0.527$\pm$0.011 (+1.344) & 1.105$\pm$0.004 (-0.099) & 1.467$\pm$0.057 (-1.734) & 0.330$\pm$0.004 (-0.067) \\
\midrule
\multirow{4}{*}{\textbf{Actor}} & DW & 0.687$\pm$0.001 (+0.222) & 0.435$\pm$0.002 (+0.088) & 0.533$\pm$0.002 (+0.138) & 0.417$\pm$0.001 (+0.587) & 0.874$\pm$0.001 (-0.081) & 0.203$\pm$0.009 (-0.105) & 0.229$\pm$0.002 (-0.017) \\
         & N2V & 0.465$\pm$0.001 (+0.356) & 0.313$\pm$0.000 (+0.282) & 0.374$\pm$0.000 (+0.312) & 0.473$\pm$0.003 (+0.293) & 1.023$\pm$0.001 (-0.083) & 0.179$\pm$0.007 (-0.387) & 0.176$\pm$0.001 (-0.035) \\
         & NetSMF & 0.562$\pm$0.002 (+0.240) & 0.366$\pm$0.001 (+0.136) & 0.443$\pm$0.001 (+0.179) & 0.457$\pm$0.003 (+0.406) & 0.959$\pm$0.001 (-0.074) & 0.147$\pm$0.013 (-0.758) & 0.285$\pm$0.002 (-0.025) \\
         & GCN & 0.373$\pm$0.001 (+0.226) & 0.263$\pm$0.000 (+0.211) & 0.308$\pm$0.001 (+0.218) & 0.505$\pm$0.003 (+0.446) & 1.086$\pm$0.001 (-0.045) & 0.280$\pm$0.008 (-0.349) & 0.153$\pm$0.002 (-0.049) \\
\midrule
\multirow{4}{*}{\textbf{Facebook}} & DW & 0.441$\pm$0.001 (+0.028) & 0.471$\pm$0.001 (+0.066) & 0.456$\pm$0.001 (+0.046) & 0.519$\pm$0.006 (+1.745) & 1.061$\pm$0.001 (-0.009) & 0.494$\pm$0.002 (+0.213) & 0.077$\pm$0.001 (+0.006) \\
         & N2V & 0.468$\pm$0.000 (+0.018) & 0.487$\pm$0.001 (+0.026) & 0.477$\pm$0.001 (+0.022) & 0.444$\pm$0.002 (+1.581) & 1.033$\pm$0.001 (-0.007) & 0.545$\pm$0.001 (+0.499) & 0.090$\pm$0.001 (+0.050) \\
         & NetSMF & 0.454$\pm$0.001 (+0.022) & 0.502$\pm$0.002 (+0.098) & 0.476$\pm$0.001 (+0.059) & 0.457$\pm$0.002 (+0.570) & 1.050$\pm$0.001 (-0.006) & 0.424$\pm$0.007 (+0.418) & 0.081$\pm$0.001 (+0.041) \\
         & GCN & 0.342$\pm$0.001 (+0.061) & 0.364$\pm$0.001 (+0.100) & 0.352$\pm$0.001 (+0.078) & 0.371$\pm$0.004 (+1.026) & 1.157$\pm$0.001 (-0.012) & 0.452$\pm$0.002 (+0.380) & 0.056$\pm$0.001 (-0.031) \\
\bottomrule\\
\end{tabular}
}
\label{tab:attack_256_results}
\end{table*}

\begin{figure*}[t]
     \centering
    \begin{subfigure}[t]{0.23\textwidth}
    \centering
        \includegraphics[width=\linewidth]{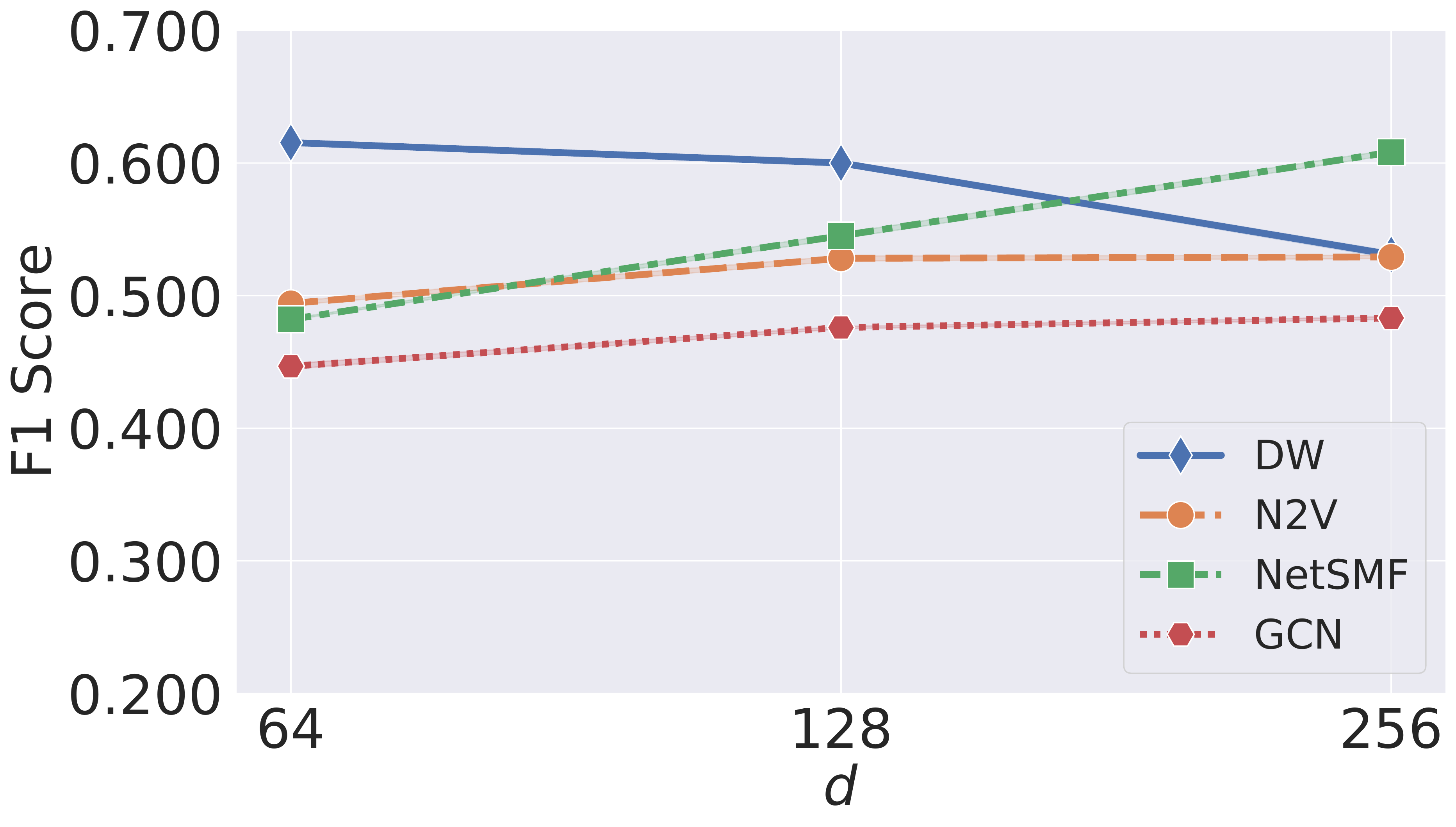}
        \caption{Cora}
        \label{fig:cora_all_dims_f1}
    \end{subfigure}
    \hfill
    \begin{subfigure}[t]{0.23\textwidth}
    \centering
        \includegraphics[width=\linewidth]{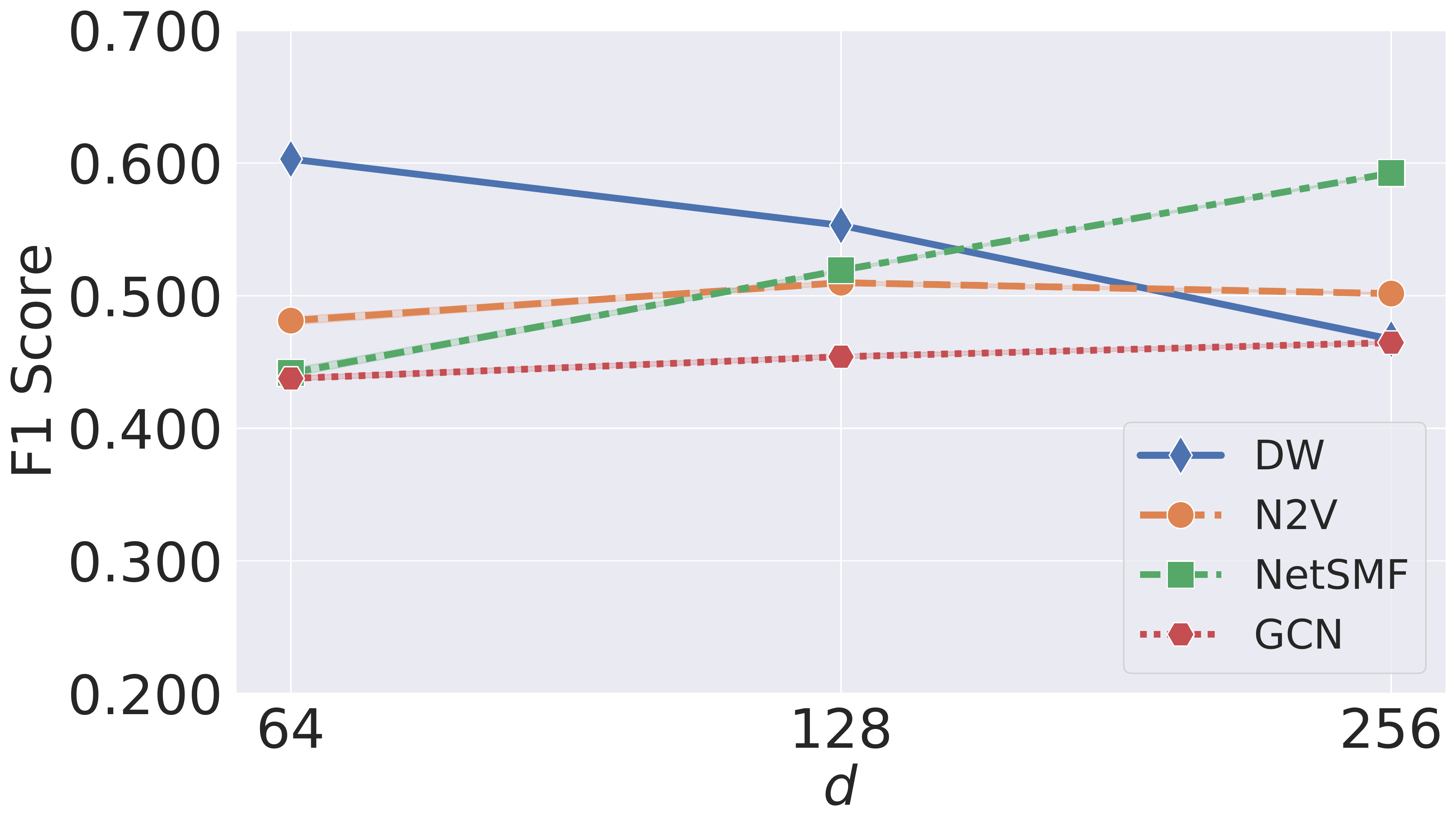}
        \caption{Citeseer}
        \label{fig:citeseer_all_dims_f1}
    \end{subfigure}
    \hfill
    \begin{subfigure}[t]{0.23\textwidth}
    \centering
        \includegraphics[width=\linewidth]{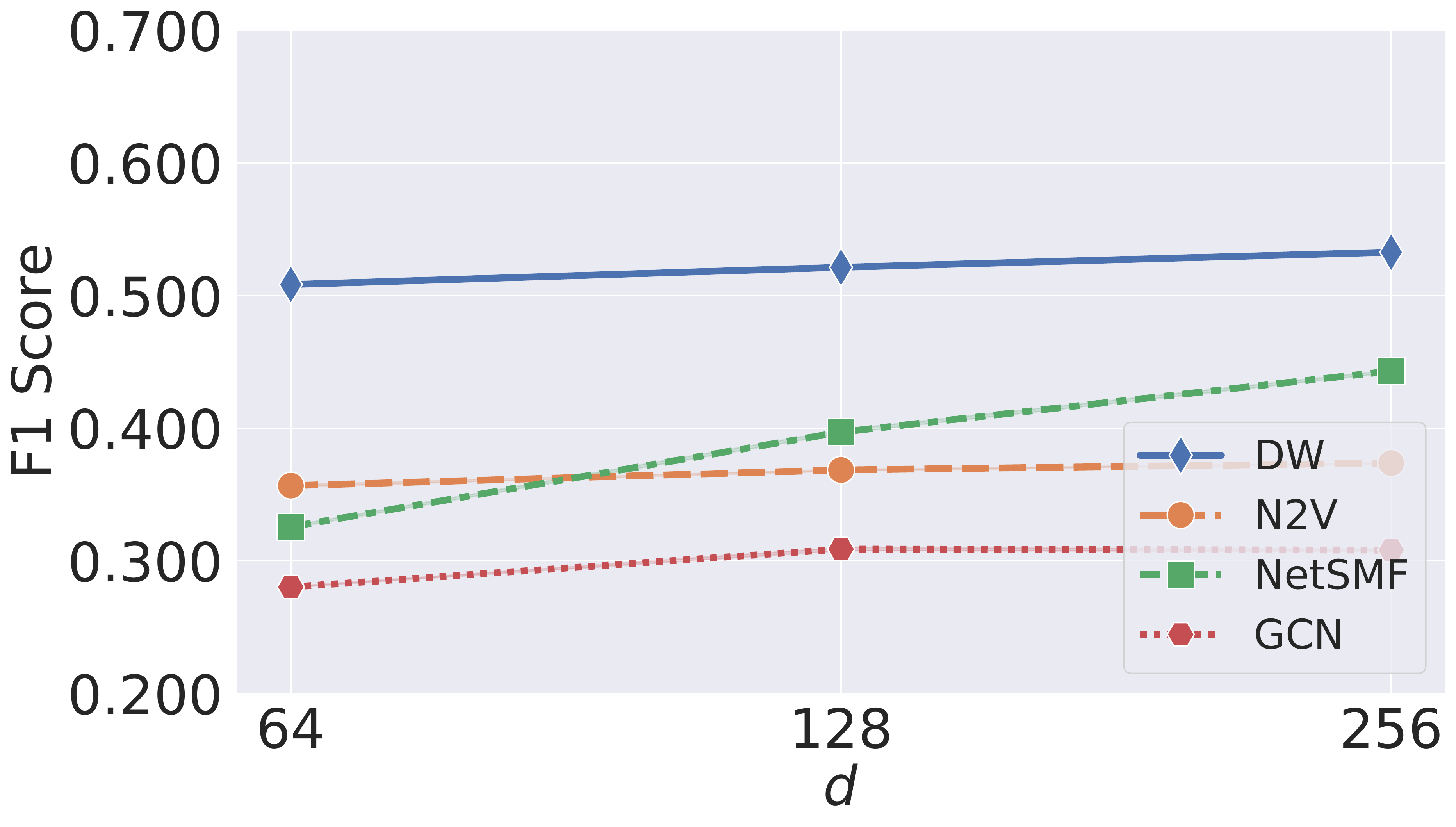}
        \caption{Actor}
        \label{fig:actor_all_dims_f1}
    \end{subfigure}
    \hfill
    \begin{subfigure}[t]{0.23\textwidth}
    \centering
        \includegraphics[width=\linewidth]{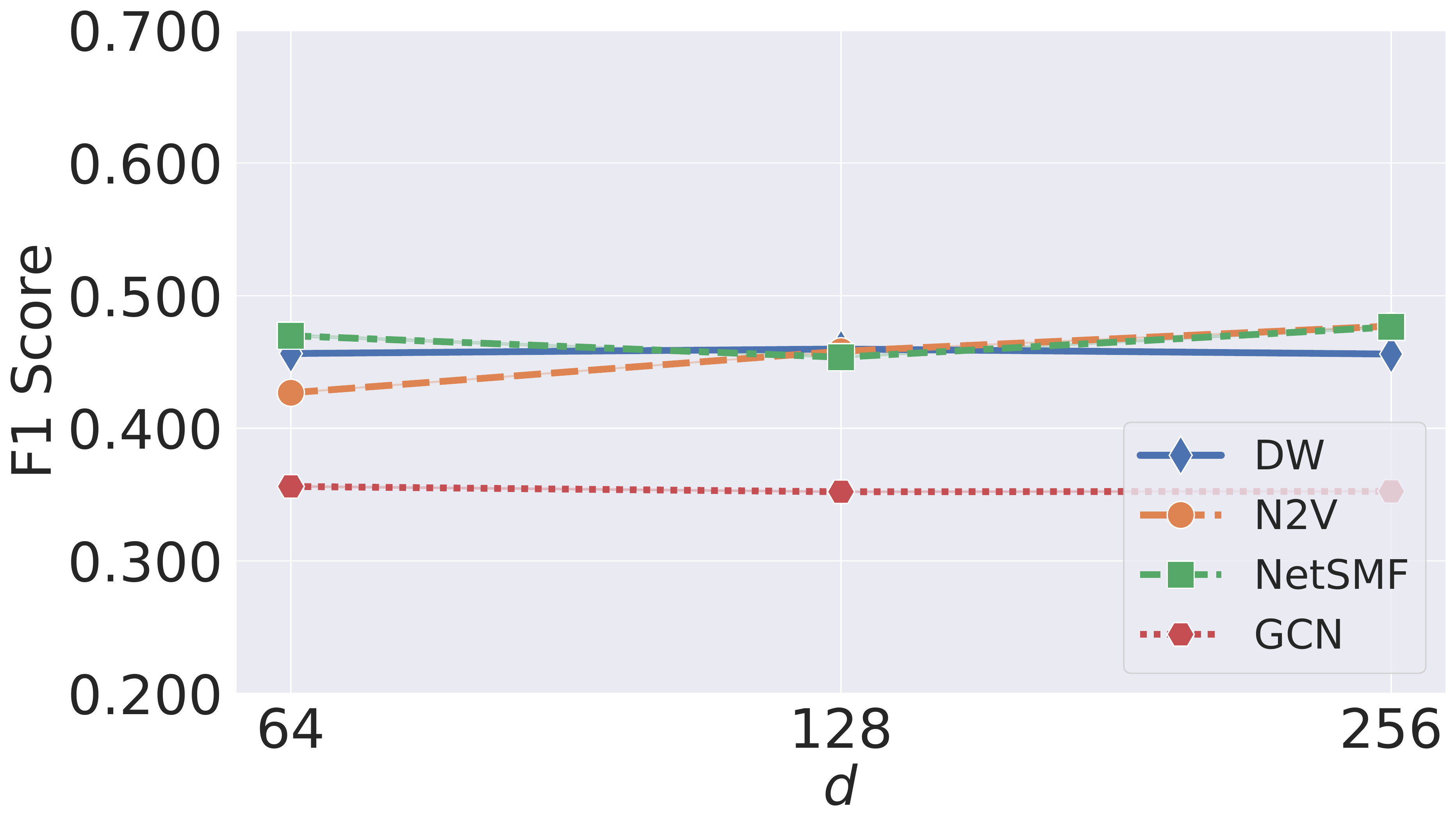}
        \caption{Facebook}
        \label{fig:facebook_all_dims_f1}
    \end{subfigure}
    \hfill
    \begin{subfigure}[t]{0.23\textwidth}
    \centering
        \includegraphics[width=\linewidth]{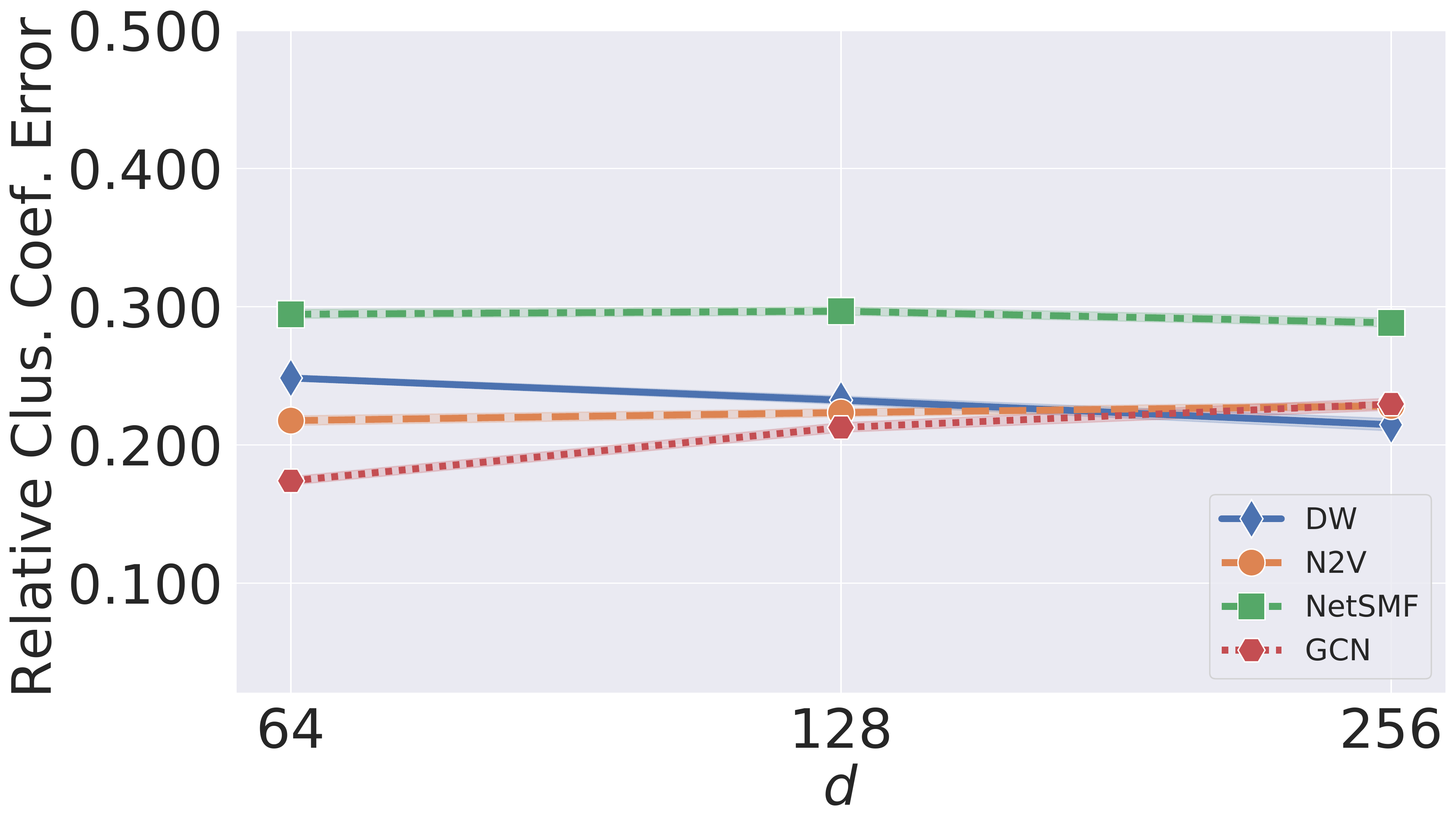}
        \caption{Cora}
        \label{fig:cora_all_dims_clus}
    \end{subfigure}
    \hfill
    \begin{subfigure}[t]{0.23\textwidth}
    \centering
        \includegraphics[width=\linewidth]{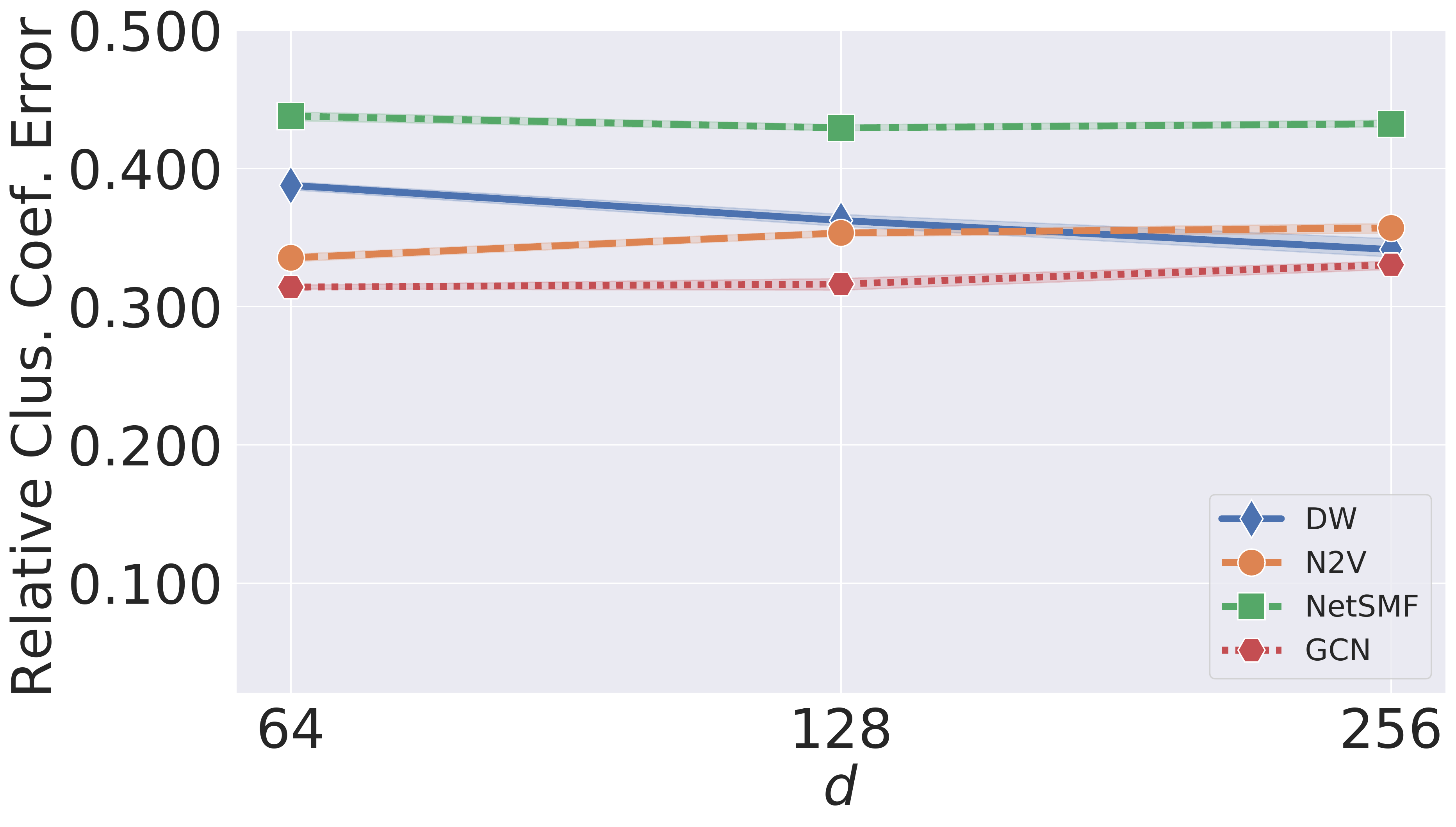}
        \caption{Citeseer}
        \label{fig:citeseer_all_dims_clus}
    \end{subfigure}
    \hfill
    \begin{subfigure}[t]{0.23\textwidth}
    \centering
        \includegraphics[width=\linewidth]{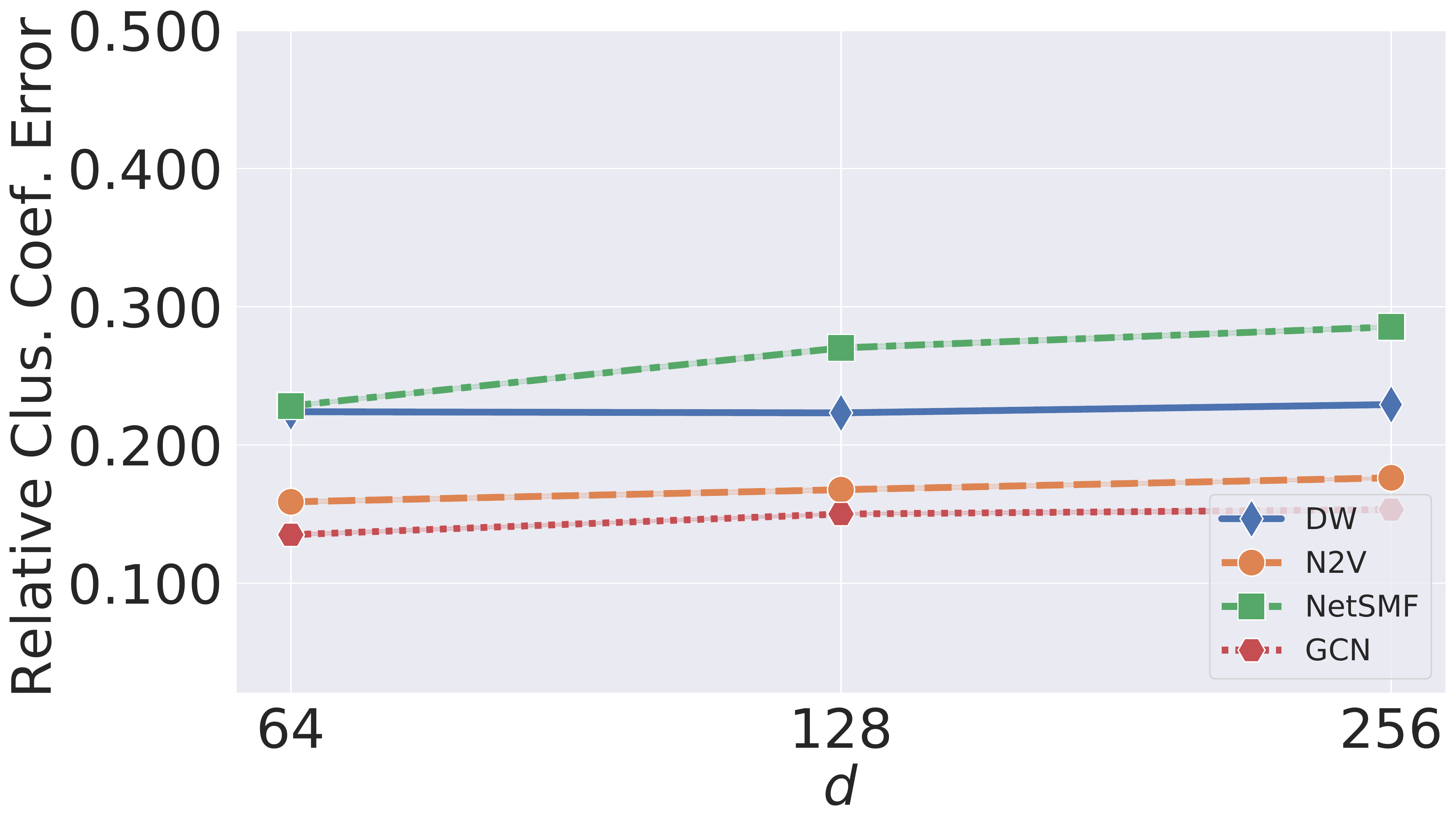}
        \caption{Actor}
        \label{fig:actor_all_dims_clus}
    \end{subfigure}
    \hfill
    \begin{subfigure}[t]{0.23\textwidth}
    \centering
        \includegraphics[width=\linewidth]{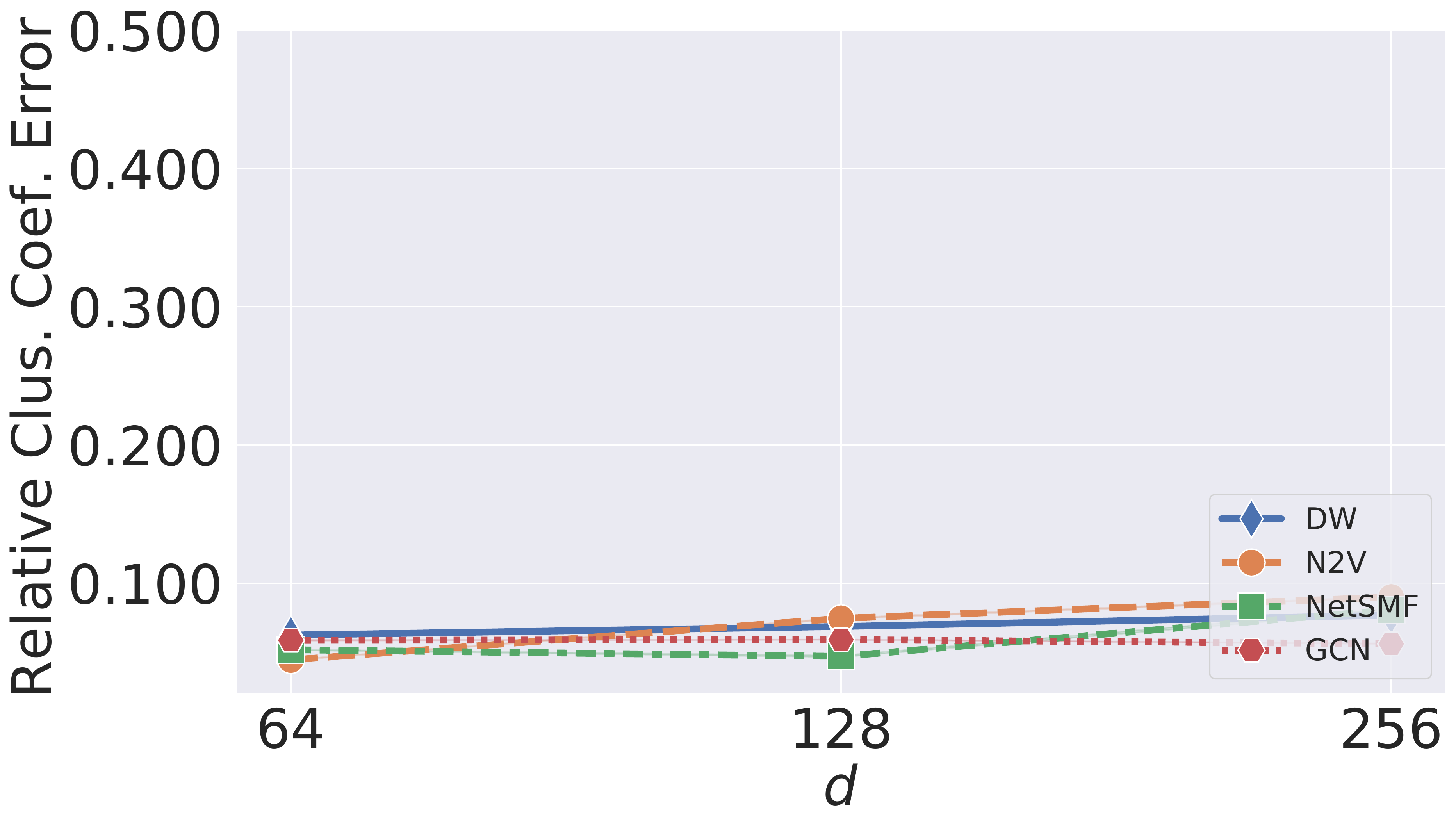}
        \caption{Facebook}
        \label{fig:facebook_all_dims_clus}
    \end{subfigure}
    \hfill
    \caption{F1 scores and relative average clustering coefficient error scores of \approach given all four datasets. We fix the node embedding size to 256. 
    }
    \label{fig:attack_all_datasets_all_dims}
\vspace{-1em}
\end{figure*}

\begin{figure*}[t]

    \begin{subfigure}[b]{0.23\textwidth}

        \includegraphics[width=\linewidth]{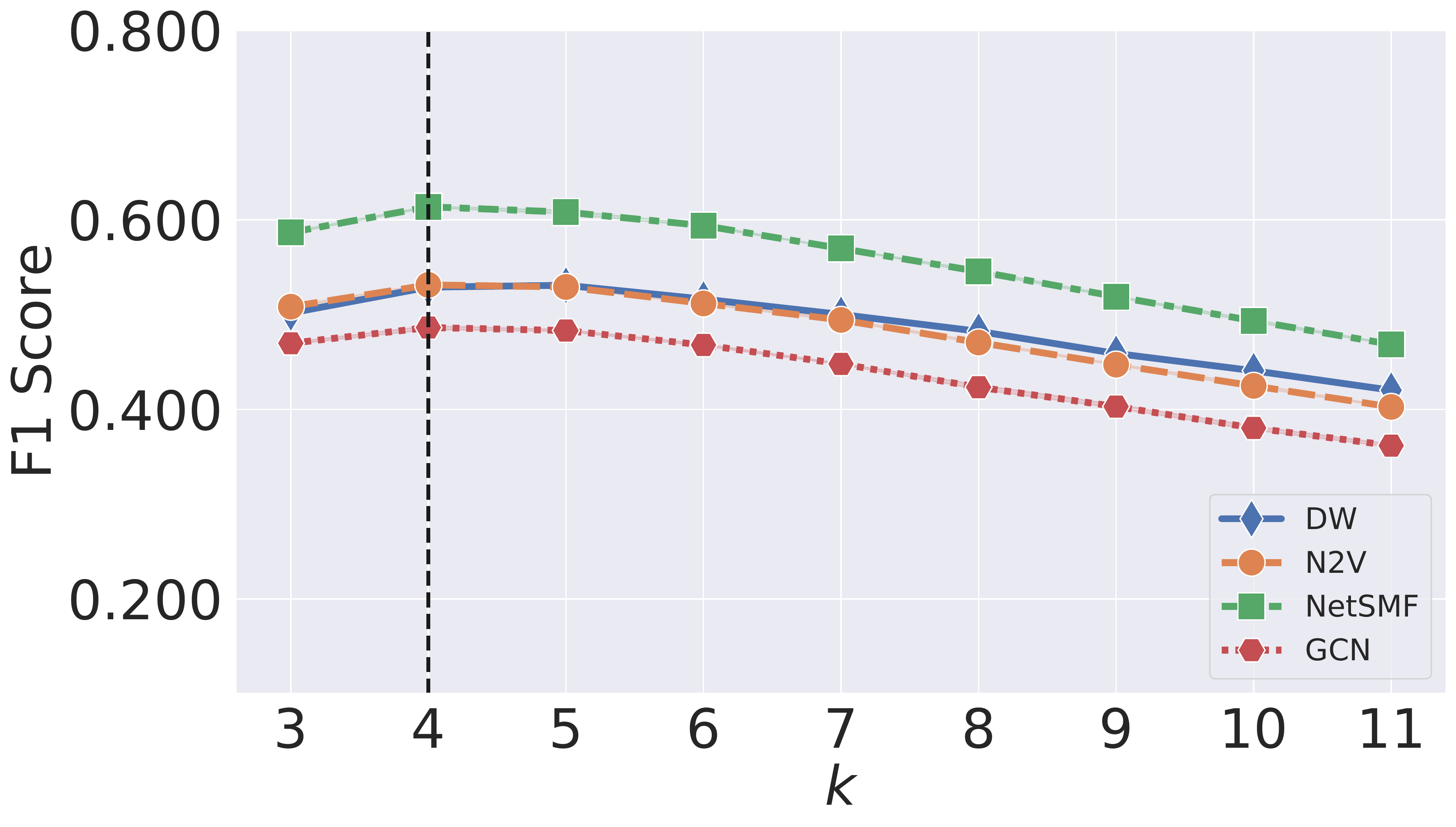}
        \caption{Cora}
        \label{fig:cora_impact_of_k_f1_256}
    \end{subfigure} %
    \hfill
    \begin{subfigure}[b]{0.23\textwidth}

        \includegraphics[width=\linewidth]{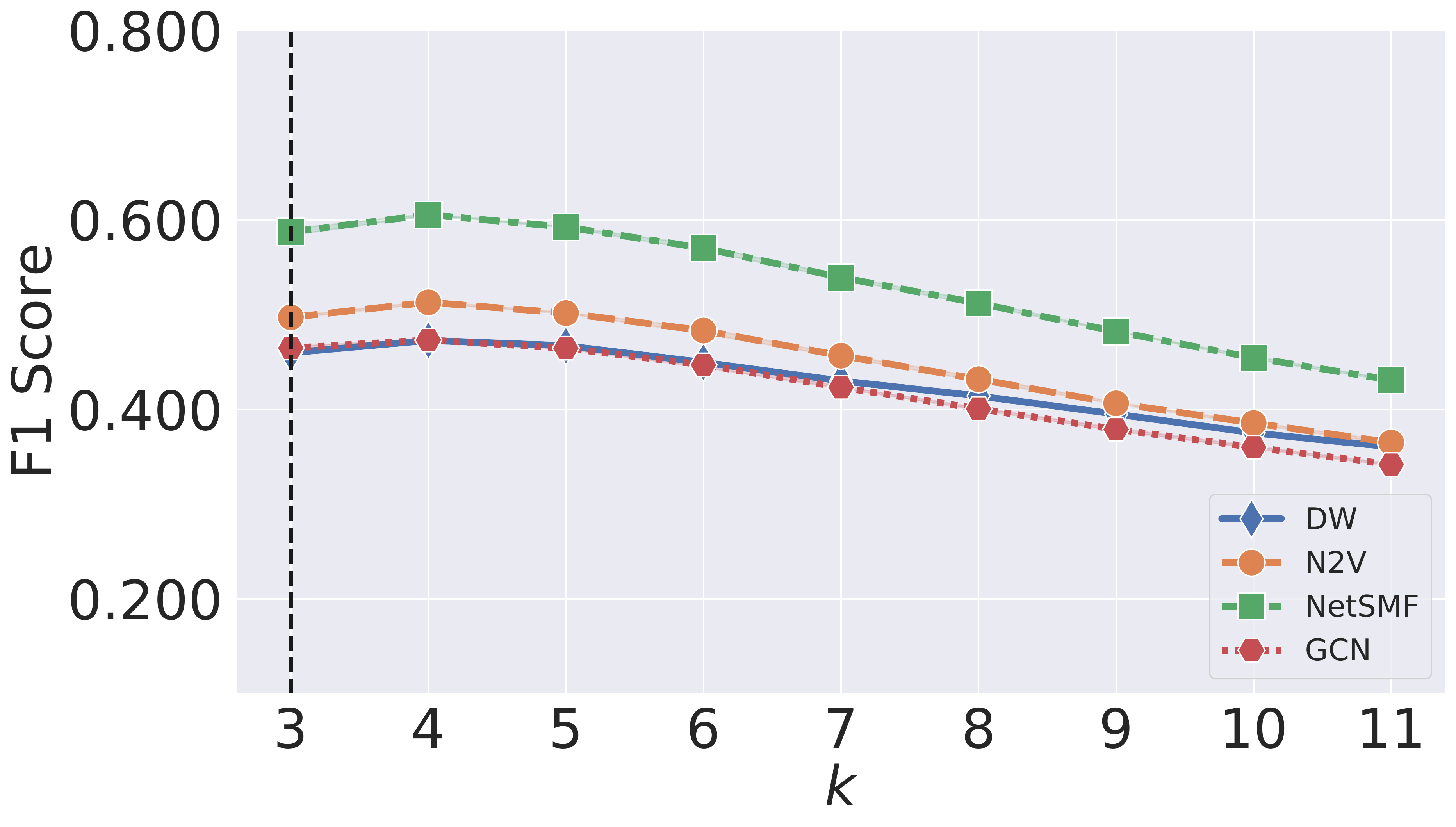}
        \caption{Citeseer}
        \label{fig:citeseer_impact_of_k_f1_256}
    \end{subfigure} %
    \hfill
    \begin{subfigure}[b]{0.23\textwidth}

        \includegraphics[width=\linewidth]{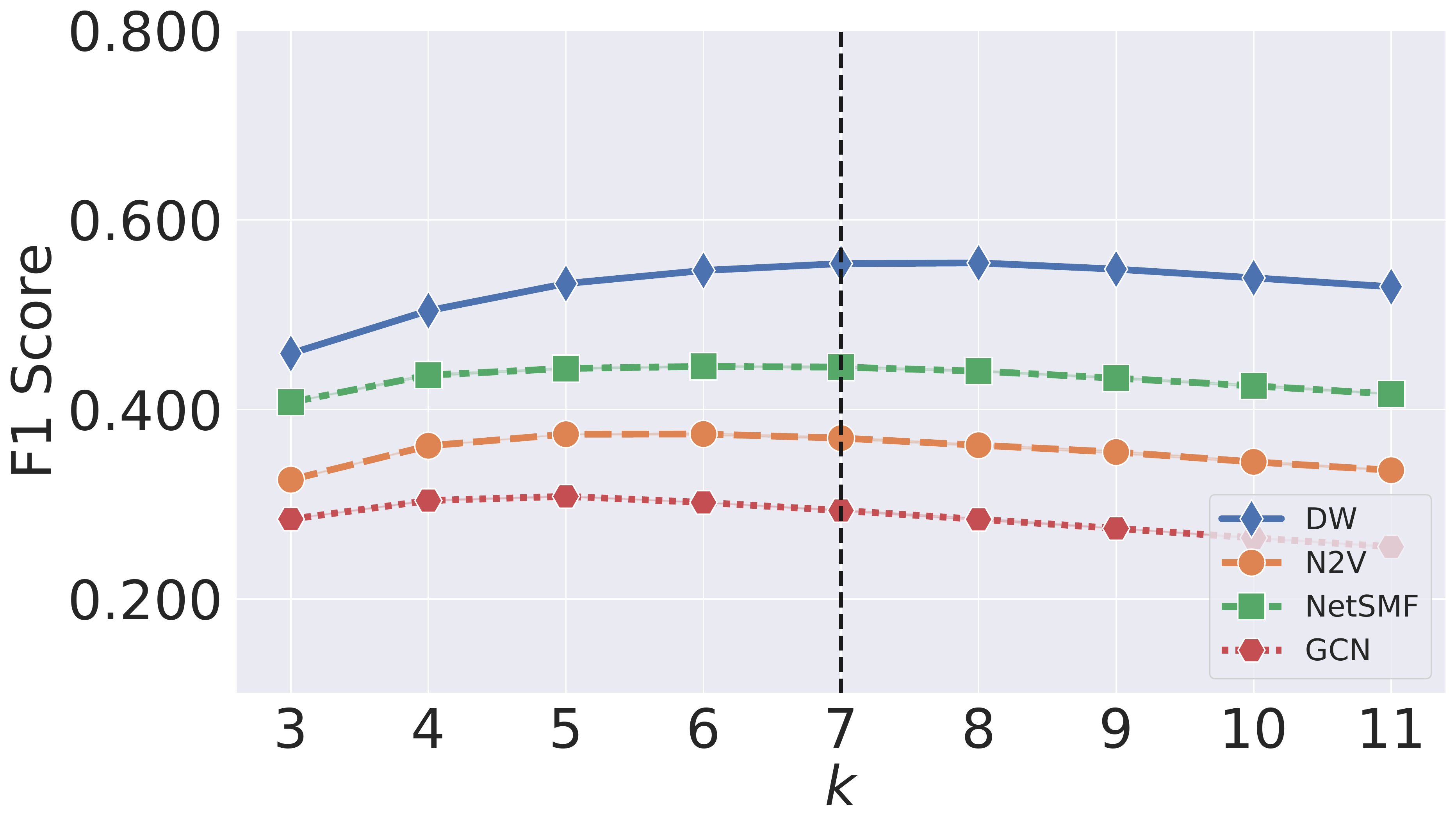}
        \caption{Actor}
        \label{fig:actor_impact_of_k_f1_256}
    \end{subfigure} %
    \hfill
    \begin{subfigure}[b]{0.23\textwidth}

        \includegraphics[width=\linewidth]{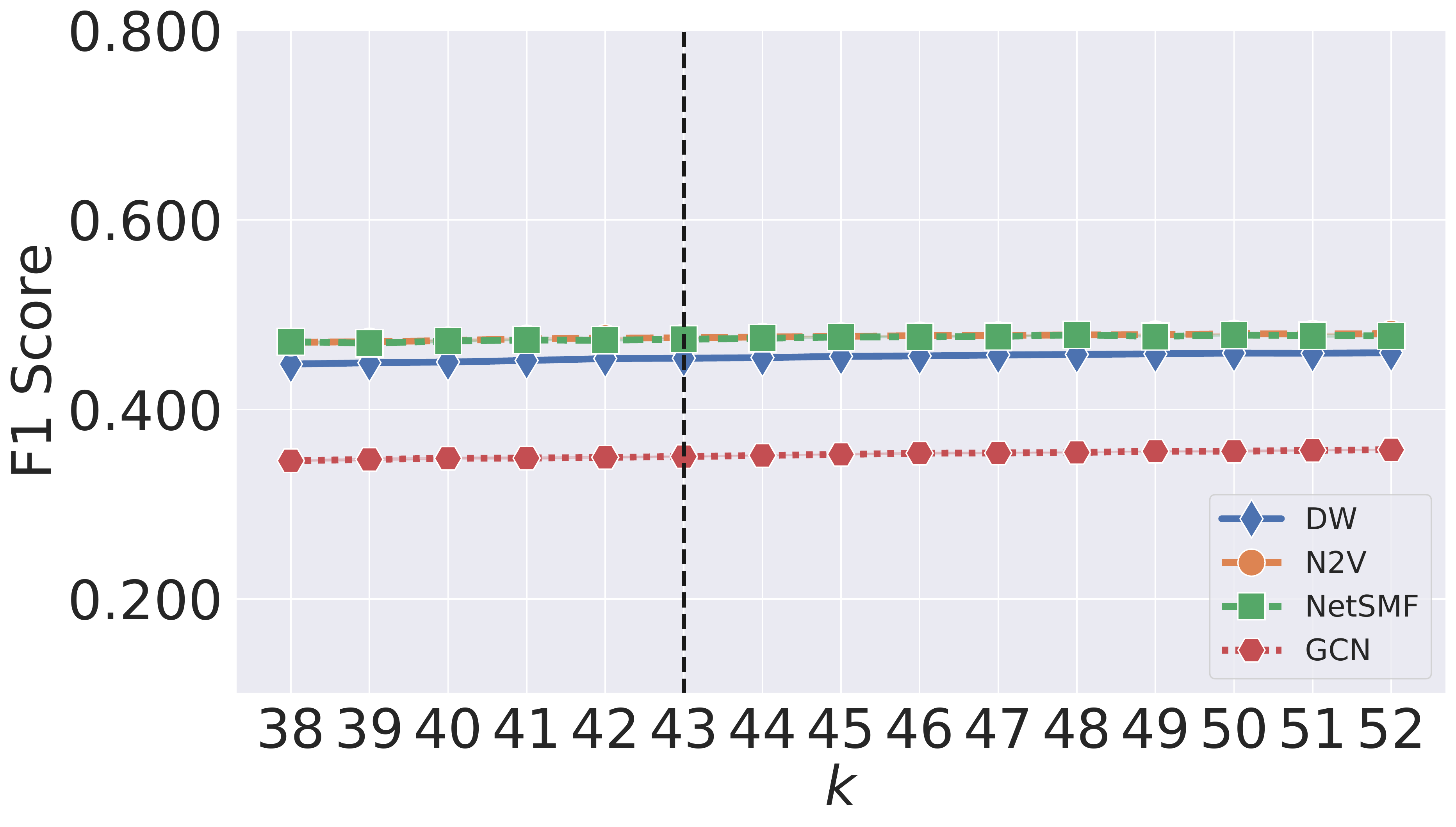}
        \caption{Facebook}
        \label{fig:facebook_impact_of_k_f1_256}
    \end{subfigure}%

    \begin{subfigure}[b]{0.23\textwidth}
        \includegraphics[width=\linewidth]{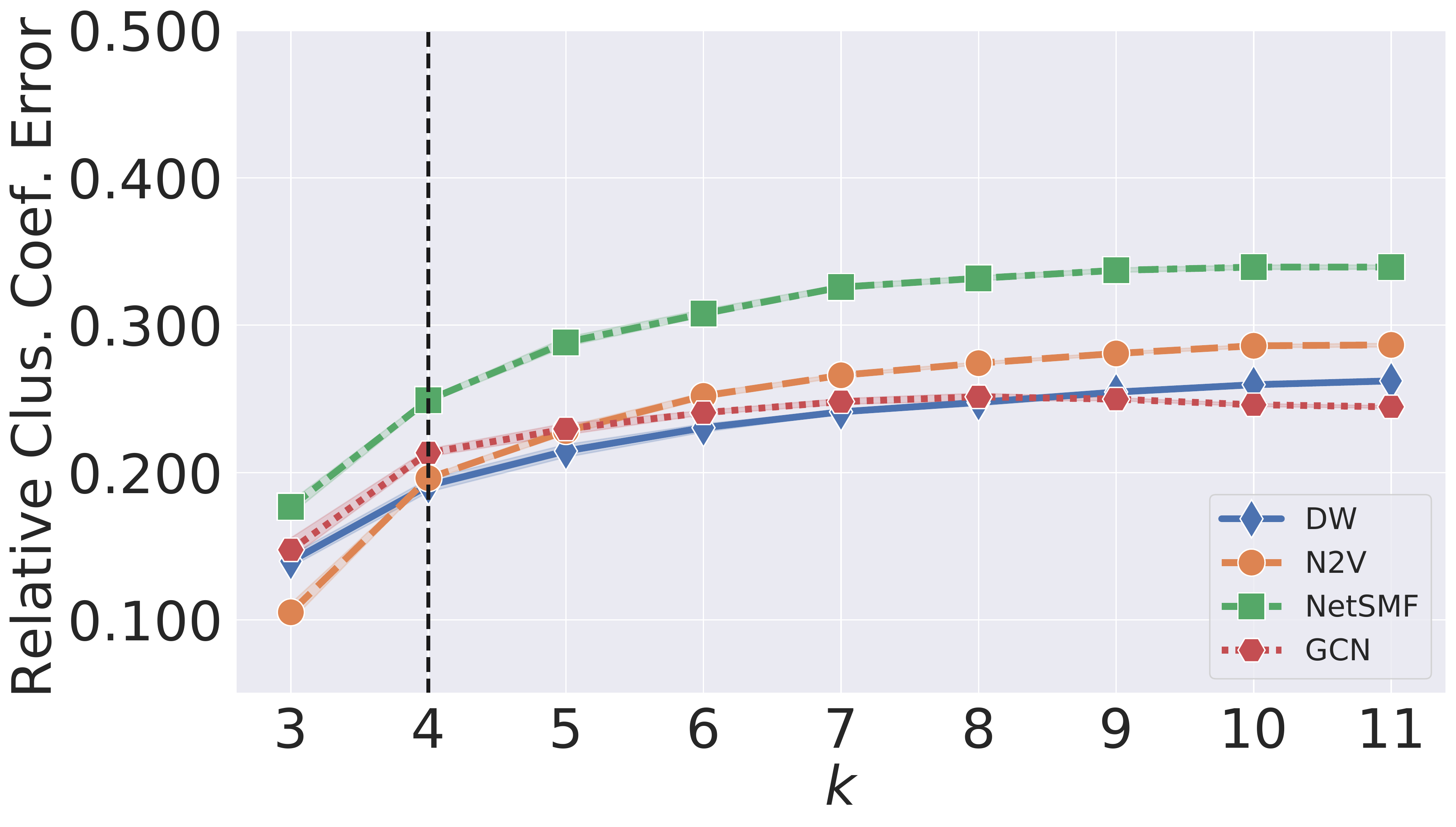}
        \caption{Cora}
        \label{fig:cora_impact_of_k_clus_256}
    \end{subfigure}%
    \hfill
    \begin{subfigure}[b]{0.23\textwidth}
        \includegraphics[width=\linewidth]{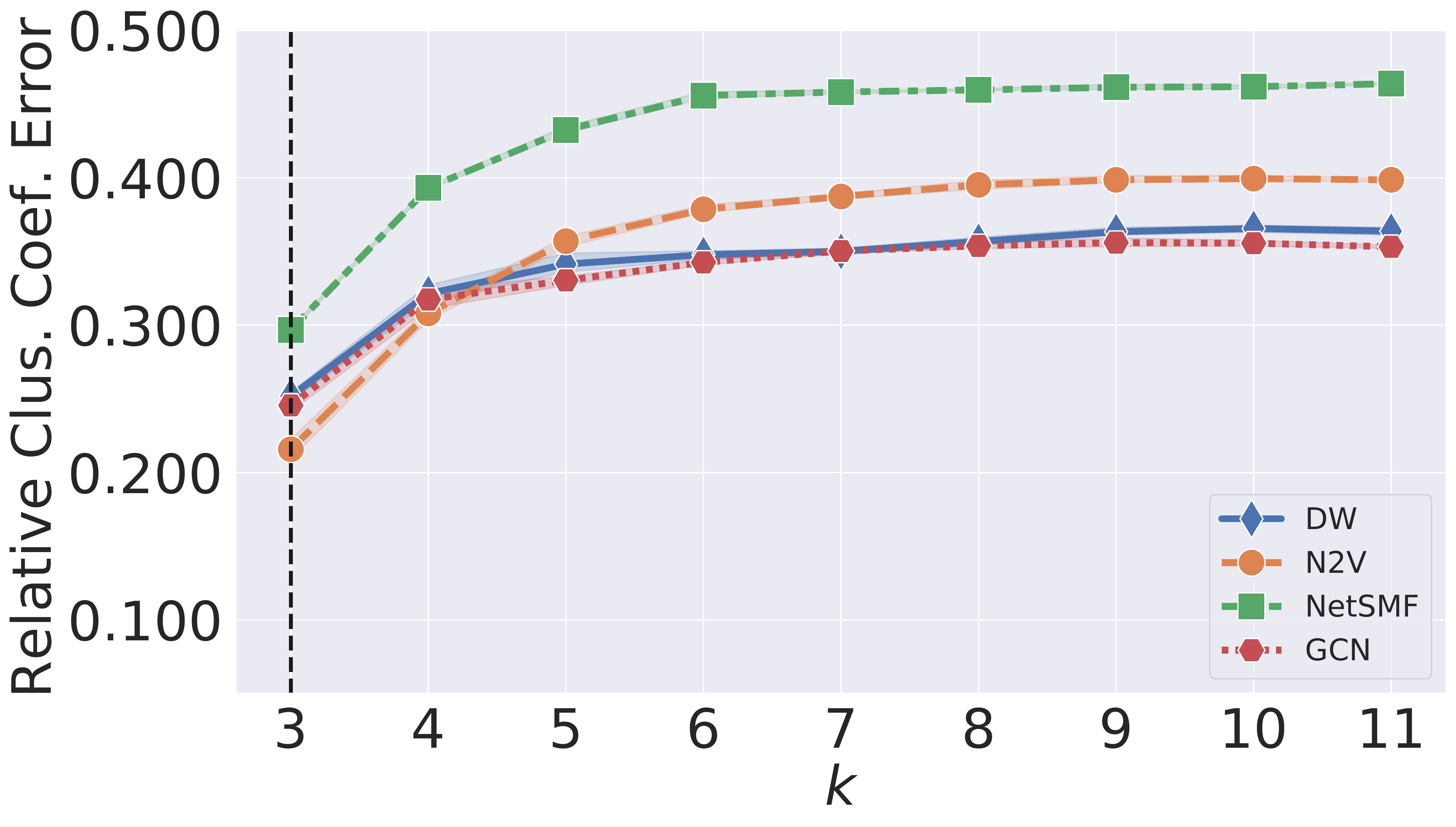}
        \caption{Citeseer}
        \label{fig:citeseer_impact_of_k_clus_256}
    \end{subfigure}%
    \hfill
    \begin{subfigure}[b]{0.23\textwidth}
        \includegraphics[width=\linewidth]{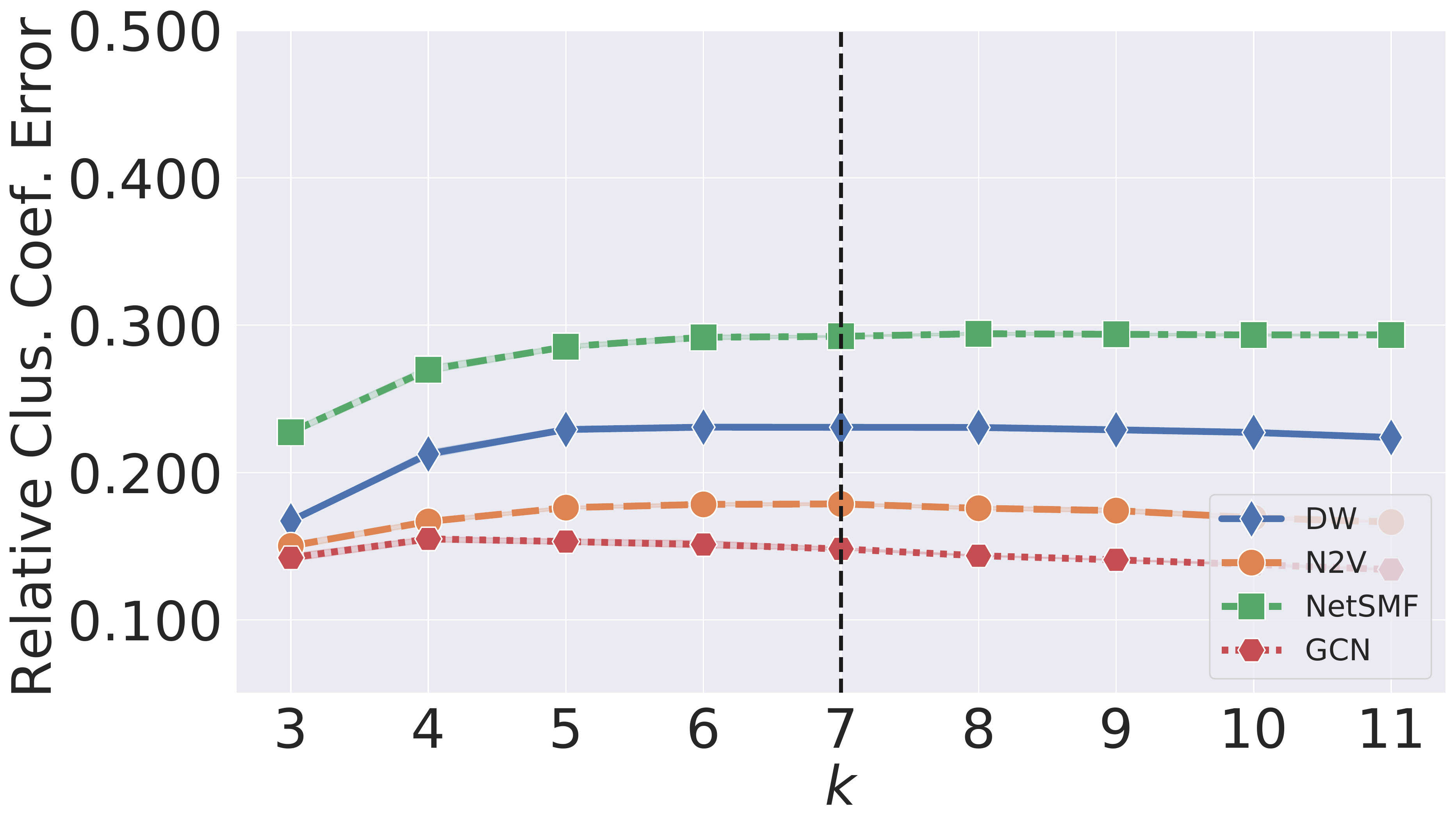}
        \caption{Actor}
        \label{fig:actor_impact_of_k_clus_256}
    \end{subfigure}%
    \hfill
    \begin{subfigure}[b]{0.23\textwidth}

        \includegraphics[width=\linewidth]{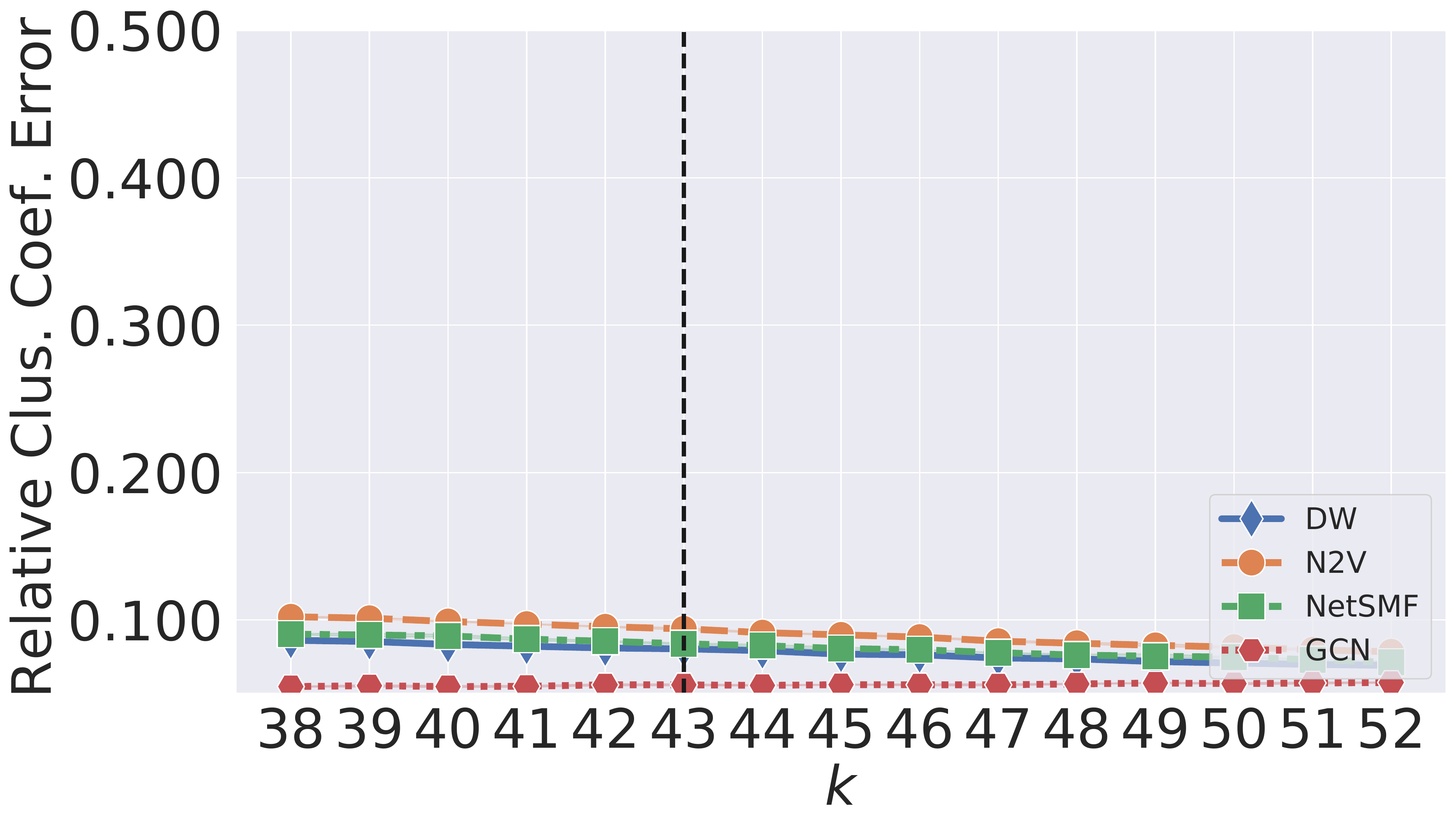}
        \caption{Facebook}
        \label{fig:facebook_impact_of_k_clus_256}
    \end{subfigure}
    \caption{F1 scores and relative average clustering coefficient error scores of \approach given all four datasets and various average node degrees. We fix the node embedding size to 256. The vertical bar indicates the actual average node degree.}
    \label{fig:impact_of_k}
    \vspace{-1em}
\end{figure*}

In this section, we evaluate \approach on all four datasets to understand its overall performance. %
The results are summarized in \autoref{tab:attack_256_results}.
Due to space limitations, we only show the results when the node embedding size is fixed to 256. 
The performance results of 64- and 128-dimensional node embeddings follow similar patterns and can be found in \autoref{sec:appendix_approach_effectivness}.

\mypara{Edge Metrics}
Recall that the primary goal of the attackers is uncovering the edges with decent accuracy from the node embedding matrix.
We thereby use edge metrics outlined in \autoref{sec:exp_setup} to measure \approach's performance.
Besides, $k$NN graph remains a viable approach to recover edges from the node embedding matrix as we explicate in \autoref{sec:exp_comparison}.
We also show the relative improvement scores in \autoref{tab:attack_256_results} to demonstrate to what extent \approach can relatively improve from $k$NN graph.
We add a positive sign (+) next to the relative improvement score to highlight the improvement.
As we can see from \autoref{tab:attack_256_results}, \approach can enable the adversary to recover edges from all node embedding matrices generated by all four node embedding models with good precision and recall.
Take the Cora dataset and the node embedding matrix generated by NetMF for example. 
\approach achieves 0.579 precision, 0.640 recall and 0.608 F1 scores. 
These scores relatively improve 0.235, 0.113, and 0.176 from those of $k$NN graph recovery.
Similarly, take the Actor dataset and the node embedding matrix generated by Deepwalk for example,
\approach achieves 0.687 precision, 0.435 recall, and 0.533 F1 scores. 
These scores relatively improve 0.222, 0.088, and 0.138 from those of $k$NN graph recovery.
The other datasets given all node embedding models follow similar patterns.
At the same time, \approach overwhelmingly outperforms $k$NN graph given joint degree distribution similarity metric.
Given the above two examples, JDD similarity scores of \approach respectively improve 1.191 and 0.587 from those of $k$NN graph. 
Our results demonstrate that \approach can recover a graph in which each pair of connected nodes share similar 1-hop neighborhood as they are in the original graph.

\mypara{Global Metrics}
Recall that the secondary goal of the attackers is recovering a graph structure that is similar to the original graph with respect to the graph properties.
We use the global metrics outlined in \autoref{sec:exp_setup} to understand \approach's performance.
Similar to the above edge metrics, we also compare \approach to $kNN$ graph.
We show the relative error reduction scores in \autoref{tab:attack_256_results} to demonstrate to what extent \approach can relatively reduce errors incurred by $k$NN graph.
We add a negative sign (-) next to the relative error reduction score to highlight the difference between \approach and $k$NN graph.
As we can see from \autoref{tab:attack_256_results}, \approach can incur relatively low error scores given all three global metrics.
Take the Actor dataset and the node embedding matrix generated by GCN for example.
\approach's relative triangle error is 0.280.
This indicates that the graph recovered by \approach contains a similar number of triangles to that of the original graph.
At the same time, this score reduces the relative error made by $k$NN  graph for 0.349.
Note that the estimated average node degree (i.e., 5) is larger than the ground truth values of both Cora and Citeseer.
This leads to higher relative triangle errors.
However, combined with the edge metrics, we can assert that such error is due to a combination of reorientation of the specific edges between the true and the recovery networks, and extra edges incurred by the overestimation of $k$.
Besides, we compare \approach with the invert embedding using the overlapping Citeseer dataset and its 256-dimensional NetSMF node embedding matrix. 
\approach can achieve 0.908 relative Frobenius error score which is close to that of the invert embedding (see Figure 4 in \cite{chanpuriya2021deepwalking}).
Note that the invert embedding in~\cite{chanpuriya2021deepwalking} is under the white box setting while \approach is under the black box setting.
In summary, our performance results demonstrate that \approach can also recover a graph that is structurally similar to the original graph with respect to the global graph properties.

Note that the clustering coefficient of a node $c_v$ is defined as $c_v = \frac{2*\mathcal{N}(v)}{deg(v) (deg(v)-1)}$ where $\mathcal{N}(v)$ represents the number of edges between the neighbors of $v$ and $deg(v)$ represents the degree of $v$.
The average clustering coefficient of the whole graph $c_\mathbf{G}$ is defined as $c_\mathbf{G} = \frac{1}{n} \sum_{v=1}^n c_v$. 
A smaller $k$ may lead to the increasing possibility that the number of triangles recovered from a graph drops closer to 0.
We therefore use different relative errors to objectively evaluate the performance from multiple perspectives.

\mypara{Impact of Node Embedding Size}
We use two metrics - F1 score and relative average clustering coefficient error - to understand the impact of node embedding size \approach across all four datasets.
The results are shown in \autoref{fig:attack_all_datasets_all_dims}.
As we can see in the figure, \approach can offer stable graph recovery performance given different embedding sizes and all embedding models.
We only observe a marginal F1 score decrease given the Deepwalk node embedding model in Cora and Citeseer datasets.
Overall, our results imply that reducing the embedding size (a common defense mechanism) may not work for \approach. 
More details can be found in \autoref{sec:defense}.

\mypara{Stability}
We run \approach 5 times on a given node embedding matrix. 
Such runtime configuration enables us to measure how widely all those metric values are dispersed from the average value (i.e., standard deviation).
At the same, it eliminates the chance of reporting opportunistically good results.  
A low standard deviation indicates low volatility.
As we can observe in \autoref{tab:attack_256_results}, the standard deviation values are low in all cases.
The results show that \approach can recover graphs from node embedding matrices with statistically stable performance.

\mypara{Ablation Study}
We also carry out an ablation study to understand the impact of GML on the performance of \approach.
To this end, we customize \approach and remove GML from the optimization.
Specifically, we initialize a graph using Gumbel-Top-$k$ trick and run GAE once.
We use edge metrics in this study and summarize the results in \autoref{table:ablation_study}.
We observe that \approach performs better given all edge metrics. 
The results exemplify that jointly optimizing GAE and GML enables us to learn more information from the node embedding matrix and further reduce noise from the recovered graph.

\begin{table}[t]
\centering
\caption{Ablation study on the impact of GML to \approach's performance. We use Citeseer dataset and fix the node embedding size to 256.}
\resizebox{0.75\linewidth}{!} {
 \begin{tabular}{cccc} 
 \toprule
 \textbf{Method} & \textbf{Precision} & \textbf{Recall}  & \textbf{F1} \\  
 \midrule
$k$NN & 0.338 & 0.483 & 0.398 \\ 
\approach w/o GML  & 0.391 & 0.511 & 0.443   \\
\approach & 0.404 & 0.557 & 0.468   \\
 \bottomrule
 \end{tabular}
 }
\label{table:ablation_study}
\end{table} 

\mypara{Takeaways}
We can observe that \approach achieves good performance on all datasets. 
Such results demonstrate that jointly optimizing the learnable distance function and adaptive graph structure combination is effective for recovering graphs from node embeddings.

\subsection{How does $k$ Affect the Attack Performance?}
\label{sec:exp_impact_of_k}

Recall that the attackers use the graph sampling algorithms to estimate the average node degree from the graphs of similar origins and transfer the estimated node degree from these graphs to facilitate the attack (see \autoref{sec:estimate_average_node_degree}).
We show that the adversary cannot obtain the precise average node degree in \autoref{sec:exp_estimate_avg_degree}.
However, the estimated average node degree $k$ directly affects the graph size of the recovered graph $\mathbf{G}_R$. 
More importantly, our attack uses this estimated $k$ to seed the initial graph using the Gumbel-Top-k trick and iterative graph structure optimization during the learning process.
It is therefore essential to study the impact of the estimated average node degree $k$ on the graph recovery performance.
To this end, we run \approach 5 times on every $k$ that falls within at least one standard deviation of the estimated average node degree in \autoref{sec:exp_estimate_avg_degree}.
For instance, the mean and standard deviation of our estimated average node degree of the Facebook dataset are 45.7 and 8.4 respectively.
In this case, we run \approach 5 times for every value between 38 and 52.
We use two metrics - F1 score and relative average clustering coefficient error - to understand the impact of $k$ across all four datasets.
Due to space limitations, we only show the attack results when the node embedding size is fixed to 256. 
The performance results using 64- and 128-dimensional node embeddings follow similar patterns and can be found in \autoref{appendix:exp_impact_of_k}.

\mypara{Performance}
The results are shown in \autoref{fig:impact_of_k}.
We show a vertical line in each figure to mark the ground truth value of the average node degree for each dataset (see \autoref{tab:graph_stats}).
In general, we can see that \approach can accommodate the inevitable estimation error.
Take the Citeseer dataset and the node embedding matrix generated by NetSMF for example. 
The ground truth average node degree is 3, and the estimated average node degree is 5.
The F1 scores achieved by \approach are respectively 0.592 and 0.605.
This means that, even though our estimated $k$ is almost twice the ground truth value, \approach can still deal with such estimation error and attain good results (i.e., the F1 score difference is only 0.013).
Similar to the F1 score metric, we can see that \approach can also achieve low relative average clustering coefficient error.
Take the Citeseer dataset and the node embedding matrix generated by NetSMF for instance, the relative average clustering coefficient errors of \approach are 0.432 and 0.332.
The difference is approximately 0.100.

\mypara{Observations}
We observe that the node embedding matrices generated by GCN are relatively harder to recover than those by the other three models.
Two factors make the graph recovery task difficult.
First, GCN considers both node features and graph structure to generate node embeddings while the other three models only use graph structures.
Second, we use ReLU as the activation function between layers.
This non-linear, element-wise function outputs the input directly if it is positive, otherwise, it outputs zero.
It is computationally efficient but leads to sparse representation, making the graph recovery harder.

\mypara{Takeaways}
Our evaluation results show that \approach can accommodate the inevitable estimation error of $k$.
The root cause of the moderate decrease (increase) of F1 scores (relative average clustering coefficient error) in \autoref{fig:impact_of_k} is due to the increasing graph size.
However, given the real world application scenarios outlined in \autoref{sec:attack_application_scenario}, we show that the estimated average node degrees can be in the vicinity of the real values as shown in \autoref{sec:exp_estimate_avg_degree}.
Combining with the adaptive learning process outlined in \autoref{sec:iterative_GAE}, \approach remains practical to recover graphs in the wild as exemplified by the results in \autoref{fig:impact_of_k}.

\section{Defense}
\label{sec:defense}

\begin{figure}[t]
\begin{center}
\includegraphics[width=0.6\linewidth]{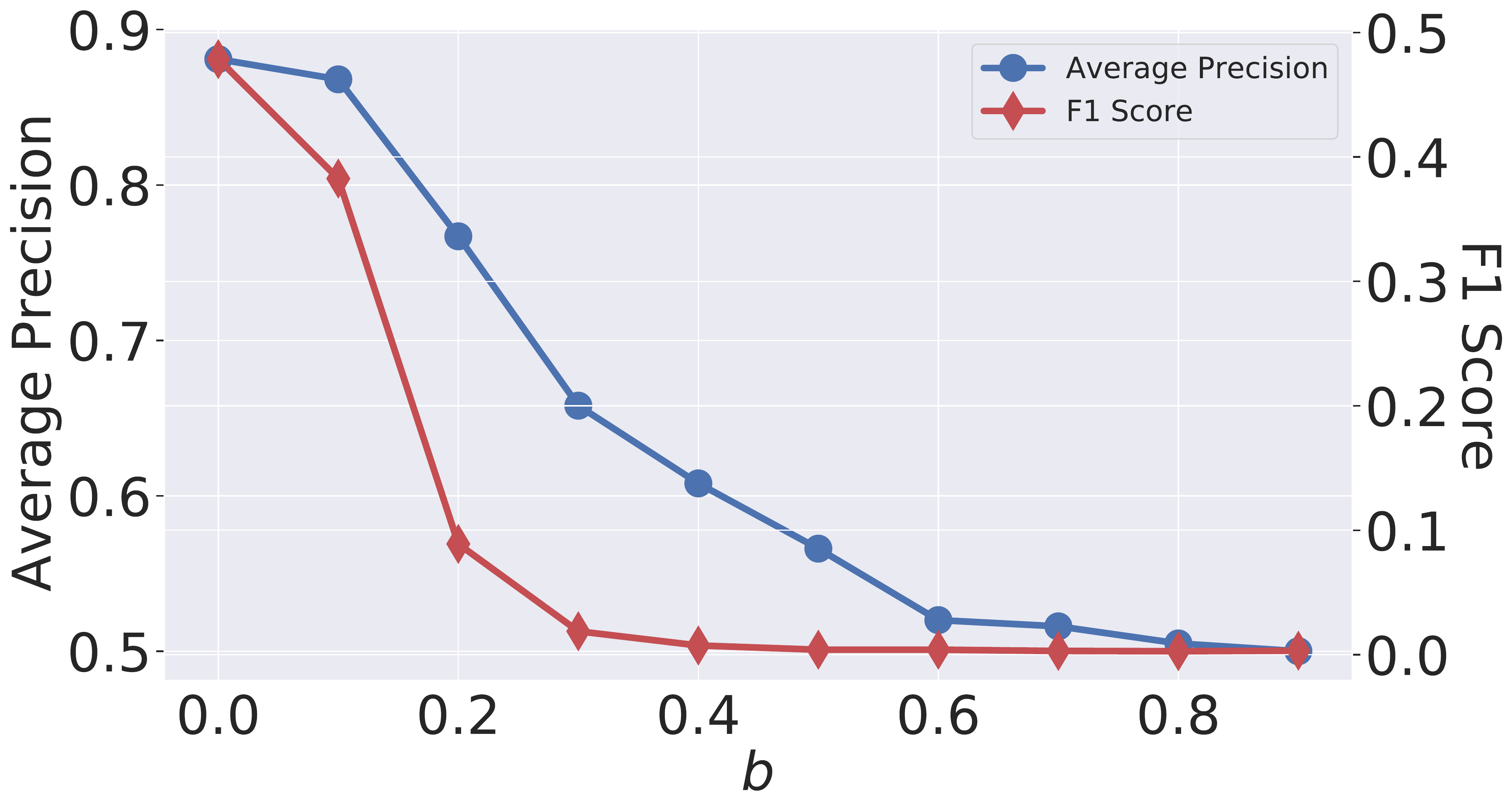}
\end{center}
\caption{
trade-off between node embedding utility (average link prediction precision) and \approach's graph recovery performance (F1 score).
We use the Cora dataset and GCN as the node embedding model.
The node embedding size is fixed to 256.
}
\label{fig:cora_utility_defense}
\vspace{-1em}
\end{figure}

In this section, we discuss node embedding perturbation as a tentative defense mechanism and empirically evaluate its effectiveness.

\mypara{Embedding Perturbation}
One possible defense of \approach is adding perturbations (i.e., noise) to the original node embeddings $\mathbf{H}_O$.
As such, the data holder only passes on a noisy but usable version $\widetilde{\mathbf{H}}_O$ to the ML pipeline.
Formally, $\widetilde{\mathbf{H}}_O = \mathbf{H}_O + \Delta(\mu, b)$ where $\Delta$ denotes the Laplace distribution, $\mu$ is a location parameter and $b>0$ is a scale parameter.
However, adding noise inevitably distorts the information contained in the node embeddings and can lead to utility loss.
We therefore focus on evaluating the trade-off between the utility and the defense in this section.
Similar defense mechanisms were also discussed in the previous literature~\cite{he2021stealing,zhang2022inference}.

\mypara{Experimental Setup}
We use the Cora dataset and the 256 dimensional node embedding matrix generated by GCN for our evaluation.
We fix $\mu$ to 0 and choose 10 evenly distributed values between 0 and 1 for $b$ (i.e., $b=\{0, 0.1, ..., 0.9\}$).
We use average link prediction precision as the utility metric and F1 score as the attack performance metric.

\mypara{Results}
The results are shown in \autoref{fig:cora_utility_defense}.
As we can see in the figure, adding perturbations could work with noticeable utility loss.
For instance, when $b=0.2$, the average link prediction precision using the perturbated node embedding matrix drops from 0.881 to 0.761.
In turn, we can see that the \approach's F1 score drops from 0.486 to 0.088.
This result shows that the data holder might choose the noise level to defend against \approach while preserving some utility.
However, it is a delicate process.
For instance, if the data holder chooses $b=0.1$, the average link prediction precision drops from 0.881 to 0.868.
In this case, \approach's F1 score drops from 0.486 to 0.401. 
In short, our results show that the trade-off is inevitable if using added perturbations to defend against \approach.
We plan to explore such research direction in the future.

\mypara{Notes}
The node embedding size affects the expressiveness of the node embeddings.
As such, another prospective defense mechanism is to reduce the dimension of the node embedding.
Its core idea is reducing the knowledge that the attackers can obtain and consequently lessening the capability of the graph recovery attack.
However, we show that \approach achieves stable graph recovery performance given different embedding sizes and all embedding models in \autoref{fig:attack_all_datasets_all_dims}.
Our results indicate that reducing the embedding size may not work for \approach. 
We plan to expand such research direction in the future.

\section{Related Work}
\label{sec:related_work}

\begin{table}[t]
\centering
\caption{Difference between our attack and the close work.}
\resizebox{0.99\linewidth}{!} {
\begin{tabular}{lcccc}
\toprule
\multicolumn{1}{l}{\textbf{Method}} &
 \textbf{\begin{tabular}[c]{@{}c@{}}Supervision from \\Auxiliary data\end{tabular}} &
  \textbf{\begin{tabular}[c]{@{}c@{}}Shadow\\ model\end{tabular}} &
  \textbf{\begin{tabular}[c]{@{}c@{}}Interaction w/\\ target model\end{tabular}} &
  \textbf{\begin{tabular}[c]{@{}c@{}}Attack\\ setting\end{tabular}} \\ \midrule
\textbf{Chanpuriya et al.}~\cite{chanpuriya2021deepwalking} & \cmark  & \xmark & N/A & whitebox \\ \midrule
\textbf{Link reidentification~\cite{he2021stealing, duddu2020quantifying, wu2022linkteller}} & \cmark & \cmark & \cmark  & blackbox \\ \midrule
\textbf{Zhang et al.}~\cite{zhang2022inference}  & \cmark & \cmark & \cmark  & blackbox \\ \midrule
\approach  & \xmark  &\xmark  & \xmark &  model-agnostic \\
\bottomrule
\end{tabular}
}
\label{tab:closest_work}
\vspace{-1em}
\end{table}

\mypara{Graph Theory Based Graph Restoration} 
Graph restoration algorithms in the graph theory realm restore a hidden graph by repeatedly querying an oracle for certain types of information about the graph structure~\cite{lauri2016topics}.   
Depending on the algorithm, different types of information can be revealed by the oracle, including node betweenness~\cite{abrahamsen2016graph}, distance or shortest path between nodes~\cite{hermelin2011distance},  edge counting~\cite{bouvel2005combinatorial}, edge detection~\cite{angluin2008learning}, etc.
The common goal among these research is identifying strategies that recover the graph with low worst-case query complexity.
However, those approaches are not learning-based and require the existence of an oracle knowing the structural information of the original graph.
They cannot be adapted to reconstruct graphs from the node embeddings.

\mypara{Graph Completion}
Graph completion~\cite{clauset2008hierarchical,guimera2009missing} aims at inferring the unobserved part of the network (i.e., missing edges and nodes) given the partially observed network.
Link prediction algorithms (see~\cite{lu2011link,dunlavy2011temporal} for an overview) have been actively investigated and successfully applied to identify missing edges~\cite{zhang2018link,trouillon2016complex}.
Probabilistic and deep learning models~\cite{sina2013solving,hric2016network} have also been investigated to deduce the missing nodes. 
However, these algorithms require the graphs to be substantially observed and high-quality attribute information provided.
Our attack assumes neither. 

\mypara{Deep Graph Structure Learning for Robust Representations}
This line of research centers on Graph Structure Learning (GSL) that jointly learns an optimized graph structure and corresponding representations~\cite{zhu2021deep}.
The goal of GSL is to generate node representations robust to noisy graph structures.
Common assumptions of these methods include the availability of node features, incomplete graph structure, and node labels. 
Different approaches then leverage metric learning~\cite{chen2020iterative}, probabilistic modeling~\cite{franceschi2019learning}, direct optimization~\cite{yang2019topology}, etc. to learn an adjacency matrix as well as the corresponding node representations.
In contrast to GSL, our attack does not assume the availability of node features and node labels.
Besides the goal difference, our attack is self-supervised while GSL approaches use node labels to supervise the learning process. 

\mypara{Close Work}
To our best knowledge, there exist five pieces of close work to our attack~\cite{he2021stealing,zhang2022inference,wu2022linkteller,chanpuriya2021deepwalking, duddu2020quantifying}.
The closest work is Chanpuriya et al.~\cite{chanpuriya2021deepwalking} presenting two optimization algorithms to recover a graph from its node embeddings generated by NetMF~\cite{qiu2018network}. 
Their algorithms assume the knowledge of the NetMF algorithm (i.e., the target model in our terminology), window size $T$, the low-ranking approximation of the finite-$T$ positive pointwise mutual information (PPMI) matrix, and the exact degree of each node, hence a specific white-box attack against NetMF only.
Another closely related work is link reidentification attack~\cite{he2021stealing,wu2022linkteller,duddu2020quantifying} from node-level information. 
In theory, those attacks can be used to reconstruct a graph upon querying the target model $n^2$ times.
However, they train a shadow model using auxiliary data and their posterior scores obtained from the target model.
Our attack assumes the attackers can not interact with the target model using auxiliary data, which renders this link stealing attack infeasible in our setting.
In addition to link re-identification attacks using node-level information, Zhang et al.~\cite{zhang2022inference} also introduce a reconstruction attack to rebuild a graph from its graph-level embedding within the context of graph classification.
This attack suffers from the same pitfalls of the link re-identification attacks, and cannot be used in our setting. 
In short, our attack is fundamentally different from the existing work by removing the assumptions of the availability of supervision information from auxiliary data, the shadow model, and the interaction with the target model. 
We summarize the differences between our attack and the closely related work in \autoref{tab:closest_work}.

\section{Conclusion}
\label{sec:conclusion}
In this paper, we presented a model-agnostic attack that uses the node embedding matrices to recover graphs.
Extensive experiments show that an adversary can recover graphs with decent accuracy by only gaining access to the node embeddings of the original graph.
Our results highlight the need for the data holders to rethink the privacy implications when integrating node embeddings for downstream analysis, even when the third party has extremely limited knowledge of the data.

\section*{Acknowledgments} 
We wish to thank the anonymous reviewers for their feedback and our shepherd Gergely Acs for his help in improving our paper. 
This work is partially funded by the Helmholtz Association within the project ``Trustworthy Federated Data Analytics'' (TFDA) (funding number ZT-I-OO1 4) and by the National Science Foundation under grant CNS-2127232.

\balance
\bibliographystyle{plain}
\bibliography{refs_clean}

\appendix

\begin{figure*}[t]
     \centering
    \begin{subfigure}[t]{0.19\textwidth}
    \centering
        \includegraphics[width=\linewidth]{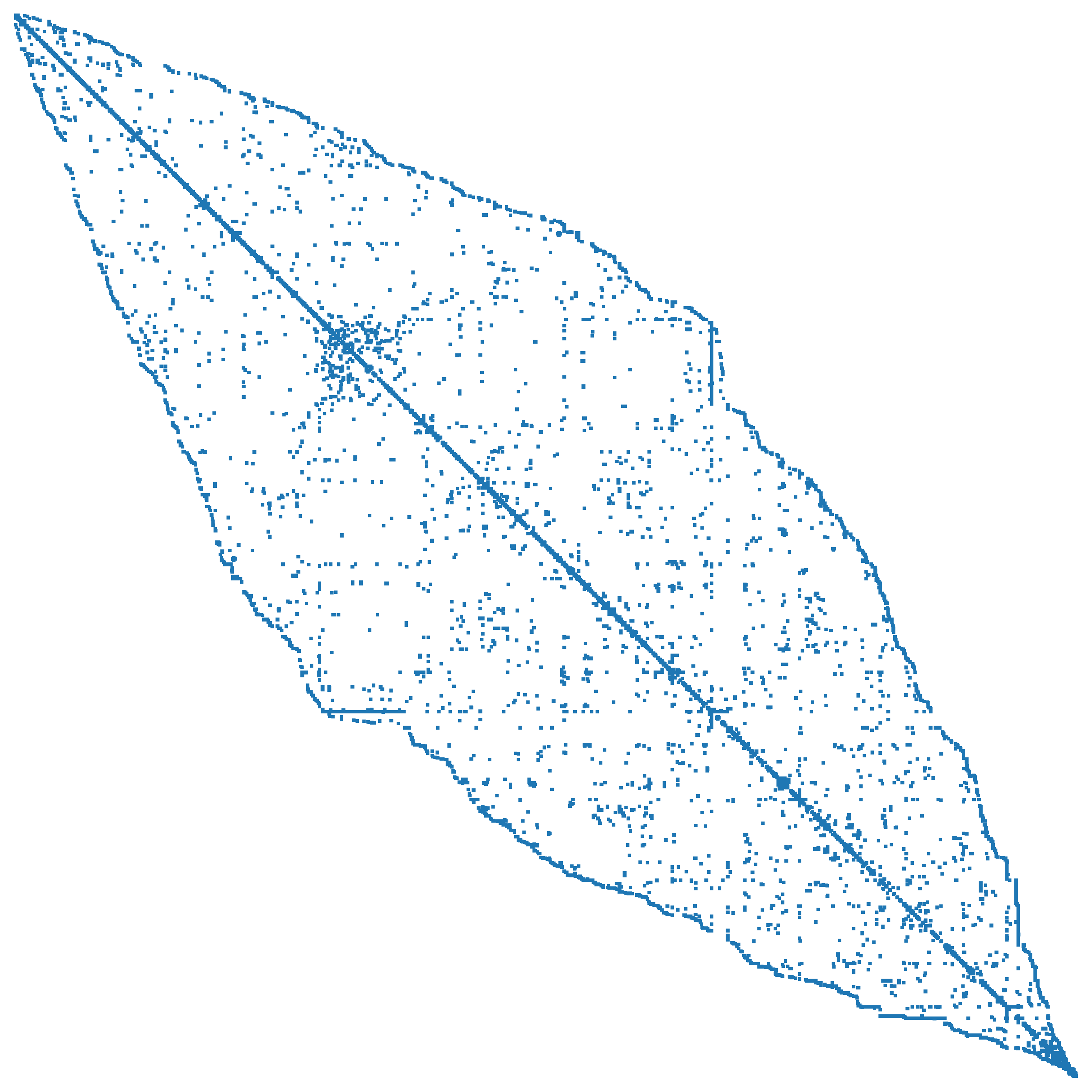}
        \caption{original}
        \label{fig:bitmap_original}
    \end{subfigure}
    \hfill
    \begin{subfigure}[t]{0.19\textwidth}
    \centering
        \includegraphics[width=\linewidth]{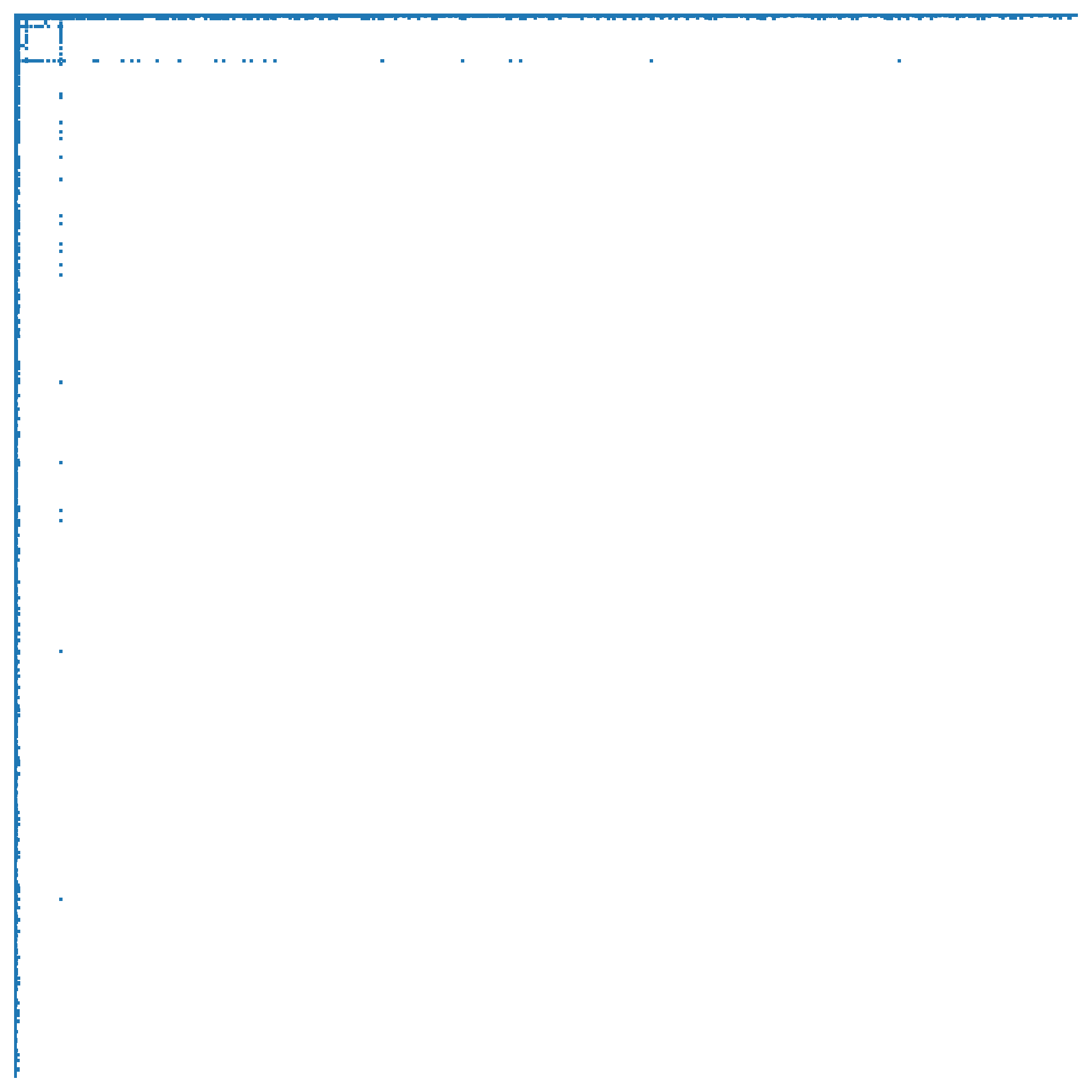}
        \caption{direct recovery}
        \label{fig:bitmap_direct}
    \end{subfigure}
    \hfill
    \begin{subfigure}[t]{0.19\textwidth}
    \centering
        \includegraphics[width=\linewidth]{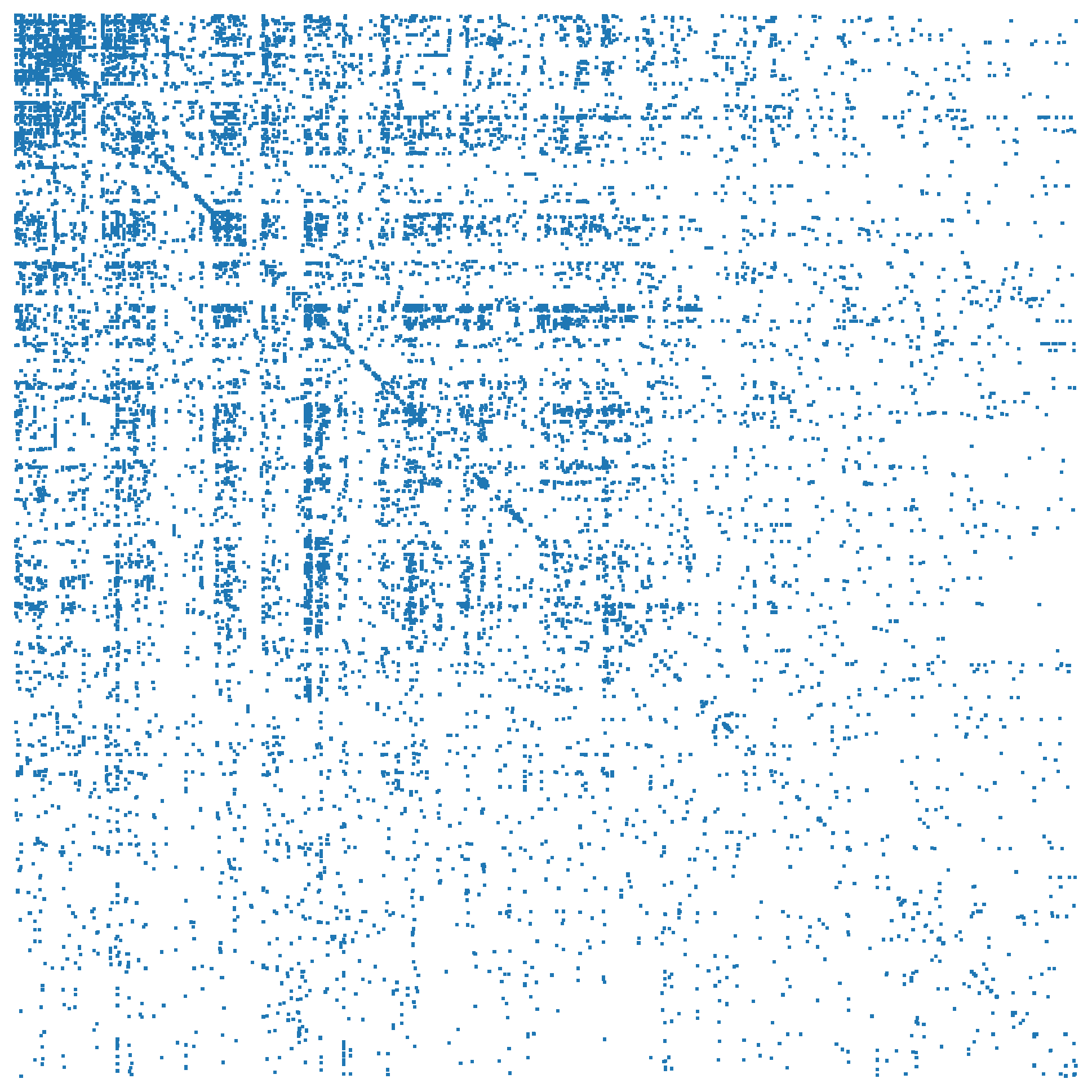}
        \caption{invert embedding}
        \label{fig:bitmap_invert}
    \end{subfigure}
    \hfill
    \begin{subfigure}[t]{0.19\textwidth}
    \centering
        \includegraphics[width=\linewidth]{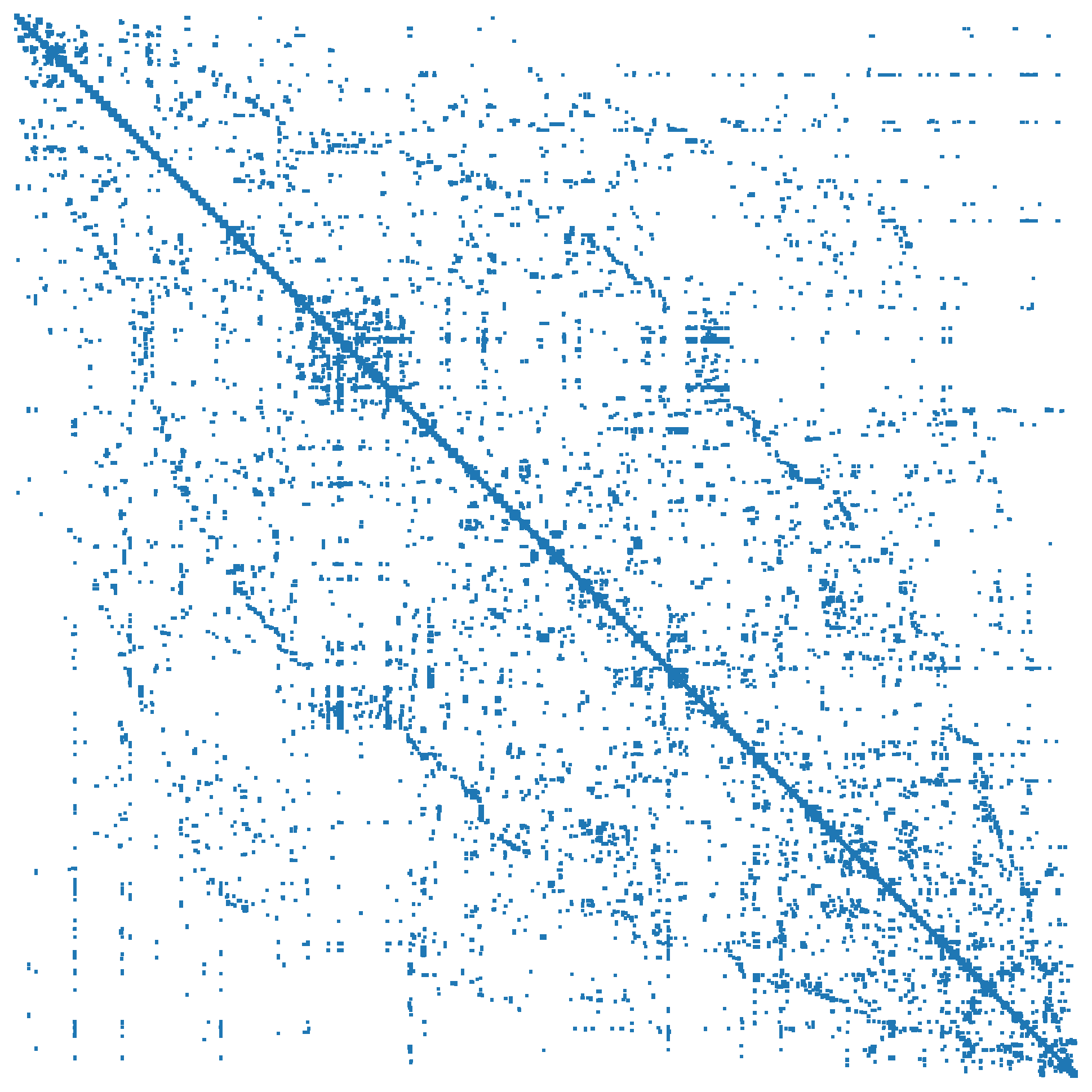}
        \caption{$k$NN graph}
        \label{fig:bitmap_knn}
    \end{subfigure}
    \hfill
    \begin{subfigure}[t]{0.19\textwidth}
    \centering
        \includegraphics[width=\linewidth]{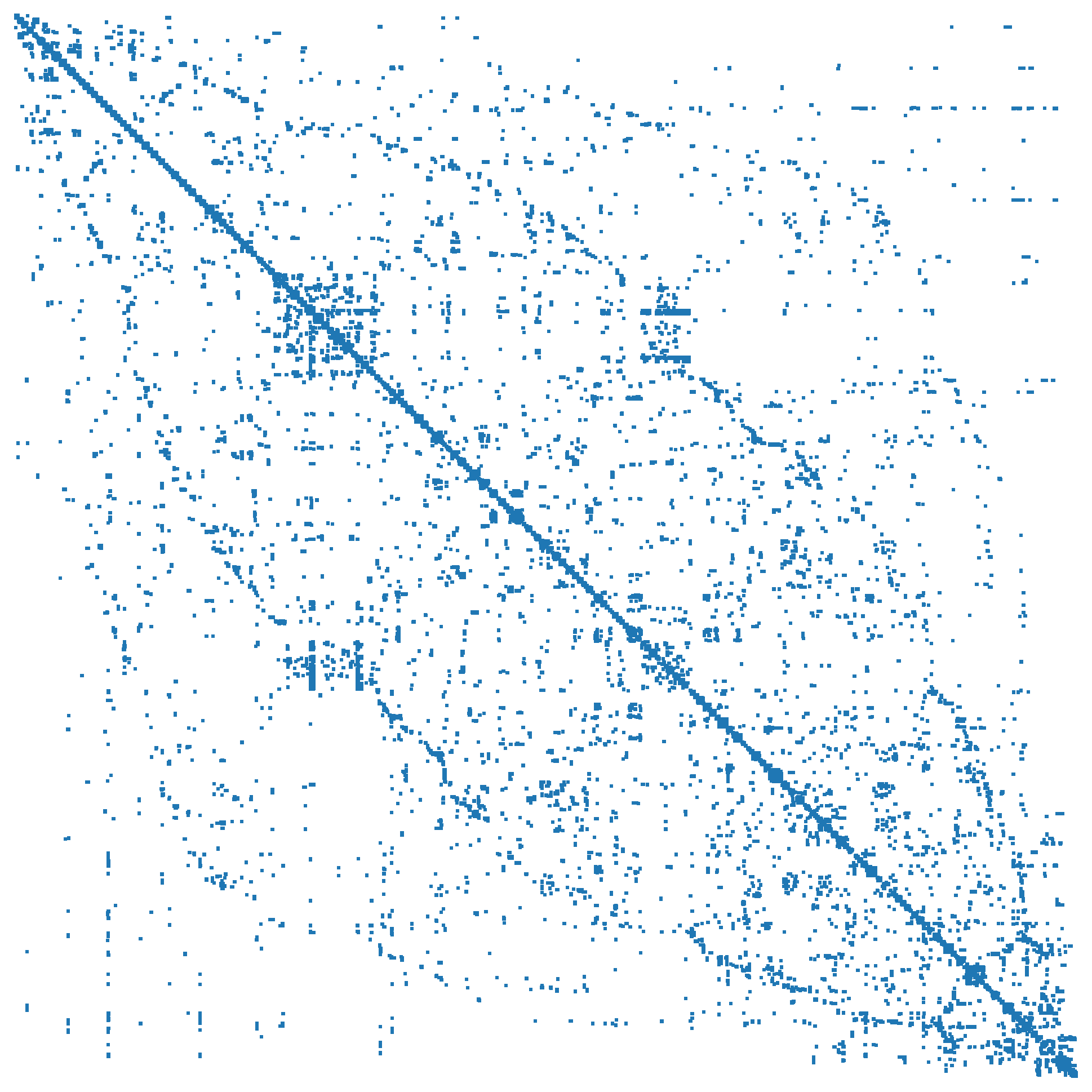}
        \caption{\approach}
        \label{fig:bitmap-ours}
    \end{subfigure}
    \caption{Bitmap visualization of the recovered graphs by all baselines and \approach. \approach, visually, removes fair amount of false positive edges.}
    \label{fig:visualization}
\end{figure*}

\section{\approach Algorithm Details}
\label{sec:attack_algorithm}

\mypara{Walk Through}
We summarize the whole learning process (i.e., \autoref{sec:metric_learning} and \autoref{sec:iterative_GAE}) in \autoref{alg:attack}.
Line 1 uses spikyball graph sampling algorithm to sample from graphs of similar origins to estimate a rough node degree of a given node embedding matrix.
Line 2 applies Gumbel-Top-$k$ trick on the fully connected probabilistic graph $\mathbf{P}$ to generate the initial seed graph structure $\mathbf{A}^0$.
Line 4 - 10 learn a distance function $\phi$ and a corresponding graph structure $\mathbf{A}^t$ in each iteration (line 5-8).
Upon termination of the learning process after a maximum iteration $T$, we obtain the recovered graph structure $\mathbf{A}^T$ (i.e., $\mathbf{G}_R$).

\smallskip
\mypara{Complexity Analysis}
Our graph initialization generates a fully connected probabilistic graph using \autoref{eq:edge_probability}.
The time and space complexity of graph initialization are both $O(n^2)$.
Our graph metric learning learns a global distance metric, leading to a $O(n^2)$ time complexity.
The encoder of the GAE has a $O(n^2)$ time and space complexity, while the inner-product decoder of the GAE has a $O(n^2)$ time complexity.
Overall, the time and space complexity of \approach are both $O(cn^2)$.
Besides, Zhang et al.~\cite{zhang2022inference} has a time complexity of $O(n^4)$ while LinkTeller~\cite{wu2022linkteller} has a space complexity of $O(n^3)$, which inevitably limit their scalability in the real world.

\section{Is \approach better than baselines?}
\label{sec:appendix_comparison}

\mypara{Comparison Study on Other Datasets}
We outline the comparison study results from the Citeseer, Actor and Facebook datasets in \autoref{tab:exp_comparison_study_citeseer_256}, \autoref{tab:exp_comparison_study_actor_256} and \autoref{tab:exp_comparison_study_facebook_256} respectively. 
The node embedding size is fixed to 256.

\begin{table*}
\centering
\caption{Comparison of all baseline methods and \approach. We use the Citeseer dataset and the node embedding size is fixed to 256.}
\resizebox{0.6\linewidth}{!} {
\begin{tabular}{ccccccccc}
\toprule
\multirow{2}{*}{\textbf{\begin{tabular}[c]{@{}c@{}}Graph\\ Reocvery\\ Method\end{tabular}}} &
  \multirow{2}{*}{$f$} &
  \multicolumn{4}{c}{\textbf{Edge Metric}} &
  \multicolumn{3}{c}{\textbf{Global Metric}} \\
\cmidrule(lr){3-6} \cmidrule(lr){7-9}
 &
   &
  \multicolumn{1}{c}{\textbf{Precision}} &
  \multicolumn{1}{c}{\textbf{Recall}} &
  \multicolumn{1}{c}{\textbf{F1}} &
  \multicolumn{1}{c}{\textbf{JDD}} &
  \multicolumn{1}{c}{\textbf{\begin{tabular}[c]{@{}c@{}}Frobenius\\ Error\end{tabular}}} &
  \multicolumn{1}{c}{\textbf{\begin{tabular}[c]{@{}c@{}}Triangle\\ Error\end{tabular}}} &
  \multicolumn{1}{c}{\textbf{\begin{tabular}[c]{@{}c@{}}Clustering Coef.\\ Error\end{tabular}}} \\
\midrule
\multirow{4}{*}{\textbf{\begin{tabular}[c]{@{}c@{}}Direct\\ Recovery\end{tabular}}}  & DW & 0.001$\pm$0.000 & 0.002$\pm$0.000 & 0.001$\pm$0.000 & 0.000$\pm$0.000 & 1.647$\pm$0.000 & 6.867$\pm$0.000 & 0.753$\pm$0.000 \\
         & N2V & 0.003$\pm$0.000 & 0.006$\pm$0.000 & 0.004$\pm$0.000 & 0.000$\pm$0.000 & 1.645$\pm$0.000 & 7.559$\pm$0.000 & 0.759$\pm$0.000 \\
         & NetSMF & 0.013$\pm$0.000 & 0.022$\pm$0.000 & 0.016$\pm$0.000 & 0.311$\pm$0.000 & 1.647$\pm$0.000 & 0.621$\pm$0.000 & 0.228$\pm$0.000 \\
         & GCN & 0.001$\pm$0.000 & 0.002$\pm$0.000 & 0.002$\pm$0.000 & 0.000$\pm$0.000 & 1.647$\pm$0.000 & 6.391$\pm$0.000 & 0.653$\pm$0.000 \\
\midrule
\multirow{4}{*}{\textbf{\begin{tabular}[c]{@{}c@{}}Invert\\ Embedding\end{tabular}}} & DW & 0.007$\pm$0.005 & 0.012$\pm$0.010 & 0.009$\pm$0.007 & 0.667$\pm$0.092 & 1.660$\pm$0.010 & 0.866$\pm$0.661 & 0.198$\pm$0.008 \\
         & N2V & 0.021$\pm$0.008 & 0.037$\pm$0.014 & 0.027$\pm$0.010 & 0.665$\pm$0.030 & 1.645$\pm$0.008 & 1.869$\pm$0.160 & 0.148$\pm$0.005 \\
         & NetSMF & 0.003$\pm$0.001 & 0.005$\pm$0.003 & 0.004$\pm$0.002 & 0.462$\pm$0.064 & 1.676$\pm$0.007 & 0.842$\pm$0.789 & 0.221$\pm$0.011 \\
         & GCN & 0.015$\pm$0.003 & 0.026$\pm$0.006 & 0.019$\pm$0.004 & 0.675$\pm$0.004 & 1.657$\pm$0.005 & 0.255$\pm$0.158 & 0.188$\pm$0.005 \\
\midrule
\multirow{4}{*}{\textbf{\begin{tabular}[c]{@{}c@{}}$k$NN\\ Graph\end{tabular}}} & DW & 0.338$\pm$0.000 & 0.483$\pm$0.000 & 0.398$\pm$0.000 & 0.234$\pm$0.000 & 1.210$\pm$0.000 & 4.528$\pm$0.000 & 0.421$\pm$0.000 \\
         & N2V & 0.350$\pm$0.000 & 0.500$\pm$0.000 & 0.412$\pm$0.000 & 0.237$\pm$0.000 & 1.196$\pm$0.000 & 3.399$\pm$0.000 & 0.416$\pm$0.000 \\
         & NetSMF & 0.431$\pm$0.000 & 0.616$\pm$0.000 & 0.507$\pm$0.000 & 0.225$\pm$0.000 & 1.094$\pm$0.000 & 3.420$\pm$0.000 & 0.488$\pm$0.000 \\
         & GCN & 0.343$\pm$0.000 & 0.490$\pm$0.000 & 0.403$\pm$0.000 & 0.225$\pm$0.000 & 1.204$\pm$0.000 & 3.201$\pm$0.000 & 0.397$\pm$0.000 \\
\midrule
\multirow{4}{*}{\approach}  & DW & 0.403$\pm$0.002 & 0.555$\pm$0.005 & 0.467$\pm$0.003 & 0.617$\pm$0.011 & 1.125$\pm$0.003 & 1.877$\pm$0.075 & 0.341$\pm$0.009 \\
         & N2V & 0.445$\pm$0.001 & 0.575$\pm$0.002 & 0.502$\pm$0.001 & 0.506$\pm$0.007 & 1.069$\pm$0.002 & 1.734$\pm$0.039 & 0.357$\pm$0.005 \\
         & NetSMF & 0.530$\pm$0.002 & 0.672$\pm$0.001 & 0.592$\pm$0.001 & 0.461$\pm$0.005 & 0.961$\pm$0.002 & 2.001$\pm$0.056 & 0.432$\pm$0.003 \\
         & GCN & 0.414$\pm$0.003 & 0.529$\pm$0.002 & 0.465$\pm$0.001 & 0.527$\pm$0.011 & 1.105$\pm$0.004 & 1.467$\pm$0.057 & 0.330$\pm$0.004 \\
\bottomrule
\end{tabular}
}
\label{tab:exp_comparison_study_citeseer_256}
\end{table*}

\begin{table*}
\centering
\caption{Comparison of all baseline methods and \approach. We use the Actor dataset and the node embedding size is fixed to 256.}
\resizebox{0.6\linewidth}{!} {
\begin{tabular}{ccccccccc}
\toprule
\multirow{2}{*}{\textbf{\begin{tabular}[c]{@{}c@{}}Graph\\ Reocvery\\ Method\end{tabular}}} &
  \multirow{2}{*}{$f$} &
  \multicolumn{4}{c}{\textbf{Edge Metric}} &
  \multicolumn{3}{c}{\textbf{Global Metric}} \\
\cmidrule(lr){3-6} \cmidrule(lr){7-9}
 &
   &
  \multicolumn{1}{c}{\textbf{Precision}} &
  \multicolumn{1}{c}{\textbf{Recall}} &
  \multicolumn{1}{c}{\textbf{F1}} &
  \multicolumn{1}{c}{\textbf{JDD}} &
  \multicolumn{1}{c}{\textbf{\begin{tabular}[c]{@{}c@{}}Frobenius\\ Error\end{tabular}}} &
  \multicolumn{1}{c}{\textbf{\begin{tabular}[c]{@{}c@{}}Triangle\\ Error\end{tabular}}} &
  \multicolumn{1}{c}{\textbf{\begin{tabular}[c]{@{}c@{}}Clustering Coef.\\ Error\end{tabular}}} \\
\midrule
\multirow{4}{*}{\textbf{\begin{tabular}[c]{@{}c@{}}Direct\\ Recovery\end{tabular}}}  & DW & 0.001$\pm$0.000 & 0.002$\pm$0.000 & 0.001$\pm$0.000 & 0.000$\pm$0.000 & 1.647$\pm$0.000 & 6.867$\pm$0.000 & 0.753$\pm$0.000 \\
         & N2V & 0.003$\pm$0.000 & 0.006$\pm$0.000 & 0.004$\pm$0.000 & 0.000$\pm$0.000 & 1.645$\pm$0.000 & 7.559$\pm$0.000 & 0.759$\pm$0.000 \\
         & NetSMF & 0.013$\pm$0.000 & 0.022$\pm$0.000 & 0.016$\pm$0.000 & 0.311$\pm$0.000 & 1.647$\pm$0.000 & 0.621$\pm$0.000 & 0.228$\pm$0.000 \\
         & GCN & 0.001$\pm$0.000 & 0.002$\pm$0.000 & 0.002$\pm$0.000 & 0.000$\pm$0.000 & 1.647$\pm$0.000 & 6.391$\pm$0.000 & 0.653$\pm$0.000 \\
\midrule
\multirow{4}{*}{\textbf{\begin{tabular}[c]{@{}c@{}}Invert\\ Embedding\end{tabular}}} & DW & 0.007$\pm$0.005 & 0.012$\pm$0.010 & 0.009$\pm$0.007 & 0.667$\pm$0.092 & 1.660$\pm$0.010 & 0.866$\pm$0.661 & 0.198$\pm$0.008 \\
         & N2V & 0.021$\pm$0.008 & 0.037$\pm$0.014 & 0.027$\pm$0.010 & 0.665$\pm$0.030 & 1.645$\pm$0.008 & 1.869$\pm$0.160 & 0.148$\pm$0.005 \\
         & NetSMF & 0.003$\pm$0.001 & 0.005$\pm$0.003 & 0.004$\pm$0.002 & 0.462$\pm$0.064 & 1.676$\pm$0.007 & 0.842$\pm$0.789 & 0.221$\pm$0.011 \\
         & GCN & 0.015$\pm$0.003 & 0.026$\pm$0.006 & 0.019$\pm$0.004 & 0.675$\pm$0.004 & 1.657$\pm$0.005 & 0.255$\pm$0.158 & 0.188$\pm$0.005 \\
\midrule
\multirow{4}{*}{\textbf{\begin{tabular}[c]{@{}c@{}}$k$NN\\ Graph\end{tabular}}} & DW & 0.562$\pm$0.000 & 0.400$\pm$0.000 & 0.468$\pm$0.000 & 0.263$\pm$0.000 & 0.955$\pm$0.000 & 0.308$\pm$0.000 & 0.246$\pm$0.000 \\
         & N2V & 0.343$\pm$0.000 & 0.244$\pm$0.000 & 0.285$\pm$0.000 & 0.366$\pm$0.000 & 1.106$\pm$0.000 & 0.566$\pm$0.000 & 0.211$\pm$0.000 \\
         & NetSMF & 0.453$\pm$0.000 & 0.322$\pm$0.000 & 0.376$\pm$0.000 & 0.325$\pm$0.000 & 1.033$\pm$0.000 & 0.905$\pm$0.000 & 0.310$\pm$0.000 \\
         & GCN & 0.304$\pm$0.000 & 0.217$\pm$0.000 & 0.253$\pm$0.000 & 0.349$\pm$0.000 & 1.131$\pm$0.000 & 0.629$\pm$0.000 & 0.202$\pm$0.000 \\
\midrule
\multirow{4}{*}{\approach}  & DW & 0.687$\pm$0.001 & 0.435$\pm$0.002 & 0.533$\pm$0.002 & 0.417$\pm$0.001 & 0.874$\pm$0.001 & 0.203$\pm$0.009 & 0.229$\pm$0.002 \\
         & N2V & 0.465$\pm$0.001 & 0.313$\pm$0.000 & 0.374$\pm$0.000 & 0.473$\pm$0.003 & 1.023$\pm$0.001 & 0.179$\pm$0.007 & 0.176$\pm$0.001 \\
         & NetSMF & 0.562$\pm$0.002 & 0.366$\pm$0.001 & 0.443$\pm$0.001 & 0.457$\pm$0.003 & 0.959$\pm$0.001 & 0.147$\pm$0.013 & 0.285$\pm$0.002 \\
         & GCN & 0.373$\pm$0.001 & 0.263$\pm$0.000 & 0.308$\pm$0.001 & 0.505$\pm$0.003 & 1.086$\pm$0.001 & 0.280$\pm$0.008 & 0.153$\pm$0.002 \\
\bottomrule
\end{tabular}
}
\label{tab:exp_comparison_study_actor_256}
\end{table*}

\begin{table*}[t]
\centering
\caption{Comparison of all baseline methods and \approach. We use the Facebook dataset and the node embedding size is fixed to 256.}
\resizebox{0.6\linewidth}{!} {
\begin{tabular}{ccccccccc}
\toprule
\multirow{2}{*}{\textbf{\begin{tabular}[c]{@{}c@{}}Graph\\ Reocvery\\ Method\end{tabular}}} &
  \multirow{2}{*}{$f$} &
  \multicolumn{4}{c}{\textbf{Edge Metric}} &
  \multicolumn{3}{c}{\textbf{Global Metric}} \\
\cmidrule(lr){3-6} \cmidrule(lr){7-9}
 &
   &
  \multicolumn{1}{c}{\textbf{Precision}} &
  \multicolumn{1}{c}{\textbf{Recall}} &
  \multicolumn{1}{c}{\textbf{F1}} &
  \multicolumn{1}{c}{\textbf{JDD}} &
  \multicolumn{1}{c}{\textbf{\begin{tabular}[c]{@{}c@{}}Frobenius\\ Error\end{tabular}}} &
  \multicolumn{1}{c}{\textbf{\begin{tabular}[c]{@{}c@{}}Triangle\\ Error\end{tabular}}} &
  \multicolumn{1}{c}{\textbf{\begin{tabular}[c]{@{}c@{}}Clustering Coef.\\ Error\end{tabular}}} \\
\midrule
\multirow{4}{*}{\textbf{\begin{tabular}[c]{@{}c@{}}Direct\\ Recovery\end{tabular}}}  & DW & 0.001$\pm$0.000 & 0.002$\pm$0.000 & 0.001$\pm$0.000 & 0.000$\pm$0.000 & 1.647$\pm$0.000 & 6.867$\pm$0.000 & 0.753$\pm$0.000 \\
         & N2V & 0.003$\pm$0.000 & 0.006$\pm$0.000 & 0.004$\pm$0.000 & 0.000$\pm$0.000 & 1.645$\pm$0.000 & 7.559$\pm$0.000 & 0.759$\pm$0.000 \\
         & NetSMF & 0.013$\pm$0.000 & 0.022$\pm$0.000 & 0.016$\pm$0.000 & 0.311$\pm$0.000 & 1.647$\pm$0.000 & 0.621$\pm$0.000 & 0.228$\pm$0.000 \\
         & GCN & 0.001$\pm$0.000 & 0.002$\pm$0.000 & 0.002$\pm$0.000 & 0.000$\pm$0.000 & 1.647$\pm$0.000 & 6.391$\pm$0.000 & 0.653$\pm$0.000 \\
\midrule
\multirow{4}{*}{\textbf{\begin{tabular}[c]{@{}c@{}}Invert\\ Embedding\end{tabular}}} & DW & 0.007$\pm$0.005 & 0.012$\pm$0.010 & 0.009$\pm$0.007 & 0.667$\pm$0.092 & 1.660$\pm$0.010 & 0.866$\pm$0.661 & 0.198$\pm$0.008 \\
         & N2V & 0.021$\pm$0.008 & 0.037$\pm$0.014 & 0.027$\pm$0.010 & 0.665$\pm$0.030 & 1.645$\pm$0.008 & 1.869$\pm$0.160 & 0.148$\pm$0.005 \\
         & NetSMF & 0.003$\pm$0.001 & 0.005$\pm$0.003 & 0.004$\pm$0.002 & 0.462$\pm$0.064 & 1.676$\pm$0.007 & 0.842$\pm$0.789 & 0.221$\pm$0.011 \\
         & GCN & 0.015$\pm$0.003 & 0.026$\pm$0.006 & 0.019$\pm$0.004 & 0.675$\pm$0.004 & 1.657$\pm$0.005 & 0.255$\pm$0.158 & 0.188$\pm$0.005 \\
\midrule
\multirow{4}{*}{\textbf{\begin{tabular}[c]{@{}c@{}}$k$NN\\ Graph\end{tabular}}} & DW & 0.429$\pm$0.000 & 0.442$\pm$0.000 & 0.436$\pm$0.000 & 0.189$\pm$0.000 & 1.070$\pm$0.000 & 0.281$\pm$0.000 & 0.071$\pm$0.000 \\
         & N2V & 0.460$\pm$0.000 & 0.474$\pm$0.000 & 0.467$\pm$0.000 & 0.172$\pm$0.000 & 1.040$\pm$0.000 & 0.046$\pm$0.000 & 0.040$\pm$0.000 \\
         & NetSMF & 0.444$\pm$0.000 & 0.457$\pm$0.000 & 0.450$\pm$0.000 & 0.291$\pm$0.000 & 1.056$\pm$0.000 & 0.006$\pm$0.000 & 0.040$\pm$0.000 \\
         & GCN & 0.322$\pm$0.000 & 0.331$\pm$0.000 & 0.327$\pm$0.000 & 0.183$\pm$0.000 & 1.169$\pm$0.000 & 0.072$\pm$0.000 & 0.087$\pm$0.000 \\
\midrule
\multirow{4}{*}{\approach}  & DW & 0.441$\pm$0.001 & 0.471$\pm$0.001 & 0.456$\pm$0.001 & 0.519$\pm$0.006 & 1.061$\pm$0.001 & 0.494$\pm$0.002 & 0.077$\pm$0.001 \\
         & N2V & 0.468$\pm$0.000 & 0.487$\pm$0.001 & 0.477$\pm$0.001 & 0.444$\pm$0.002 & 1.033$\pm$0.001 & 0.545$\pm$0.001 & 0.090$\pm$0.001 \\
         & NetSMF & 0.454$\pm$0.001 & 0.502$\pm$0.002 & 0.476$\pm$0.001 & 0.457$\pm$0.002 & 1.050$\pm$0.001 & 0.424$\pm$0.007 & 0.081$\pm$0.001 \\
         & GCN & 0.342$\pm$0.001 & 0.364$\pm$0.001 & 0.352$\pm$0.001 & 0.371$\pm$0.004 & 1.157$\pm$0.001 & 0.452$\pm$0.002 & 0.056$\pm$0.001 \\
\bottomrule
\end{tabular}
}
\label{tab:exp_comparison_study_facebook_256}
\end{table*}

\mypara{Visual Explanation}
We use bitmap images to further exemplify \approach's capability in recovery graphs from the node embedding matrices. 
Each adjacency matrix $\mathbf{A}$ is represented as an image where the pixel at a coordinate $(i, j)$ is blue if $\mathbf{A}_{ij}=1$ and white otherwise. 
In this way, bitmap images can give an overall impression of the graph topology, offering a straightforward qualitative visual assessment between the original graphs and the recovered graphs.
We visualize the original graph and the recovered graphs by both baseline methods and \approach from the node embeddings generated by Node2Vec in \autoref{fig:visualization}.
As we can see in \autoref{fig:visualization}, the graph structures recovered by direct recovery and invert embedding do not resemble the original graph. 
At the same time, $k$NN graph can recover some graph topology.
However, because $k$NN is not learning-based and uses a predefined distance function, we can see in \autoref{fig:visualization} that the recovered graph by $k$NN is noisy.
Thanks to the learnable distance function and adaptive graph structure combination (see \autoref{sec:approach}), \approach can reduce a reasonable amount of false edges and recover a better graph topology, consequently leading to better performance as shown in \autoref{tab:exp_comparison_study_cora_256}.

\section{How Effective is \approach?}
\label{sec:appendix_approach_effectivness}

\mypara{Additional Experimental Results}
We list additional experimental results using all four datasets in \autoref{tab:attack_128_results} and \autoref{tab:attack_64_results}.

\begin{table*}[t]
\centering
\caption{
The performance results of \approach using all four datasets. 
We fix the node embedding size to 128. 
We show the relative improvement scores in edge metrics to demonstrate to what extent \approach can relatively improve from $k$NN graph.
We add a positive sign (+) next to the relative improvement score to highlight the improvement.
We also show the relative error reduction scores in global metrics to demonstrate to what extent \approach can relatively reduce errors incurred by $k$NN graph.
We add a negative sign (-) next to the relative error reduction score to highlight the difference.
}
\resizebox{0.75\linewidth}{!} {
\begin{tabular}{ccccccccc}
\toprule
\multirow{2}{*}{\textbf{Dataset}} &
  \multirow{2}{*}{$f$} &
  \multicolumn{4}{c}{\textbf{Edge Metrics}} &
  \multicolumn{3}{c}{\textbf{Global Metrics (Relative Error)}} \\
\cmidrule(lr){3-6} \cmidrule(lr){7-9}
 &
   &
  \multicolumn{1}{c}{\textbf{Precision}} &
  \multicolumn{1}{c}{\textbf{Recall}} &
  \multicolumn{1}{c}{\textbf{F1}} &
  \multicolumn{1}{c}{\textbf{\begin{tabular}[c]{@{}c@{}}Deg. \\ Dist.\end{tabular}}} &
  \multicolumn{1}{c}{\textbf{\begin{tabular}[c]{@{}c@{}}Frobenius\\ Error\end{tabular}}} &
  \multicolumn{1}{c}{\textbf{\begin{tabular}[c]{@{}c@{}}Triangle\\ Error\end{tabular}}} &
  \multicolumn{1}{c}{\textbf{\begin{tabular}[c]{@{}c@{}}Clus. Coef.\\ Error\end{tabular}}} \\
\midrule
\multirow{4}{*}{\textbf{Cora}} & DW & 0.570$\pm$0.001 (+0.224) & 0.633$\pm$0.005 (+0.109) & 0.600$\pm$0.003 (+0.170) & 0.786$\pm$0.012 (+1.311) & 0.919$\pm$0.002 (-0.122) & 0.987$\pm$0.039 (-1.376) & 0.232$\pm$0.004 (-0.055) \\
         & N2V & 0.504$\pm$0.002 (+0.276) & 0.554$\pm$0.003 (+0.145) & 0.528$\pm$0.002 (+0.214) & 0.726$\pm$0.010 (+1.205) & 0.995$\pm$0.002 (-0.126) & 0.997$\pm$0.035 (-1.140) & 0.223$\pm$0.004 (-0.050) \\
         & NetSMF & 0.516$\pm$0.003 (+0.225) & 0.579$\pm$0.003 (+0.121) & 0.545$\pm$0.003 (+0.175) & 0.716$\pm$0.006 (+1.188) & 0.982$\pm$0.004 (-0.111) & 1.453$\pm$0.033 (-1.139) & 0.297$\pm$0.004 (-0.047) \\
         & GCN & 0.456$\pm$0.002 (+0.231) & 0.499$\pm$0.003 (+0.101) & 0.476$\pm$0.001 (+0.170) & 0.763$\pm$0.007 (+1.305) & 1.048$\pm$0.003 (-0.101) & 0.806$\pm$0.039 (-1.235) & 0.213$\pm$0.004 (-0.049) \\
\midrule
\multirow{4}{*}{\textbf{Citeseer}} & DW & 0.488$\pm$0.001 (+0.204) & 0.639$\pm$0.003 (+0.104) & 0.553$\pm$0.001 (+0.159) & 0.576$\pm$0.006 (+1.482) & 1.016$\pm$0.000 (-0.112) & 1.772$\pm$0.058 (-2.132) & 0.362$\pm$0.006 (-0.071) \\
         & N2V & 0.453$\pm$0.001 (+0.270) & 0.583$\pm$0.003 (+0.141) & 0.510$\pm$0.002 (+0.214) & 0.500$\pm$0.004 (+1.155) & 1.058$\pm$0.001 (-0.129) & 1.715$\pm$0.021 (-1.581) & 0.353$\pm$0.003 (-0.057) \\
         & NetSMF & 0.458$\pm$0.001 (+0.197) & 0.599$\pm$0.003 (+0.095) & 0.519$\pm$0.001 (+0.154) & 0.455$\pm$0.006 (+1.014) & 1.053$\pm$0.002 (-0.102) & 2.319$\pm$0.064 (-1.266) & 0.429$\pm$0.003 (-0.055) \\
         & GCN & 0.404$\pm$0.003 (+0.201) & 0.519$\pm$0.001 (+0.079) & 0.454$\pm$0.002 (+0.146) & 0.518$\pm$0.005 (+1.270) & 1.117$\pm$0.004 (-0.095) & 1.404$\pm$0.025 (-1.641) & 0.316$\pm$0.005 (-0.061) \\
\midrule
\multirow{4}{*}{\textbf{Actor}} & DW & 0.678$\pm$0.002 (+0.225) & 0.424$\pm$0.001 (+0.078) & 0.521$\pm$0.002 (+0.133) & 0.400$\pm$0.002 (+0.564) & 0.882$\pm$0.001 (-0.079) & 0.219$\pm$0.009 (-0.032) & 0.223$\pm$0.002 (-0.016) \\
         & N2V & 0.461$\pm$0.001 (+0.348) & 0.307$\pm$0.001 (+0.264) & 0.369$\pm$0.001 (+0.298) & 0.473$\pm$0.003 (+0.296) & 1.026$\pm$0.001 (-0.081) & 0.209$\pm$0.006 (-0.330) & 0.168$\pm$0.002 (-0.037) \\
         & NetSMF & 0.498$\pm$0.002 (+0.221) & 0.330$\pm$0.002 (+0.135) & 0.397$\pm$0.002 (+0.172) & 0.472$\pm$0.003 (+0.409) & 1.001$\pm$0.001 (-0.062) & 0.148$\pm$0.013 (-0.828) & 0.270$\pm$0.002 (-0.033) \\
         & GCN & 0.374$\pm$0.002 (+0.220) & 0.263$\pm$0.001 (+0.206) & 0.309$\pm$0.001 (+0.212) & 0.512$\pm$0.002 (+0.481) & 1.085$\pm$0.001 (-0.044) & 0.303$\pm$0.007 (-0.263) & 0.150$\pm$0.001 (-0.043) \\
\midrule
\multirow{4}{*}{\textbf{Facebook}} & DW & 0.448$\pm$0.001 (+0.025) & 0.472$\pm$0.002 (+0.049) & 0.460$\pm$0.001 (+0.038) & 0.509$\pm$0.004 (+1.649) & 1.053$\pm$0.001 (-0.010) & 0.500$\pm$0.003 (+0.337) & 0.069$\pm$0.001 (-0.003) \\
         & N2V & 0.450$\pm$0.001 (+0.018) & 0.467$\pm$0.001 (+0.025) & 0.458$\pm$0.001 (+0.023) & 0.421$\pm$0.003 (+1.407) & 1.050$\pm$0.001 (-0.008) & 0.535$\pm$0.002 (+0.506) & 0.074$\pm$0.001 (+0.031) \\
         & NetSMF & 0.435$\pm$0.001 (+0.026) & 0.474$\pm$0.002 (+0.087) & 0.454$\pm$0.001 (+0.055) & 0.415$\pm$0.004 (+0.579) & 1.069$\pm$0.000 (-0.007) & 0.417$\pm$0.004 (+0.407) & 0.047$\pm$0.001 (-0.002) \\
         & GCN & 0.341$\pm$0.001 (+0.062) & 0.364$\pm$0.002 (+0.100) & 0.352$\pm$0.001 (+0.080) & 0.370$\pm$0.004 (+0.851) & 1.157$\pm$0.001 (-0.013) & 0.450$\pm$0.004 (+0.377) & 0.059$\pm$0.000 (-0.026) \\
\bottomrule\\
\end{tabular}
}
\label{tab:attack_128_results}
\end{table*}

\begin{table*}[t]
\centering
\caption{
The performance results of \approach using all four datasets. 
We fix the node embedding size to 64. 
We show the relative improvement scores in edge metrics to demonstrate to what extent \approach can relatively improve from $k$NN graph.
We add a positive sign (+) next to the relative improvement score to highlight the improvement.
We also show the relative error reduction scores in global metrics to demonstrate to what extent \approach can relatively reduce errors incurred by $k$NN graph.
We add a negative sign (-) next to the relative error reduction score to highlight the difference.
}
\resizebox{0.75\linewidth}{!} {
\begin{tabular}{ccccccccc}
\toprule
\multirow{2}{*}{\textbf{Dataset}} &
  \multirow{2}{*}{$f$} &
  \multicolumn{4}{c}{\textbf{Edge Metrics}} &
  \multicolumn{3}{c}{\textbf{Global Metrics (Relative Error)}} \\
\cmidrule(lr){3-6} \cmidrule(lr){7-9}
 &
   &
  \multicolumn{1}{c}{\textbf{Precision}} &
  \multicolumn{1}{c}{\textbf{Recall}} &
  \multicolumn{1}{c}{\textbf{F1}} &
  \multicolumn{1}{c}{\textbf{\begin{tabular}[c]{@{}c@{}}Deg. \\ Dist.\end{tabular}}} &
  \multicolumn{1}{c}{\textbf{\begin{tabular}[c]{@{}c@{}}Frobenius\\ Error\end{tabular}}} &
  \multicolumn{1}{c}{\textbf{\begin{tabular}[c]{@{}c@{}}Triangle\\ Error\end{tabular}}} &
  \multicolumn{1}{c}{\textbf{\begin{tabular}[c]{@{}c@{}}Clus. Coef.\\ Error\end{tabular}}} \\
\midrule
\multirow{4}{*}{\textbf{Cora}} & DW & 0.596$\pm$0.003 (+0.231) & 0.636$\pm$0.003 (+0.073) & 0.615$\pm$0.001 (+0.155) & 0.729$\pm$0.015 (+1.195) & 0.892$\pm$0.002 (-0.128) & 0.918$\pm$0.044 (-1.137) & 0.248$\pm$0.003 (-0.046) \\
         & N2V & 0.467$\pm$0.002 (+0.270) & 0.524$\pm$0.002 (+0.162) & 0.494$\pm$0.002 (+0.220) & 0.730$\pm$0.007 (+1.231) & 1.036$\pm$0.002 (-0.115) & 1.046$\pm$0.017 (-1.156) & 0.218$\pm$0.004 (-0.048) \\
         & NetSMF & 0.441$\pm$0.003 (+0.166) & 0.532$\pm$0.002 (+0.149) & 0.482$\pm$0.001 (+0.159) & 0.727$\pm$0.009 (+1.237) & 1.069$\pm$0.004 (-0.071) & 2.193$\pm$0.169 (-0.512) & 0.295$\pm$0.004 (-0.045) \\
         & GCN & 0.426$\pm$0.003 (+0.217) & 0.470$\pm$0.002 (+0.097) & 0.447$\pm$0.003 (+0.161) & 0.777$\pm$0.007 (+1.306) & 1.079$\pm$0.004 (-0.091) & 0.684$\pm$0.032 (-1.235) & 0.174$\pm$0.004 (-0.054) \\
\midrule
\multirow{4}{*}{\textbf{Citeseer}} & DW & 0.541$\pm$0.002 (+0.221) & 0.681$\pm$0.003 (+0.074) & 0.603$\pm$0.002 (+0.155) & 0.516$\pm$0.002 (+1.295) & 0.947$\pm$0.003 (-0.131) & 1.669$\pm$0.036 (-1.618) & 0.388$\pm$0.004 (-0.065) \\
         & N2V & 0.423$\pm$0.004 (+0.275) & 0.557$\pm$0.004 (+0.172) & 0.481$\pm$0.004 (+0.231) & 0.479$\pm$0.004 (+1.028) & 1.096$\pm$0.005 (-0.121) & 1.763$\pm$0.038 (-1.469) & 0.335$\pm$0.004 (-0.055) \\
         & NetSMF & 0.362$\pm$0.004 (+0.049) & 0.567$\pm$0.003 (+0.153) & 0.442$\pm$0.004 (+0.091) & 0.462$\pm$0.004 (+1.000) & 1.197$\pm$0.007 (-0.005) & 7.019$\pm$0.920 (+3.300) & 0.438$\pm$0.004 (-0.039) \\
         & GCN & 0.389$\pm$0.003 (+0.196) & 0.501$\pm$0.002 (+0.077) & 0.438$\pm$0.002 (+0.143) & 0.524$\pm$0.010 (+1.297) & 1.134$\pm$0.004 (-0.091) & 1.435$\pm$0.030 (-1.586) & 0.314$\pm$0.002 (-0.054) \\
\midrule
\multirow{4}{*}{\textbf{Actor}} & DW & 0.660$\pm$0.002 (+0.224) & 0.414$\pm$0.001 (+0.080) & 0.508$\pm$0.001 (+0.135) & 0.401$\pm$0.002 (+0.520) & 0.894$\pm$0.001 (-0.078) & 0.212$\pm$0.003 (-0.019) & 0.224$\pm$0.003 (-0.013) \\
         & N2V & 0.447$\pm$0.001 (+0.330) & 0.298$\pm$0.001 (+0.246) & 0.357$\pm$0.001 (+0.279) & 0.468$\pm$0.001 (+0.283) & 1.035$\pm$0.001 (-0.076) & 0.231$\pm$0.010 (-0.252) & 0.159$\pm$0.002 (-0.034) \\
         & NetSMF & 0.402$\pm$0.002 (+0.186) & 0.274$\pm$0.001 (+0.135) & 0.326$\pm$0.001 (+0.155) & 0.480$\pm$0.005 (+0.370) & 1.065$\pm$0.001 (-0.044) & 0.238$\pm$0.089 (-0.566) & 0.228$\pm$0.002 (-0.031) \\
         & GCN & 0.338$\pm$0.001 (+0.211) & 0.239$\pm$0.001 (+0.209) & 0.280$\pm$0.001 (+0.209) & 0.523$\pm$0.003 (+0.460) & 1.109$\pm$0.001 (-0.038) & 0.325$\pm$0.008 (-0.192) & 0.135$\pm$0.002 (-0.045) \\
\midrule
\multirow{4}{*}{\textbf{Facebook}} & DW & 0.445$\pm$0.001 (+0.025) & 0.468$\pm$0.001 (+0.047) & 0.456$\pm$0.001 (+0.037) & 0.473$\pm$0.004 (+1.688) & 1.056$\pm$0.001 (-0.010) & 0.498$\pm$0.002 (+0.435) & 0.062$\pm$0.001 (-0.007) \\
         & N2V & 0.418$\pm$0.001 (+0.022) & 0.436$\pm$0.001 (+0.033) & 0.427$\pm$0.001 (+0.028) & 0.385$\pm$0.003 (+1.253) & 1.082$\pm$0.001 (-0.007) & 0.495$\pm$0.001 (+0.422) & 0.044$\pm$0.001 (-0.016) \\
         & NetSMF & 0.427$\pm$0.001 (+0.082) & 0.522$\pm$0.002 (+0.283) & 0.470$\pm$0.001 (+0.172) & 0.394$\pm$0.002 (+0.958) & 1.085$\pm$0.001 (-0.018) & 0.038$\pm$0.022 (+0.014) & 0.052$\pm$0.000 (-0.029) \\
         & GCN & 0.342$\pm$0.001 (+0.062) & 0.371$\pm$0.002 (+0.122) & 0.356$\pm$0.001 (+0.093) & 0.372$\pm$0.003 (+0.876) & 1.159$\pm$0.001 (-0.010) & 0.432$\pm$0.003 (+0.350) & 0.059$\pm$0.001 (-0.025) \\
\bottomrule\\
\end{tabular}
}
\label{tab:attack_64_results}
\end{table*}

\section{How does $k$ Affect the Attack Performance}
\label{appendix:exp_impact_of_k}

\begin{figure*}[t]
    \begin{subfigure}[b]{0.23\textwidth}

        \includegraphics[width=\linewidth]{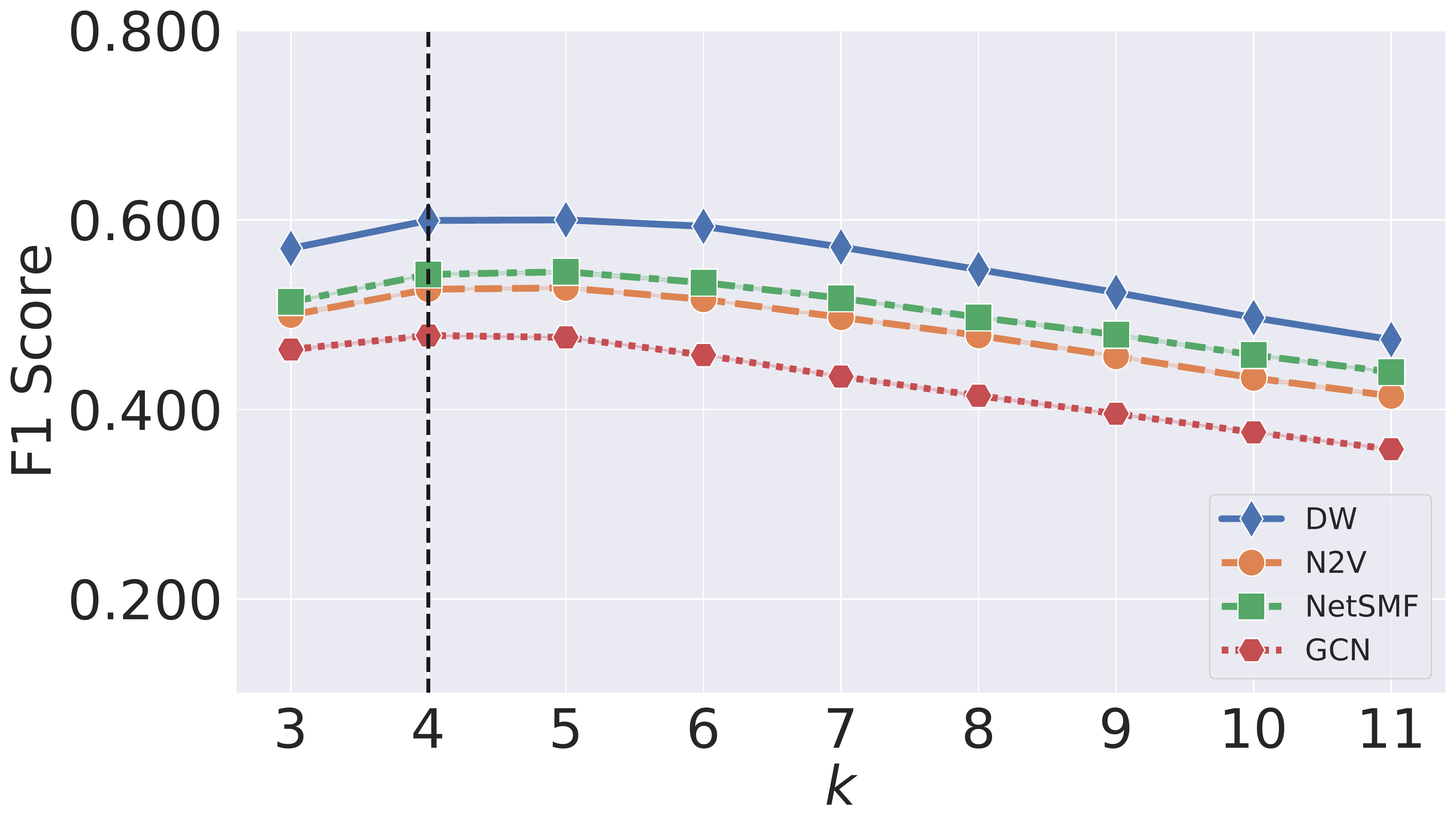}
        \caption{Cora}
        \label{fig:citation_network_k_128_cora}
    \end{subfigure} %
    \hfill
    \begin{subfigure}[b]{0.23\textwidth}

        \includegraphics[width=\linewidth]{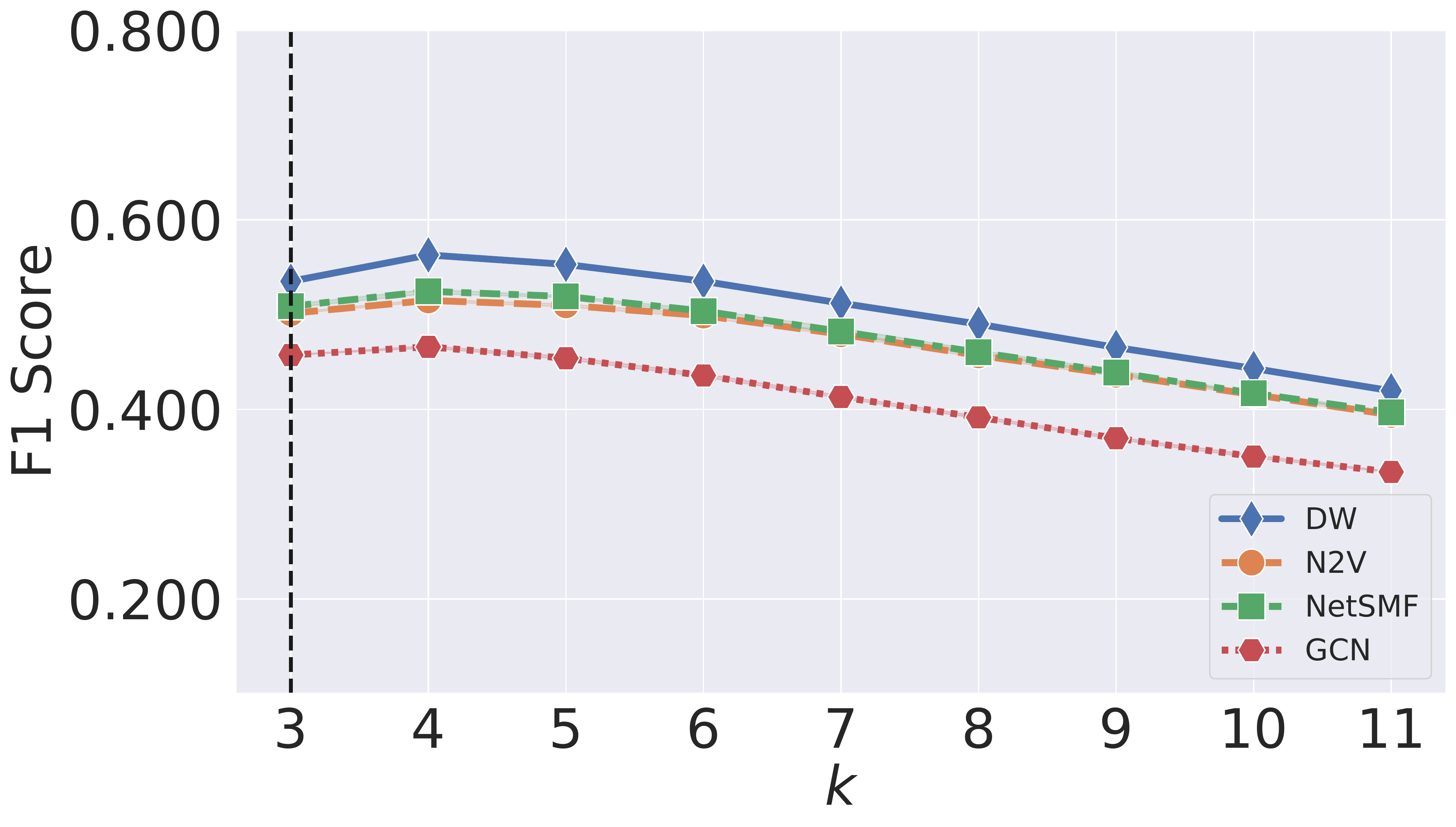}
        \caption{Citeseer}
        \label{fig:social_network_k_128_citeseer}
    \end{subfigure} %
    \hfill
    \begin{subfigure}[b]{0.23\textwidth}

        \includegraphics[width=\linewidth]{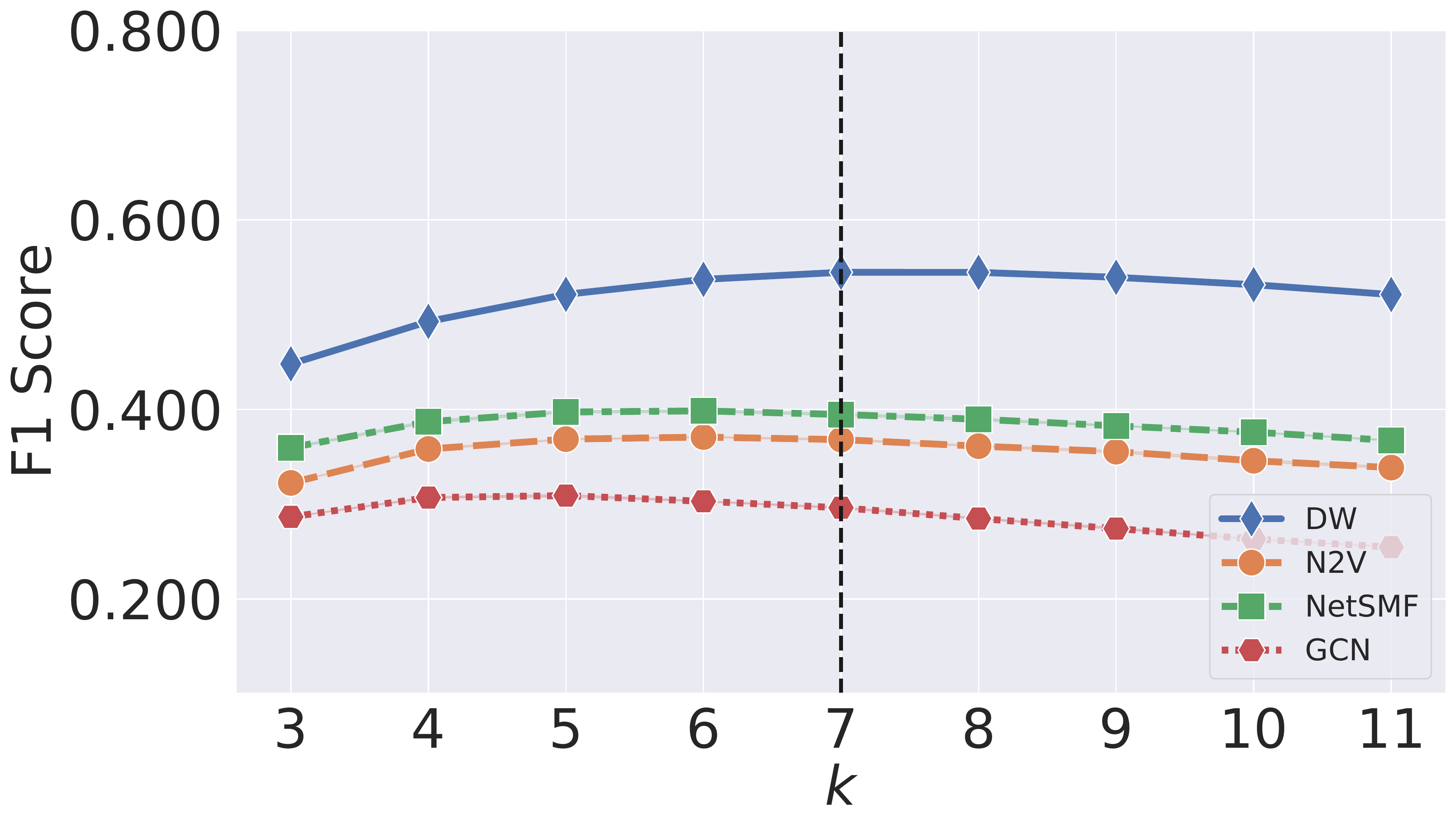}
        \caption{Actor}
        \label{fig:citation_network_k_128_actor}
    \end{subfigure} %
    \hfill
    \begin{subfigure}[b]{0.23\textwidth}

        \includegraphics[width=\linewidth]{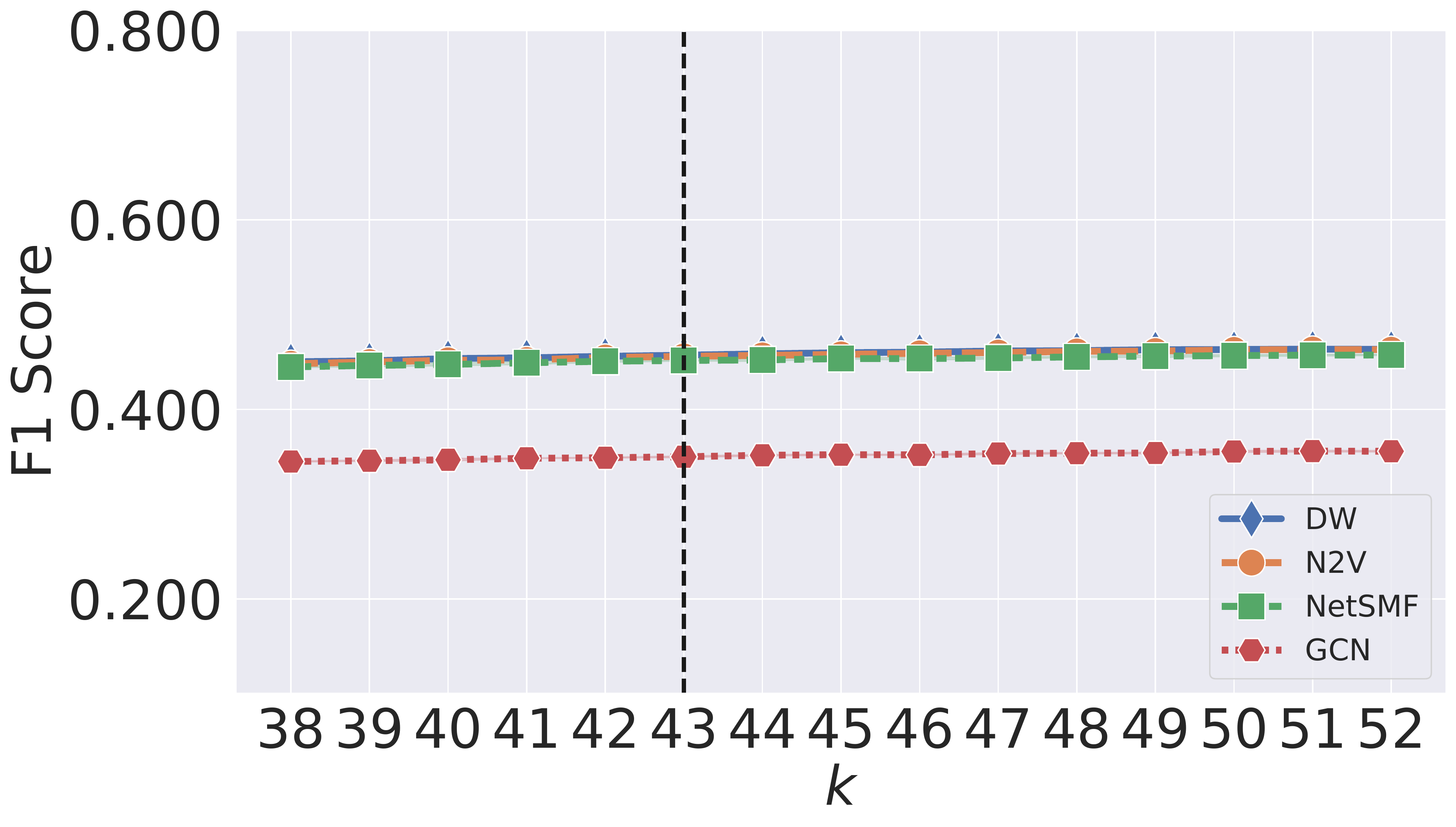}
        \caption{Facebook}
        \label{fig:social_network_k_128_facebook}
    \end{subfigure}%

    \begin{subfigure}[b]{0.23\textwidth}
        \includegraphics[width=\linewidth]{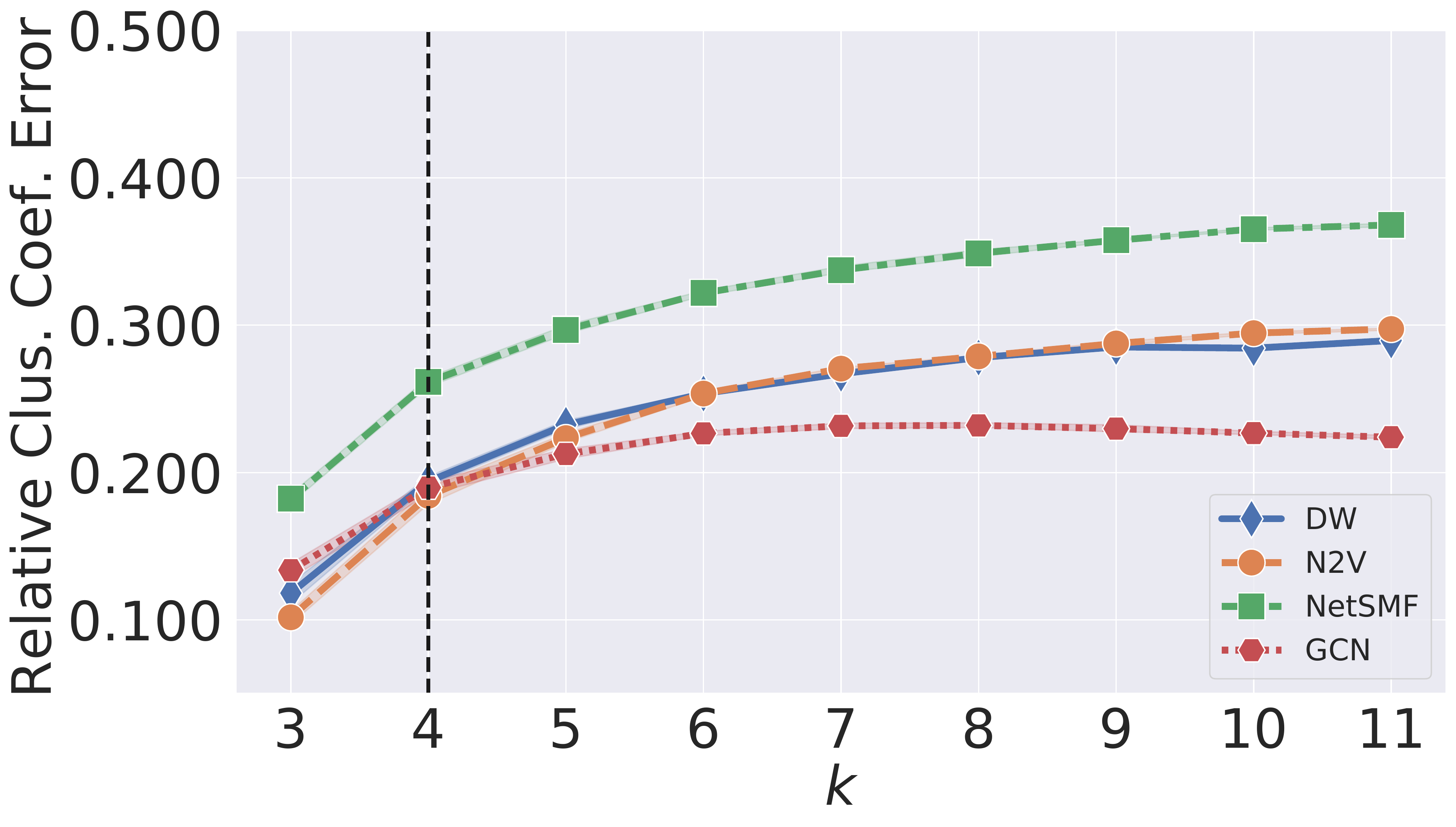}
        \caption{Cora}
        \label{fig:citation_network_k_128_cora_clus}
    \end{subfigure}%
    \hfill
    \begin{subfigure}[b]{0.23\textwidth}
        \includegraphics[width=\linewidth]{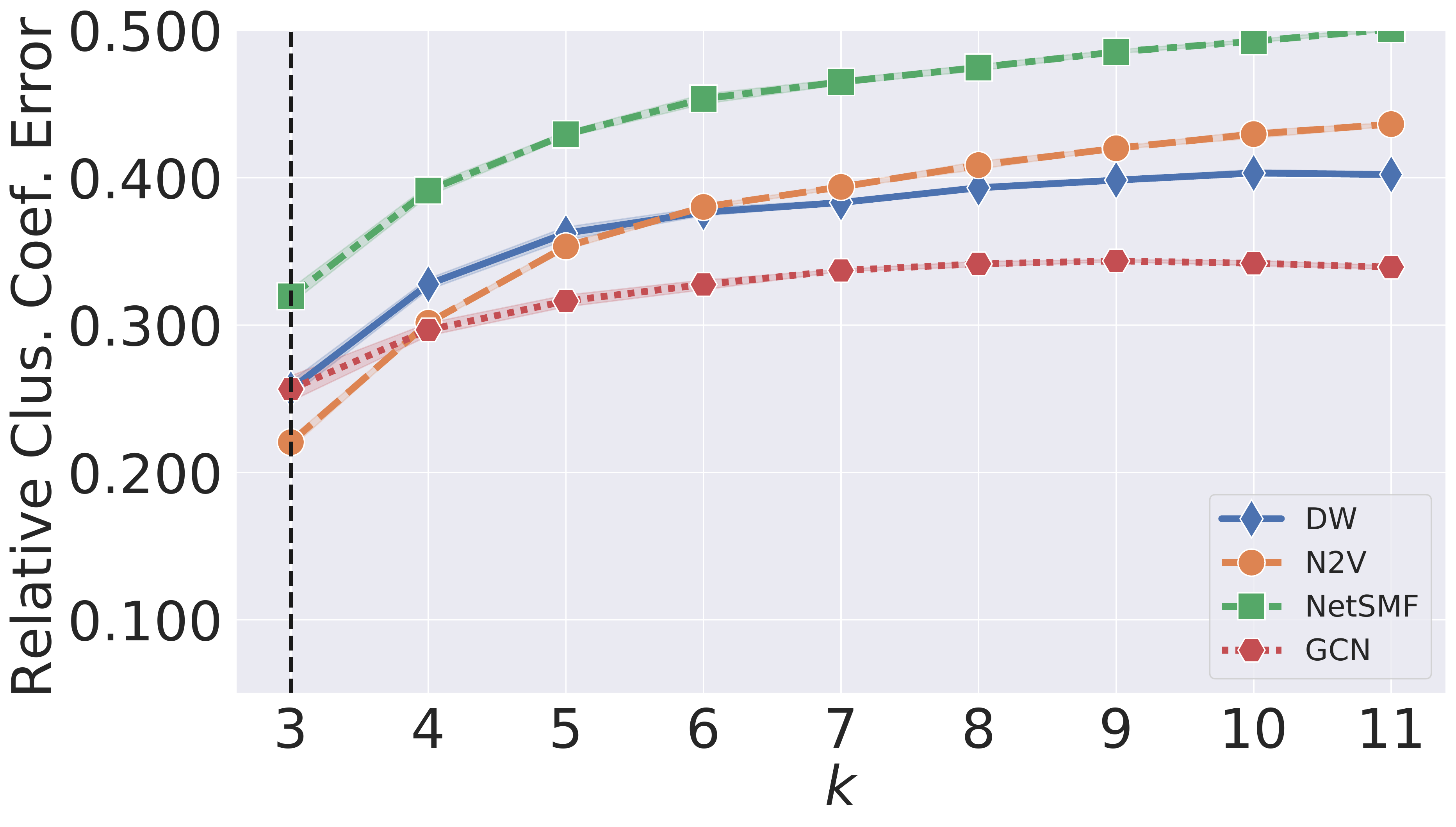}
        \caption{Citeseer}
        \label{fig:social_network_k_128_citeseer_clus}
    \end{subfigure}%
    \hfill
    \begin{subfigure}[b]{0.23\textwidth}
        \includegraphics[width=\linewidth]{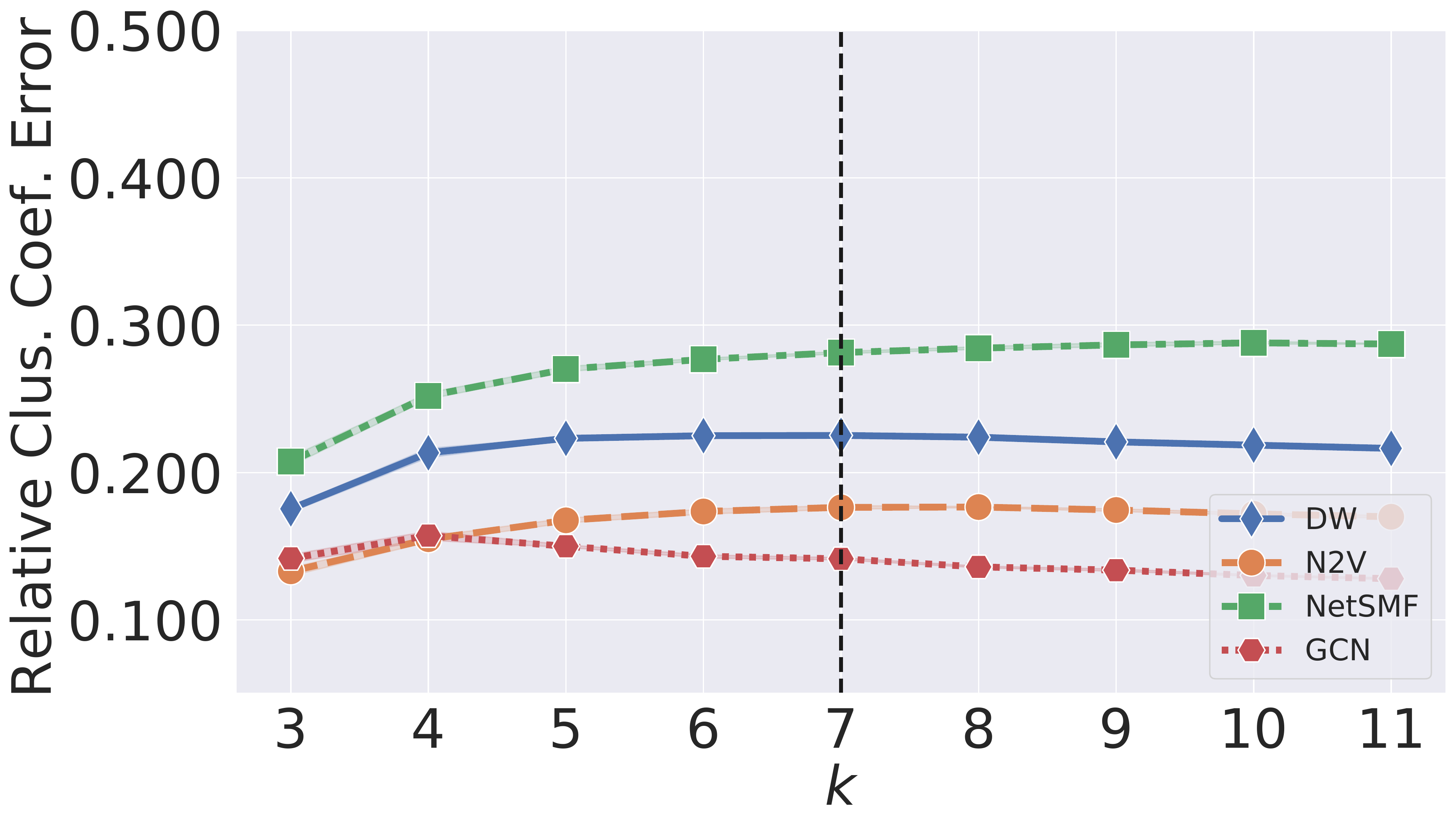}
        \caption{Actor}
        \label{fig:citation_network_k_128_actor_clus}
    \end{subfigure}%
    \hfill
    \begin{subfigure}[b]{0.23\textwidth}

        \includegraphics[width=\linewidth]{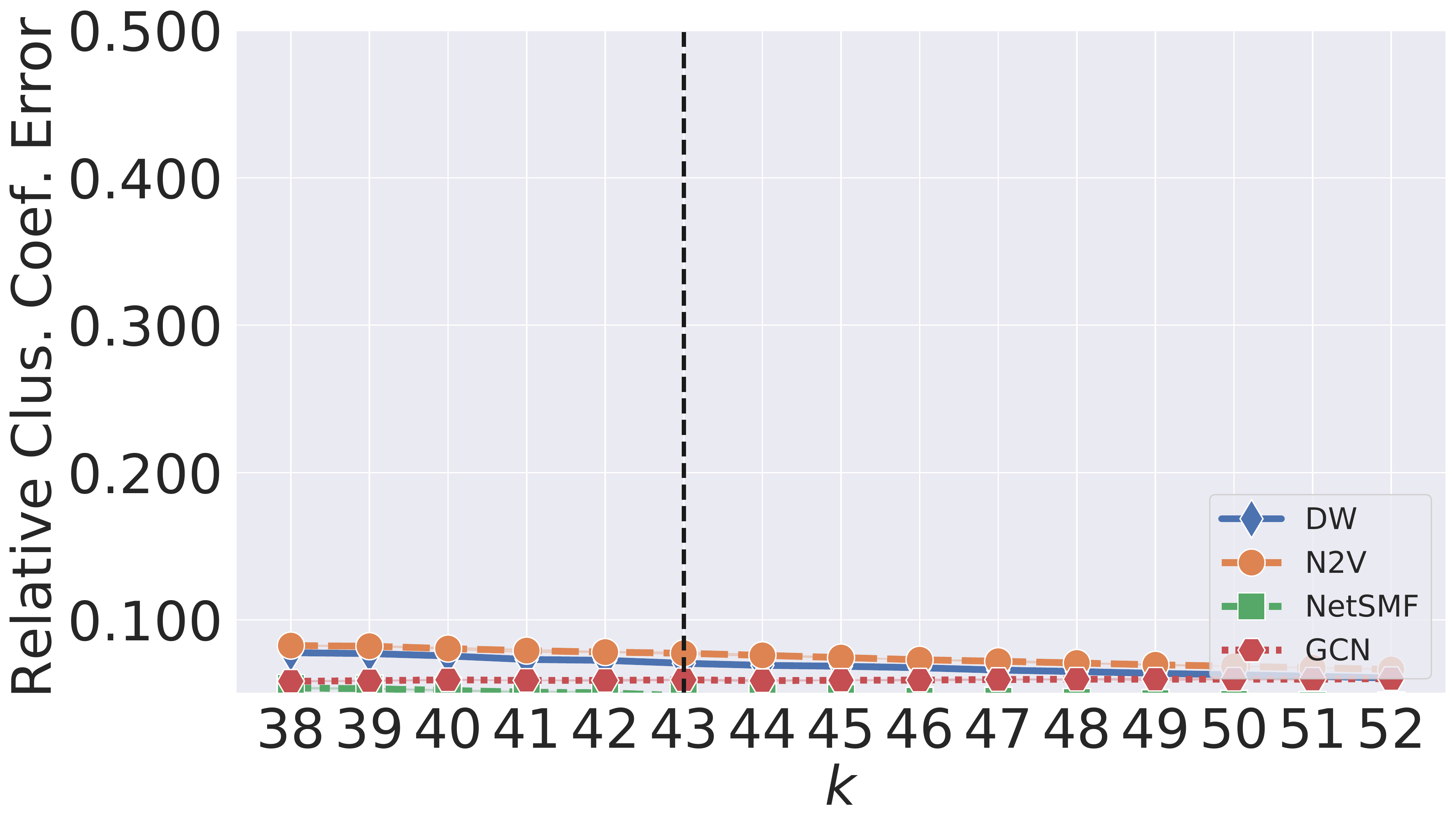}
        \caption{Facebook}
        \label{fig:social_network_k_128_facebook_clus}
    \end{subfigure}
    \caption{F1 scores and relative average clustering coefficient error scores of \approach given all four datasets. We fix the node embedding size to 128. The estimated average node degree of Cora, Citeseer and Actor datasets is 5. The estimated average node degree of Facebook is 46.}
    \label{fig:impact_of_k_128}
\end{figure*}

\begin{figure*}[t]

    \begin{subfigure}[b]{0.23\textwidth}

        \includegraphics[width=\linewidth]{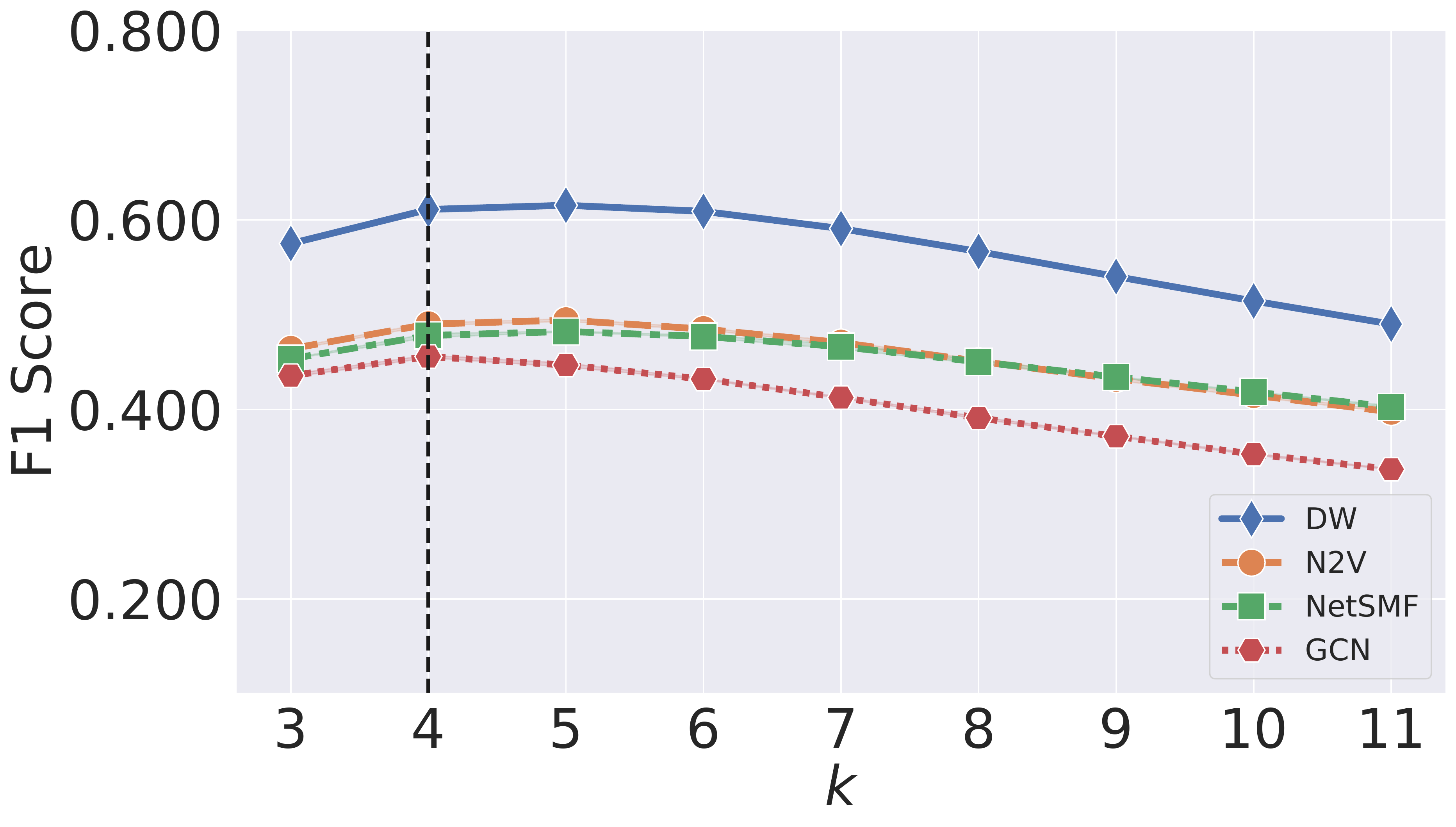}
        \caption{Cora}
        \label{fig:citation_network_k_64_cora_f1}
    \end{subfigure} %
    \hfill
    \begin{subfigure}[b]{0.23\textwidth}

        \includegraphics[width=\linewidth]{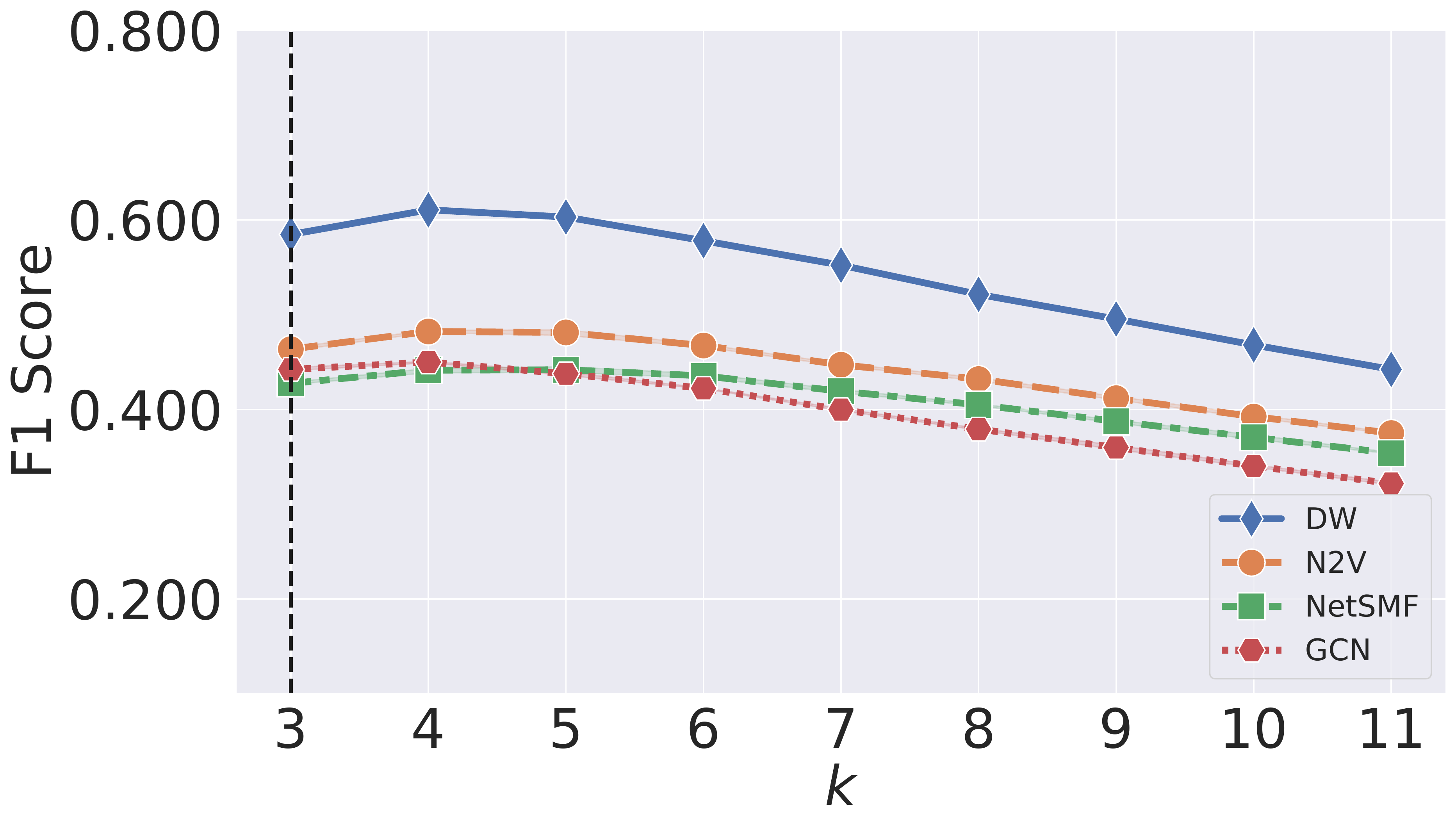}
        \caption{Citeseer}
        \label{fig:social_network_k_64_citeseer_f1}
    \end{subfigure} %
    \hfill
    \begin{subfigure}[b]{0.23\textwidth}

        \includegraphics[width=\linewidth]{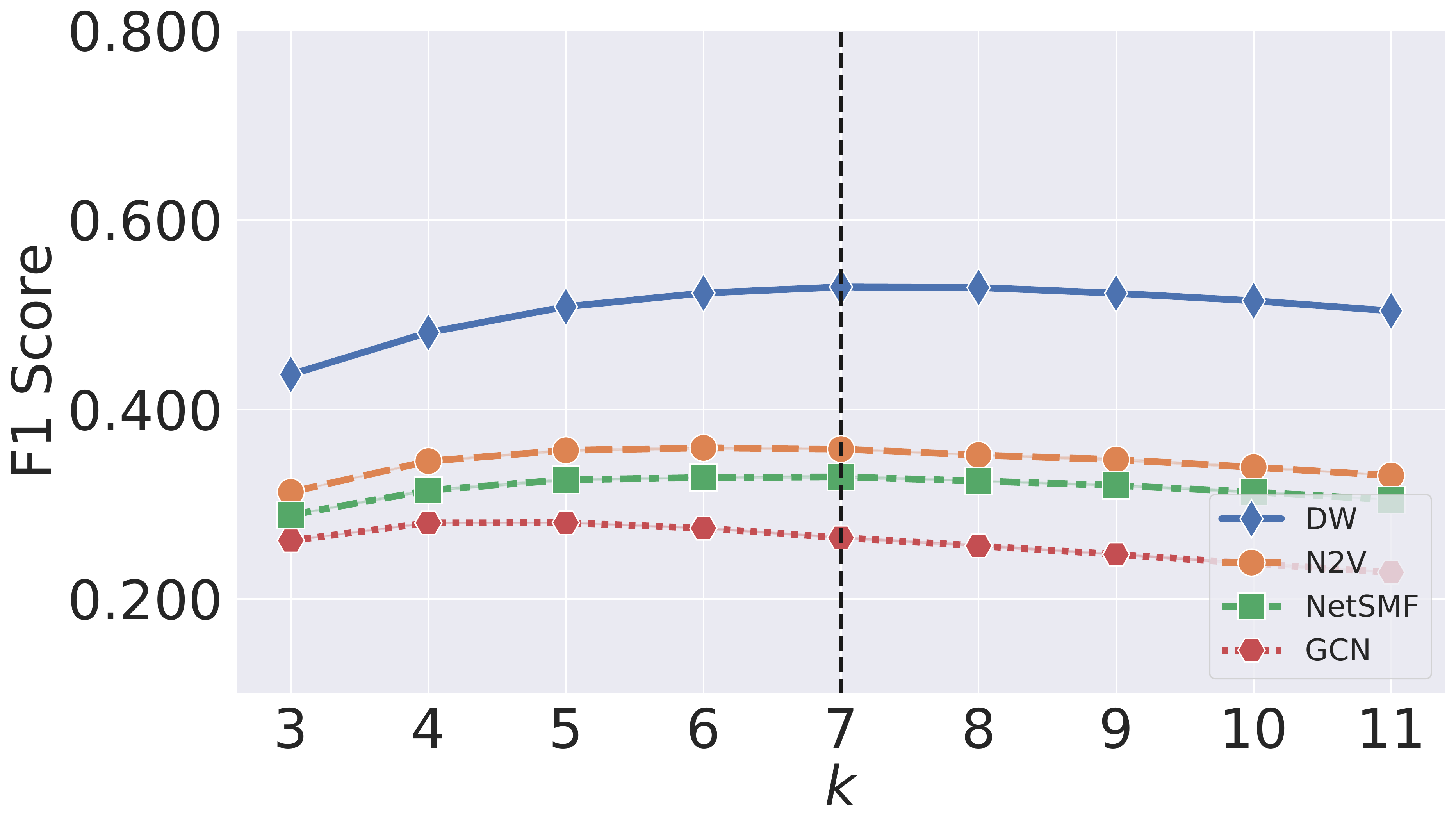}
        \caption{Actor}
        \label{fig:citation_network_k_64_actor_f1}
    \end{subfigure} %
    \hfill
    \begin{subfigure}[b]{0.23\textwidth}

        \includegraphics[width=\linewidth]{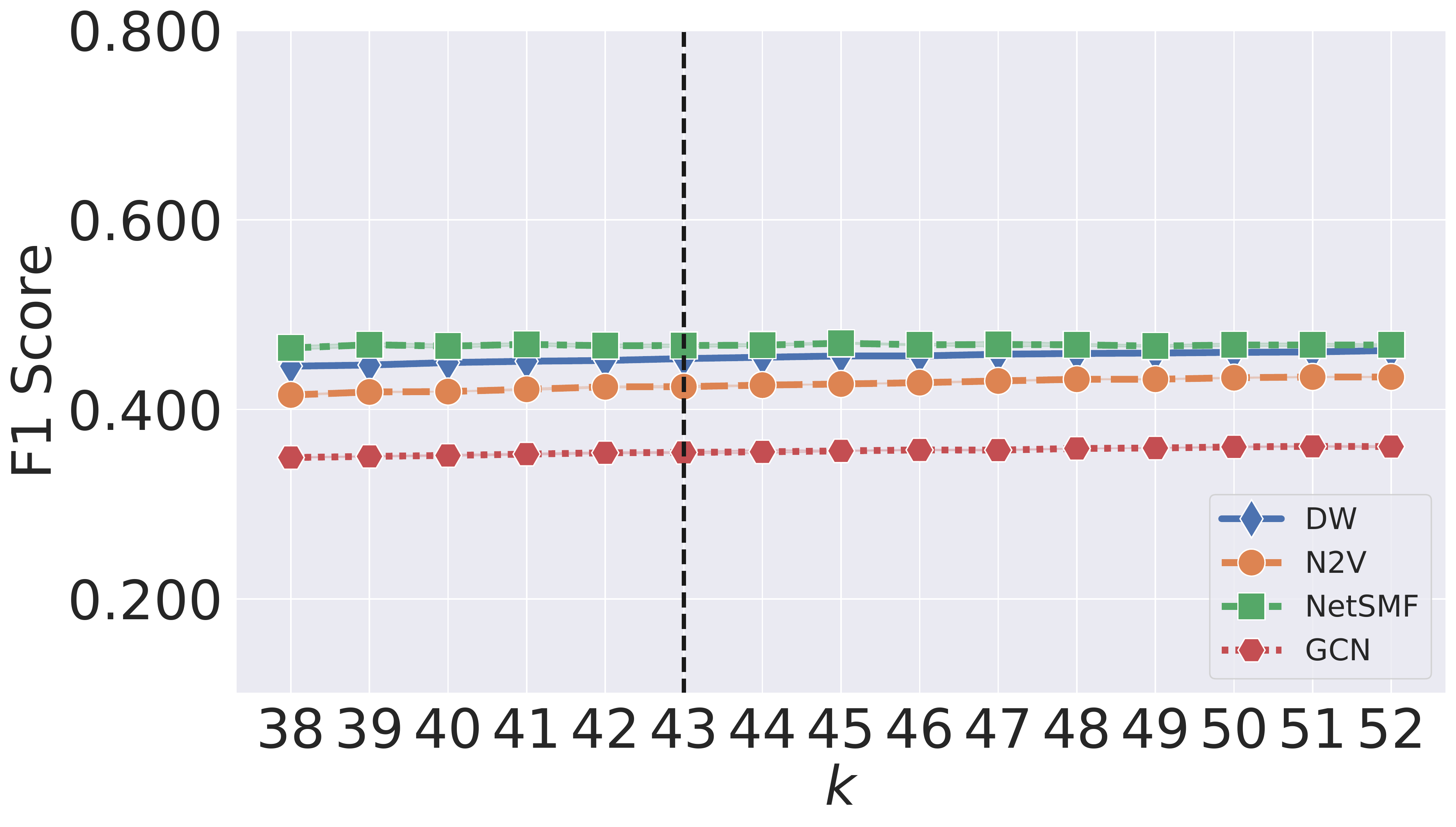}
        \caption{Facebook}
        \label{fig:social_network_k_64_facebook_f1}
    \end{subfigure}%

    \begin{subfigure}[b]{0.23\textwidth}
        \includegraphics[width=\linewidth]{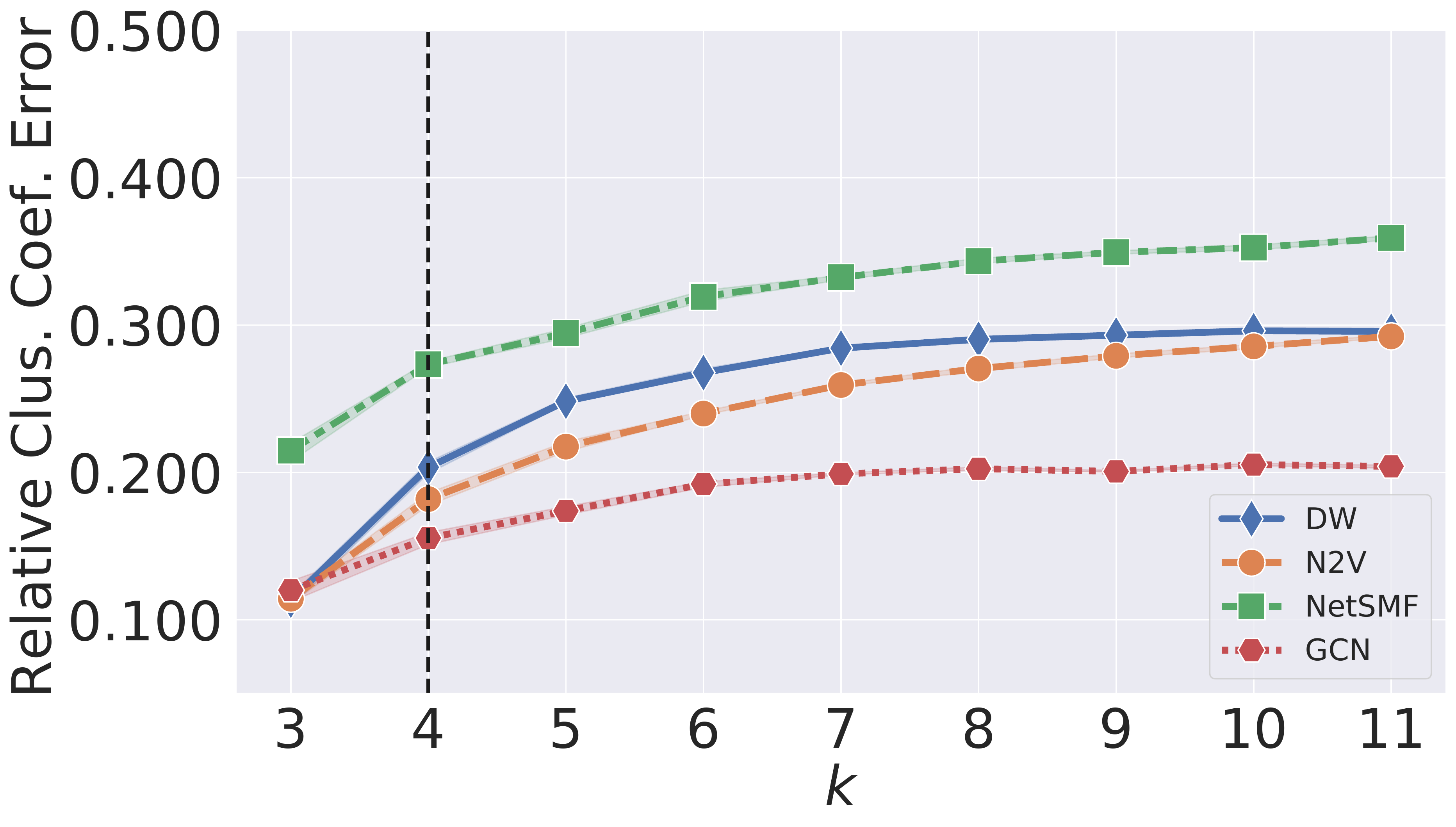}
        \caption{Cora}
        \label{fig:citation_network_k_64_cora_clus}
    \end{subfigure}%
    \hfill
    \begin{subfigure}[b]{0.23\textwidth}
        \includegraphics[width=\linewidth]{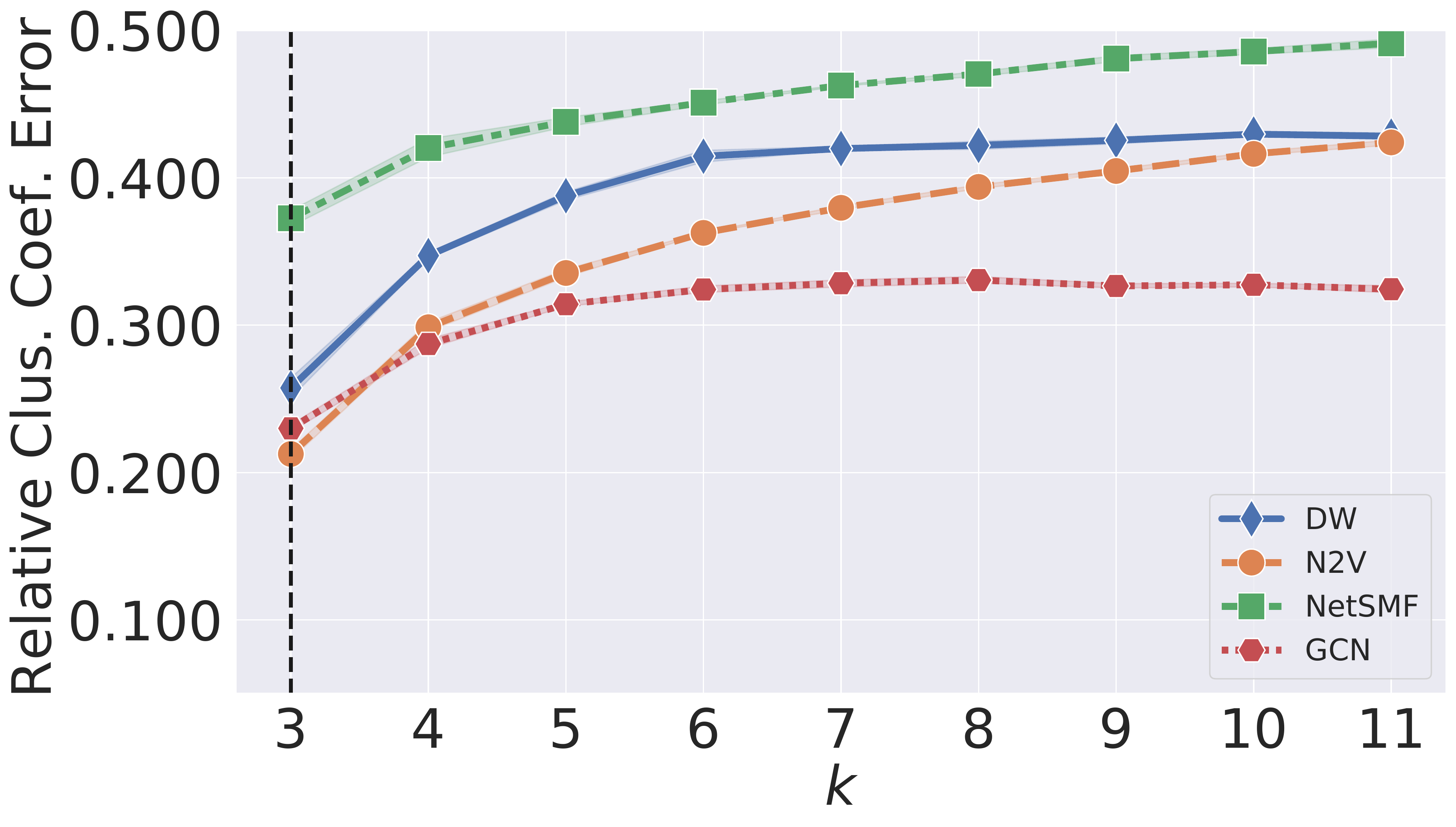}
        \caption{Citeseer}
        \label{fig:social_network_k_64_citeseer_clus}
    \end{subfigure}%
    \hfill
    \begin{subfigure}[b]{0.23\textwidth}
        \includegraphics[width=\linewidth]{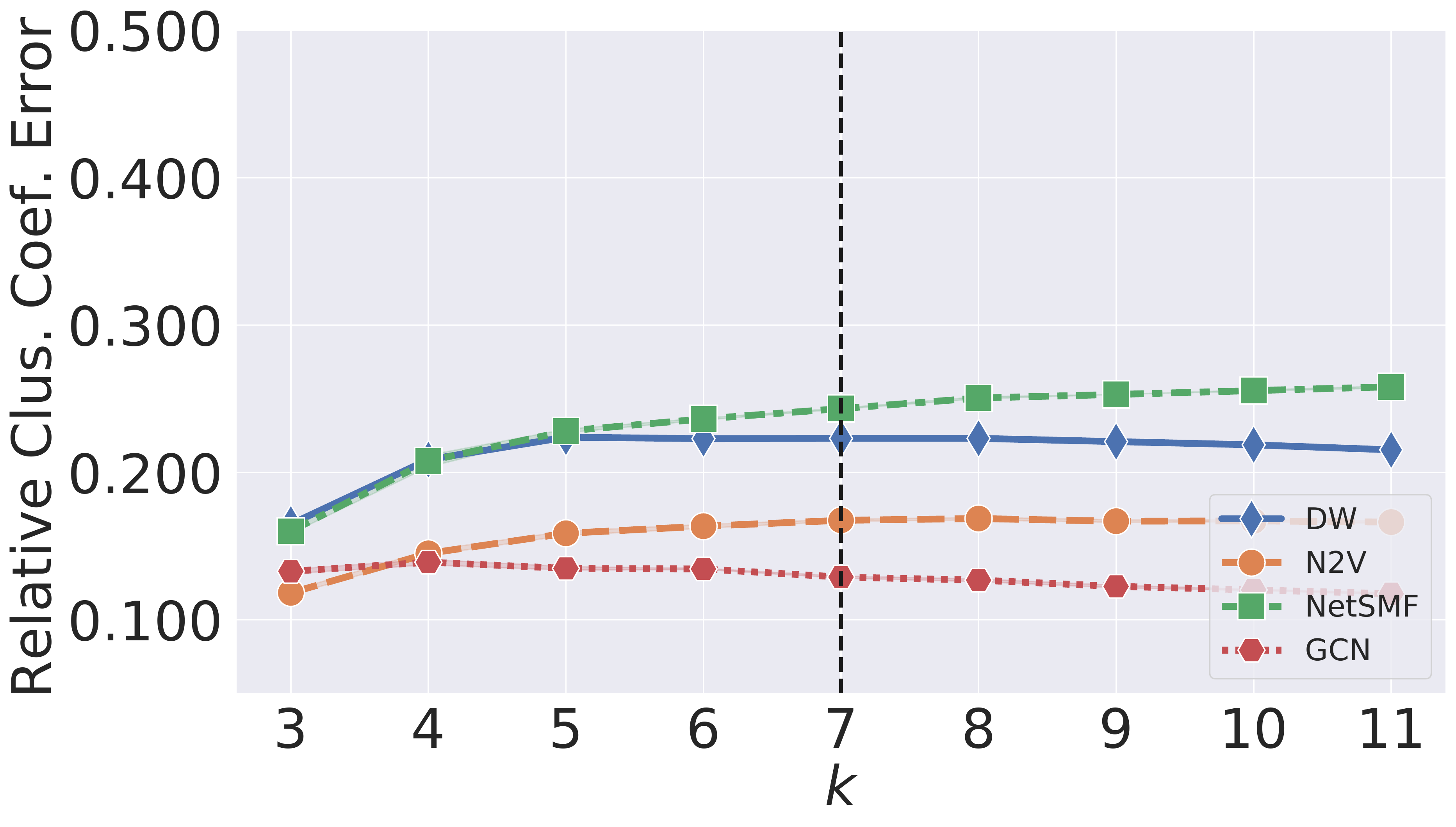}
        \caption{Actor}
        \label{fig:citation_network_k_64_actor_clus}
    \end{subfigure}%
    \hfill
    \begin{subfigure}[b]{0.23\textwidth}

        \includegraphics[width=\linewidth]{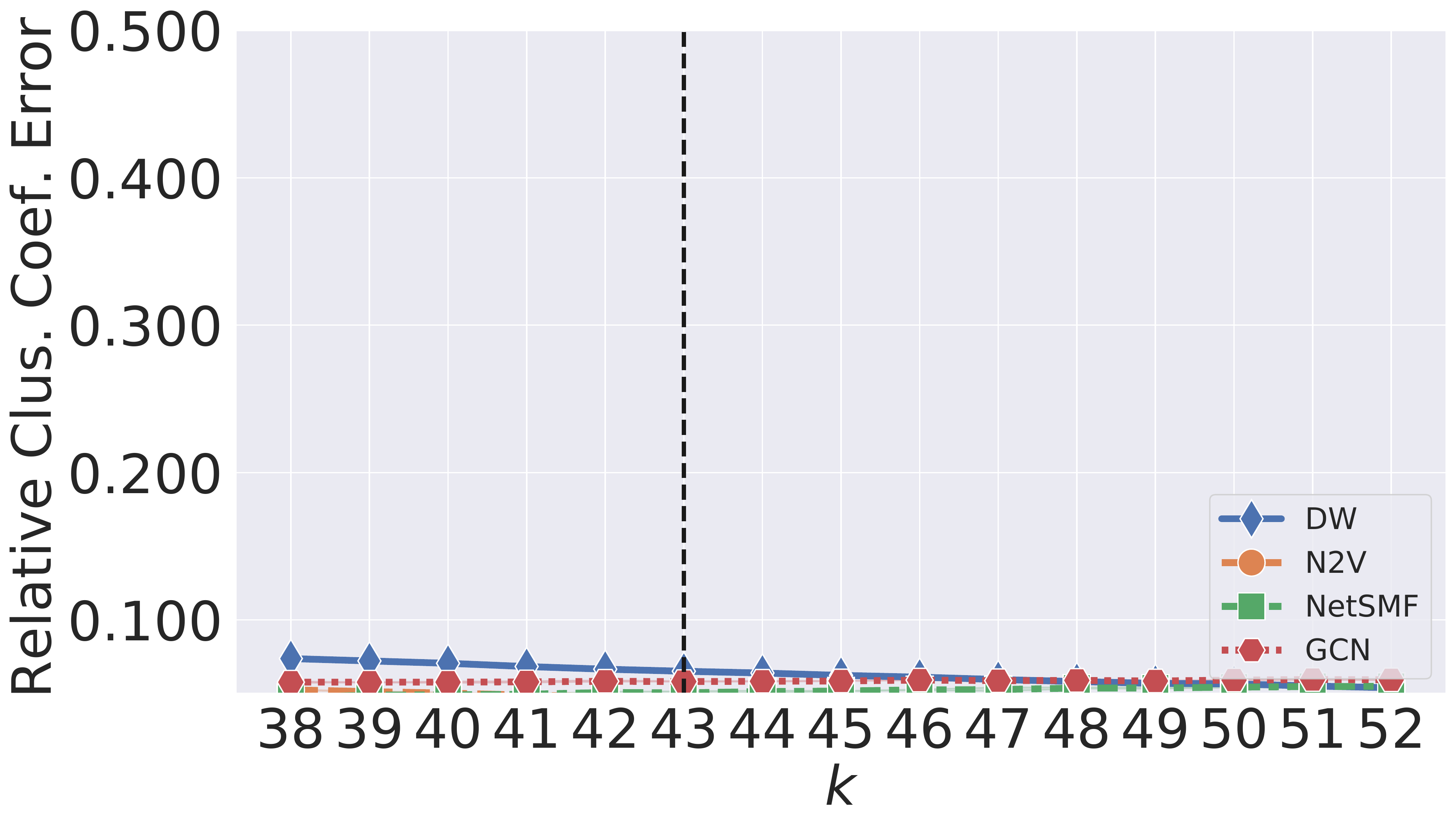}
        \caption{Facebook}
        \label{fig:social_network_k_64_facebook_clus}
    \end{subfigure}

    \caption{F1 scores and relative average clustering coefficient error scores of \approach given all four datasets. We fix the node embedding size to 64. The estimated average node degree of Cora, Citeseer and Actor datasets is 5. The estimated average node degree of Facebook is 46.}
    \label{fig:impact_of_k_64}
\end{figure*}

\mypara{Additional Experimental Results}
We list additional experimental results using all four datasets in \autoref{fig:impact_of_k_128} and \autoref{fig:impact_of_k_64}.

\end{document}